\documentclass[preprint,12pt,authoryear,nonatbib]{elsarticle}

\usepackage{amssymb}
\usepackage{amsmath}

\usepackage{url}
\usepackage{colortbl}
\usepackage{multirow} 
\usepackage{arydshln}
\usepackage{pdflscape}
\usepackage{booktabs}
\usepackage{caption}
\usepackage{xcolor}
\usepackage[cal=boondox]{mathalfa}
\usepackage[natbibapa]{apacite}

\newcolumntype{x}[1]{>{\centering\arraybackslash\hspace{0pt}}p{#1}}
\newcolumntype{m}[1]{>{\arraybackslash\hspace{0pt}}p{#1}}

\journal{}
\usepackage[margin=3cm]{geometry}
\begin{document}

\begin{frontmatter}

\title{Realized Volatility Forecasting for New Issues and Spin-Offs using Multi-Source Transfer Learning}

\author[label1]{Andreas Teller}

\affiliation[label1]{organization={Faculty of Economics and Business Administration, Friedrich Schiller University Jena},
            addressline={Carl-Zeiß-Straße 3}, 
            city={Jena},
            postcode={07743}, 
            country={Germany}}

\author[label2]{Uta Pigorsch\corref{cor1}}

\affiliation[label2]{organization={Schumpeter School of Business and Economics, University of Wuppertal},
            addressline={Gaußstraße 20}, 
            city={Wuppertal},
            postcode={42119}, 
            country={Germany}}

\author[label1]{Christian Pigorsch}

\begin{abstract}
    Forecasting the volatility of financial assets is essential for various financial applications. This paper addresses the challenging task of forecasting the volatility of financial assets with limited historical data, such as new issues or spin-offs, by proposing a multi-source transfer learning approach. Specifically, we exploit complementary source data of assets with a substantial historical data record by selecting source time series instances that are most similar to the limited target data of the new issue/spin-off. Based on these instances and the target data, we estimate linear and non-linear realized volatility models and compare their forecasting performance to forecasts of models trained exclusively on the target data, and models trained on the entire source and target data. The results show that our transfer learning approach outperforms the alternative models and that the integration of complementary data is also beneficial immediately after the initial trading day of the new issue/spin-off.
\end{abstract}

\begin{keyword}
Realized Variance \sep Transfer Learning \sep Machine Learning \sep New Issues \sep Data Scarcity
\end{keyword}
\end{frontmatter}

\clearpage

\section{Introduction}
\label{Introduction}
\noindent
Volatility forecasting is of significant importance in financial market trading and risk management. In the past, a range of methods for predicting future volatility has emerged, starting with the stochastic volatility models of \citet{taylor1982financial}, and the ARCH and GARCH models of \citet{engle1982autoregressive} and \citet{bollerslev1986generalized}, respectively. The availability of high-frequency financial data has led to the development of new volatility measures like realized volatility \citep{andersen2001distributionexchange} and corresponding forecast models such as the the heterogeneous autoregressive (HAR) model proposed by \citet{corsi2009simple}. Moreover, numerous authors have explored the application of data-driven machine learning methods for realized volatility forecasting. For example, \citet{luong2018forecasting} utilize random forest models to predict future volatility. Extreme gradient boosting (XGBoost) models for realized volatility forecasting across different S\&P 500 assets and forecast horizons are applied by \citet{teller2022short}. \citet{bucci2020realized} forecast realized volatility using various artificial neural network models. For a detailed comparison of different machine learning algorithms evaluated for realized volatility forecasting in a joint framework, we refer to \citet{christensen2022amachine}. 

Interestingly, all of the aforementioned studies base their empirical evaluation of realized volatility forecasting models on financial assets with a substantial historical data record. On the other hand, this is not surprising, as even parsimonious forecasting methods such as the HAR model rely on adequate data set sizes to generate accurate forecasts. However, this also means that the literature has not yet considered realized volatility forecasts for financial assets with very limited historical data, i.e., financial new issues and spin-offs. It can be assumed that the main reason for this shortage in the existing research is the lack of historical data at the beginning of a financial asset's life cycle. In this paper, we propose to overcome this limitation through complementary data from financial assets with a substantial data history in combination with transfer learning. 

Transfer learning represents a common knowledge transfer approach in fields such as natural language processing \citep{houlsby2019parameter, blitzer2006domain}, computer vision \citep{deng2018active, hussain2019study}, medical applications \citep{lan2018domain, goetz2015dalsa}, and in industrial settings \citep{zhou2023time}. In financial contexts, transfer learning has been considered by \citet{zhang2023novel}, who forecast financial return series using an adversarial domain adaptation model, and by \citet{he2023instance}, who predict stock price movements using a transfer learning attention mechanism. Furthermore, \citet{he2020multi} propose multi-source transfer learning methods to forecast 1-day-ahead stock prices. However, the existing literature has largely ignored transfer learning for volatility forecasting in general and for predicting the future volatility of new issues or spin-offs in particular. To the best of our knowledge, \citet{liu2023deepvol} is the only study that explicitly addresses the use of transfer learning for volatility forecasting. Their main focus is on developing a global volatility model trained on pooled data from multiple assets, which, in their empirical analysis, outperforms models trained exclusively on the target data of individual assets. Although they incorporate a fine-tuning stage to transfer knowledge from the global model to individual asset forecasting, this approach, contrary to prior expectations, does not significantly improve forecasting performance compared to the standalone global model. Moreover, their study exclusively focuses on stocks with at least five years of historical data during the training period, excluding those with limited data histories, such as new issues or spin-offs. Thus, while \citet{liu2023deepvol} demonstrate an initial application of transfer learning for volatility forecasting, it neither delivers notable forecast improvements nor addresses the challenges posed by assets with sparse historical data.

We address this research gap on volatility forecasting for new issues and spin-offs. To this end, we propose an instance-based multi-source transfer learning approach that selects subsequences from complementary financial assets based on the dynamic time warping (DTW) distance between the new issue/spin-off and complementary data series. The selected subsequences are then combined with all the available target data for model training. Our proposed transfer learning approach has the advantage of being model-agnostic, meaning it can be applied to any type of model one might consider for forecasting realized volatility. In this paper, we integrate our transfer learning method with HAR models, feedforward neural networks (FNNs), and XGBoost models. Thus, we use linear time series models and non-linear machine learning models that are well established and commonly used for forecasting realized volatility. 

Furthermore, we compare our transfer learning approach with a naive pooling method, combining new issue/spin-off with the entire complementary data sets as examined by \citet{liu2023deepvol} and \citet{frank2023forecasting}, as well as with forecasting models solely estimated on the limited historical data of the new issues/spin-offs. Each model and training data set combination is evaluated for 1-day-ahead realized variance forecasts on 10 new issues and spin-offs from 50 to 550 days post-distribution date. Additionally, we examine the subsequences selected by our transfer learning approach to gain further insights into their properties. 

Finally, we conduct a second forecast assessment to evaluate the forecast precision immediately after the first trading day of the new issue/spin-off. This includes the most extreme and practically interesting scenario: forecasting realized volatility after only one day of trading.

The remainder of this paper is organized as follows. Section \ref{tl_for_rv} provides a introduction to transfer learning and introduces our novel instance-based transfer learning approach for forecasting volatility. Section \ref{sec:rv_data_pred_sets} presents our data sets, the construction of the realized variance series and the predictors we employ in this paper. The forecasting models considered (HAR models, FNNs, and XGBoost models) are detailed in Section \ref{sec:rv_forecasting_models}. Section \ref{sec:experimental_setup} presents the empirical setup including data preparation steps and model estimation. The empirical results of our forecast evaluation study are reported in Section \ref{sec:results}. Section \ref{sec:conclusion} concludes.

\section{Transfer Learning for Volatility Forecasting}
\label{tl_for_rv}
\noindent
This section establishes a methodology that allows for leveraging information from financial assets with a considerable data history, in order to improve the prediction of daily volatility of data-scarce new issues and spin-offs. Importantly, our approach enables the prediction of volatility immediately after the first trading day, addressing the challenge of limited historical data for these newly listed assets.

Before introducing our proposed transfer learning approach for predicting the volatility of new issues and spin-offs in Section \ref{sec:da_rv}, we first review transfer learning in general and then focus on different instance-based transfer learning methods. Since historical volatility observations (e.g., realized volatility) are naturally structured as time series, we primarily consider instance-based transfer learning approaches in time series contexts.

\subsection{Transfer Learning}
\label{sec:transfer_learning}
\noindent
Transfer learning, as a paradigm, addresses the challenges of knowledge transfer across disparate tasks and domains. Following the notation of \citet{pan2009survey}, we define a domain by $\mathcal{D} =(\mathcal{X},P(X))$, where $\mathcal{X}$ is the feature space, $P(X)$ is the marginal probability distribution of the domain, and $X$ represents a learning sample, where $X = \{x_{1}, \dots, x_{n}\} \in \mathcal{X}$. Given a domain, $\mathcal{D}$, a task is defined by $\mathcal{T} = \{\mathcal{Y}, \mathcal{h}(\cdot)\}$, where $\mathcal{Y}$ is the label space and $\mathcal{h}(\cdot)$ is a predictive function. This predictive function can be learned from the training data pairs $\{x_{i}, y_{i}\}$, where $x_{i} \in X$ and $y_{i} \in \mathcal{Y}$. Tasks and domains to which knowledge is transferred are commonly referred to as target tasks, $\mathcal{T}_{T}$, and target domains, $\mathcal{D}_{T}$, while tasks and domains from which knowledge is derived are referred to as source tasks, $\mathcal{T}_{S}$, and source domains $\mathcal{D}_{S}$. The specific source domain data, ${D}_{S}$, is denoted by $D_{S} = \{(x_{S,1}, y_{S,1}), \dots, (x_{S,i}, y_{S,i}), \dots, (x_{S,n}, y_{S,n}) \}$, where $x_{S,i} \in \mathcal{X}_{S}$ is the $i$-th data instance and $y_{S,i} \in \mathcal{Y}_{S}$ is the corresponding label. Similarly, the target domain data, $D_{T}$, is defined by $D_{T} = (x_{T,1}, y_{T,1}), \dots, (x_{T,i}, y_{T,i}), \dots, (x_{T,n}, y_{T,n})$, where $x_{T,i} \in \mathcal{X}_{T}$ is the $i$-th data instance of the target domain data and $y_{T,i} \in \mathcal{Y}_{T}$ is the corresponding label. Based on $\mathcal{T}_{S}$ and $\mathcal{D}_{S}$, transfer learning seeks to improve the learning of the target predictive function $\mathcal{h}_{T}(\cdot)$ in $\mathcal{D}_{T}$, where $\mathcal{D}_{S} \neq \mathcal{D}_{T}$ or $\mathcal{T}_{S} \neq \mathcal{T}_{T}$. In general, transfer learning can also include the transfer of knowledge from multiple source domains \citep{weiss2016survey}.

According to \citet{pan2009survey}, transfer learning can be further categorized into deductive transfer learning for $\mathcal{D}_{S} \neq \mathcal{D}_{T}$ and inductive transfer learning for $\mathcal{T}_{S} \neq \mathcal{T}_{T}$, where the domains for the latter can be the same or different. In the deductive transfer learning setting, target data labels are usually not available, or the transfer learning method itself does not necessitate such labels. Inductive transfer learning assumes that at least a small number of target labels are available to induce the target predictive function. Unsupervised transfer learning refers to scenarios where no labels are available in the target and source data. 

In addition, \citet{pan2009survey} identify four categories of transfer learning methods based on the question of what is actually being transferred: parameter-based, feature-based, relational-knowledge-based, and instance-based transfer learning methods. Parameter-based approaches adapt a model trained on one task to perform a different but related task by reusing and fine-tuning its parameters (examples in \citet{liu2023deepvol}, \citet{guo2019spottune}, and \citet{fahimi2019inter}). Feature-based methods, on the other hand, focus on either encoding one domain into a feature representation that more closely matches the other domain, or transforming the target and source domains into a common feature space to minimize the differences between the feature spaces of the target and source domains, see, e.g., \citet{blitzer2006domain} and \citet{argyriou2006multi}. Relational-knowledge-based methods leverage structured, domain-specific knowledge, such as social network data (e.g. \citet{li2012cross}) from one task to enhance learning in a related but different task. Lastly, instance-based methods assume that some instances of the source domain are more relevant to the target domain than others and that these instances can be applied directly to train a target learner (examples in \citet{jiang2007instance} and \citet{dai2007boosting}). 

When predicting the 1-day-ahead volatility of new issues/spin-offs based on past volatility observations, we encounter data sets that inherently contain labeled data. However, due to the sparse amount of labeled data available immediately after the first trading day of new issues/spin-offs, we deliberately disregard parameter-based methods, as they commonly require a large number of labeled target observations. Similarly, we do not consider feature-based methods, since they rely on the estimation of feature mappings based on target observations, which are scarce for new issues and spin-offs. Since relational-knowledge-based methods, while effective in domains with clear relational structures such as graphs or networks, are not applicable to time series transfer learning, we focus primarily on instance-based approaches in an inductive time series transfer learning framework. Specifically, we adopt a multi-source transfer learning approach to facilitate knowledge transfer from multiple financial assets, i.e., source domains.

\subsection{Time Series Instance and Source Selection}
\label{sec:inst_and_source_select}
\noindent
Instance-based transfer learning methods, with respect to time series problems, are concerned with the selection and weighting of instances from a source data set, where instance in this context refers to a single time series \citep{weber2021transfer}. Since time series instance weighting methods, as noted by \citet{weber2021transfer}, are primarily applied in the context of ensemble models, we focus exclusively on instance selection methods. To determine which instances are most likely to improve a target learner, instance selection methods typically consider a similarity measure, i.e., some notion of distance or divergence, between the target and source domain time series. This similarity measure serves as a proxy for the difference between the marginal probability and/or conditional probability distributions of the target and source domain data sets. By including only those instances that are most related to the target domain, instance selection methods mitigate the risk of negative transfer, a scenario in which the inclusion of certain source instances leads to diminished prediction accuracy \citep{wang2019characterizing}. 

In the related literature, different approaches have been introduced to evaluate the similarity between target and source instances. For example, \citet{vercruyssen2017transfer} propose to select a subset of source instances for target learner training, either based on the affiliation and proximity to target data clusters, or by applying a density-based method. In the context of source selection, which refers to the selection of one or more data sets consisting of multiple time series from different domains, several studies have adopted thresholds on distance and similarity measures to select source data sets that are relevant to the target domain. These distance and similarity measures include the DTW distance \citep{fawaz2018transfer}, Jensen-Shannon divergence \citep{ye2021implementing}, Pearson correlation \citep{xiao2014transfer}, or maximum mean discrepancy \citep{islam2019evaluation}. Although source selection is categorized by \citet{weber2021transfer} as a preprocessing step for instance selection or other transfer learning methods, the aforementioned similarity metrics are equally applicable to instance selection itself.

\subsection{Transfer Learning for Forecasting Volatility of New Issues/Spin-Offs}
\label{sec:da_rv}
\noindent
Building on the time series instance and source selection methods discussed in the previous section, we propose a multi-source transfer learning approach for predicting the 1-day-ahead volatility of new issues and spin-offs in the time period after their first trading day. 

To outline our method, we define a combined source data set $D_{S}$ consisting of $P$ financial assets with an extensive data history by 
\begin{equation}
D_{S} = \left\{D_{S_{p}} \mid D_{S_{p}} = \left\{(x^{(D_{S_{p}})}_{t}, y^{(D_{S_{p}})}_{t}) \right\}_{t=1}^{N^{(D_{S_{p}})}}, p = 1, 2, \ldots, P \right\},
\end{equation}
where $D_{S_{p}}$ denotes the data set of the $p$-th financial asset, $x^{(D_{S_{p}})}_{t}$ represents the input features for day $t$ of asset $p$, which include past volatility measures such as lagged daily volatility as well as lagged weekly and monthly volatility components, as defined in Section \ref{sec:rv_and_predictor_sets}. Furthermore, $y^{(D_{S_{p}})}_{t}$ represents the label that is to be predicted on day $t$ for asset $p$, i.e., the daily volatility for the following day, $t+1$, and $N^{(D_{S_{p}})}$ denotes the total number of daily observations for asset $p$. In addition, we define the data set of a target asset, i.e., the new issue/spin-off, $T$ by
\begin{equation}
D_{T} = \{(x_t^{(D_{T})}, y_t^{(D_{T})})\}_{t=1}^{N^{(D_{T})}},
\end{equation}
where $y_t^{(D_{T})}$ denotes the daily volatility of the target asset for day $t+1$ and $x^{(D_{T})}_{t}$ represents the input features of the target asset for day $t$, again consisting of lagged daily volatility as well as lagged volatility components. Note that the specific composition of the input features depends on the forecast horizon. For shorter forecast horizons, particularly those less than one month or one week, certain components may be unavailable and therefore excluded from the input features. For instance, when predicting volatility one day after the first trading day, the input features include only the first lag of daily volatility. $N^{(D_{T})}$ represents the number of observations in $D_{T}$ available for the target asset. 

Since our goal is to forecast the volatility of new issues/spin-offs in the period following their first trading day, $N^{(D_{T})}$ is usually small, making it difficult to predict future volatility. Therefore, we aim to extend $D_{T}$ by including additional data from $D_{S}$, which is most similar to $D_{T}$. However, this approach poses two challenges. First, for source assets with a long data history, there is a significant disparity in the number of observations of $D_{T}$ and $D_{S_{p}}$ complicating the application of time series similarity measures such as the Euclidean distance. Second, as highlighted by previous research \citep[among others]{choi2010long, yang2014realized}, volatility series are observed to exhibit structural breaks, which can potentially lead to biased similarity measurements with respect to large source asset data sets. 

To address these issues, we split the source series into non-overlapping subsequences and selectively transfer those subsequences that show the highest similarity to $D_{T}$, rather than incorporating the entire source asset data set into the extended training data set of the respective new issue/spin-off. The utilization of subsequences to account for time series of different lengths has also been considered by \citet{xiao2014transfer} in the context of forecasting port container throughput. The choice of non-overlapping subsequences over overlapping ones is primarily based on considerations of computational efficiency. Overlapping subsequences necessitate a greater number of subsequent similarity measurements, significantly increasing the computational burden. 

For a source asset $D_{S_{p}}$, we define the set of non-overlapping subsequences as follows. Let $m$ represent the length of a single subsequence, $K$ be the total number of non-overlapping subsequences that can be generated from the data set, where $K = \left\lfloor\frac{N^{(D_{S_{p}})}-e}{m}\right\rfloor$, and $e$ denoting the number of excess observations, calculated as $e = N^{(D_{S_{p}})} \pmod m$. The corresponding set of non-overlapping subsequences is then given by
\begin{equation}
\begin{split}
D_{S_{p}}^{(u)} = & \left\{D_{S_{p,k}} \mid D_{S_{p,k}} = \{z^{(D_{S_{p}})}_{t=(e+1) + (k-1)m}, z^{(D_{S_{p}})}_{t=(e+1) + (k-1)m + 1}, \ldots, \right.\\ 
& \left. z^{(D_{S_{p}})}_{t=(e+1) + km - 1}\}, k = 1, 2, \ldots, K \right\},
\end{split}
\end{equation}
where $z^{(D_{S_{p}})}_{t=(e+1) + (k-1)m}$ represents its respective input feature and label pair: 
\begin{equation}
(x^{(D_{S_{p}})}_{t=(e+1) + (k-1)m}, y^{(D_{S_{p}})}_{t=(e+1) + (k-1)m}). 
\end{equation}
Subsequences generated in this way omit excess observations at the beginning of each source asset data set. The interval length ($m$) inherently represents a trade-off, where small values of $m$ can also lead to an increased computational load due to the number of similarity comparisons required and a decreased robustness with respect to multi-day model re-estimation intervals, while large values of $m$ lead to large subsequences that potentially contain structural breaks. The resulting combined set, containing the subsequences of all elements of $D_{S}$, is defined by
\begin{equation}
U = D_{S_{1}}^{(u)} \cup D_{S_{2}}^{(u)} \cup \dots \cup D_{S_{P}}^{(u)}.
\end{equation}

The similarity between the target asset and the source subsequences in $U$ is assessed through the DTW distance, which represents a more flexible approach than the Euclidean distance. In contrast to the Euclidean distance, which requires a one-to-one alignment and is sensitive to temporal shifts or distortions, DTW allows for non-linear alignments, enabling it to effectively capture similarities between time series that may evolve at different rates \citep{ratanamahatana2004everything}. We propose computing the DTW distance between the lagged volatility predictor series (i.e., lagged daily volatility and lagged volatility components) of the target asset and the generated source subsequences. To ensure an accurate comparison with the target asset data, we address structural breaks and align the sequences by considering only the most recent, i.e., the last, $m$ input feature values of $D_{T}$. Note that this also includes the input features for the day of the forecast ($x^{(D_{T})}_{N^{(D_{T})}+1}$), which notably lack a corresponding output label. Although a direct comparison of recent target observation labels with the labels of subsequences from $U$ is not performed, such a comparison can be implicitly facilitated via the lagged daily volatility. Thus, we account not only for the similarity between the marginal probability distributions, but also indirectly for the similarity between the conditional probability distributions. A more detailed explanation of DTW is provided in \ref{sec:dtw}.

After computing the DTW distance between the most recent target observations and each subsequence within $U$, we selectively transfer into the training data set $D_{T}$ only those subsequences whose DTW distance ranks below a predefined percentile $\epsilon$ of the empirical DTW distance distribution across all source subsequences. In effect, $\epsilon$ serves as a selection threshold, determining the volume of source subsequences to be included in the new issues/spin-off training data set. The resulting pooled data set, which contains the entire available target data along with the selected source data, can then be used to train an arbitrary realized variance prediction model. In this paper, we use the resulting data set to estimate HAR, FNN, and XGBoost models.

In summary, our proposed transfer learning approach allows forecasting models to predict the volatility of target assets with limited data availability by leveraging data from source assets with more extensive historical records. This method effectively addresses the issue of data scarcity for target assets and potentially improves the accuracy and robustness of volatility forecasts. At the same time, it accounts for differences in the length of the target and source data sets as well as for potential structural breaks. Moreover, unlike other model- and feature-based transfer learning methods, our approach necessitates only a minimal amount of target data and can be combined with any realized variance forecasting model. Before discussing the models employed in this paper, we first introduce our data sets along with the predictors considered. Detailed information on the implementation of our transfer learning approach is provided in Section \ref{sec:experimental_setup}.

\section{Data, Realized Variance, and Predictor Sets}
\label{sec:rv_data_pred_sets}
\noindent
In the following, we introduce the target data set containing the new issues/spin-offs and the source data set containing complementary stock data. We also briefly present two sets of predictors that we use to forecast realized variance.

\subsection{Data}
\label{sec:data}
\noindent
As our primary target data set, we use 1-minute intraday stock price data obtained from FirstRate Data LLC on 10 new issues and spin-offs that entered the public market between 2010 and 2020 and were part of the S\&P 500 index on November 1, 2022. To ensure a representative sample, we select one new issue or spin-off from each Global Industry Classification Standard (GICS) sector \citep{gics}. Stocks from the \emph{utilities} sector are not included in our target data set, as none of these stocks started public trading during our observation period. Table~\ref{tab:target_assets} provides an overview of the new issues and spin-offs considered and their respective GICS sectors.

Furthermore, for our source data set, we consider the 1-minute intraday price data of 66 additional S\&P 500 stocks. Again, the data is retrieved from FirstRate Data LLC. The assets considered are selected based on their market capitalization within each GICS sector on November 1, 2022. In particular, we select those assets that are ranked in the top two, bottom two, or at the median level of market capitalization among all stocks in the respective GICS sector that began public trading at least one year prior to the new issue or spin-off within the same GICS sector.\thinspace\footnote{If the S\&P 500 GICS sectors consist of an odd number of stocks, the median asset as well as the asset closest to the median in terms of market capitalization are selected. Conversely, if the sectors consist of an even number of stocks, we select the two assets surrounding the median value.} To ensure a comprehensive representation of each GICS sector in the source data set, we also include six assets from the \emph{utility} sector. Again, these assets reflect a broad range of market capitalization, specifically selecting the top two, bottom two, and median-ranked \emph{utility} assets that began trading publicly prior to the most recent new issue/spin-off.

The source data set covers the period between January 1, 2010 and September 30, 2022\thinspace\footnote{The tickers META, QRVO, and AMT contain a limited number of observations because they began public trading after January 1, 2010. Specifically, META (\emph{communication services}) started trading on May 18, 2012, QRVO (\emph{information technology}) started trading on January 2, 2015, and AMT (\emph{real estate}) entered the public market on March 1, 2012. The same holds true for WRK (\emph{utilities}), which began public trading on June 24, 2015, and LIN (\emph{utilities}), which started trading on the New York Stock Exchange on November 29, 2018.}, and is used as complementary data for the naive pooling method and the transfer learning approach of our study. Table~\ref{tab:source_assets} of \ref{sec:appendix_source_data} presents a complete list of all stocks in our source data set.

The raw data from both the target and source data set is processed by excluding entries from 4:00 p.m. EST on Friday through 9:30 a.m. EST on Monday, as well as trading activity between 4:00 p.m. and 9:30 a.m. EST for consecutive business days. 

\begin{table}[t]
\centering
\caption{Overview of new issues and spin-offs considered.}
\scalebox{0.7}{
\begin{tabular}{lcllll}
\hline \hline
\textbf{Ticker}  & \textbf{First Trading Day} & \textbf{Type} & \textbf{Name} & \textbf{GICS Sector}\\ 
\hline
TWTR & 07.11.2013 & New Issue & Twitter Inc. & Communication Services\\
NCLH  & 18.01.2013 & New Issue & Norwegian Cruise Line Ltd. & Communication Discretionary\\
LW  & 10.11.2016 & Spin-Off & Lamb Weston Holdings & Consumer Staples\\
PSX & 01.05.2012 & Spin-Off & Phillips 66 & Energy\\
SYF & 31.07.2014 & Spin-Off & Synchrony Financial & Financials\\
MRNA & 07.12.2018 & New Issue & Moderna Inc. & Health Care\\
CARR  & 03.04.2020 & Spin-Off & Carrier & Industrials\\
DXC  & 03.04.2017 & Spin-Off & DXC Technology & Information Technology\\
CTVA & 24.05.2019 & Spin-Off & Corteva Inc. & Materials\\
INVH & 01.02.2017 & New Issue & Invitation Homes Inc. & Real Estate\\
\hline \hline
\end{tabular}
}
\label{tab:target_assets}
\end{table}

\subsection{Realized Variance and Predictor Sets}
\label{sec:rv_and_predictor_sets}
\noindent
As shown in \citet{andersen2001distributionexchange}, \citet{barndorff2002econometric}, and \citet{andersen1998answering}, the unobservable daily return variation can, under certain assumptions, be consistently estimated by the sum of squared intraday returns as the sampling frequency ($h$) of these returns approaches infinity. The corresponding nonparametric estimator, commonly referred to as daily realized variance, is defined by
\begin{equation}
RV_{t, d} = \sum^{h}_{j = 1} r^{2}_{t,j}, 
\label{eq:daily_rv}
\end{equation}
where $r_{t,j}$ denotes the continuously compounded $j$-th intraday return sampled at frequency $h$ on day $t$. 

Realized variance is typically constructed from 5-minute intraday returns ($h=78$, for standard NYSE and Nasdaq trading hours, 9:30 a.m. to 4:00 p.m. EST), which provides a favorable trade-off between microstructure noise, e.g., caused by bid-ask bounces or rounding errors, and estimation accuracy \citep{andersen2001distributionexchange}. We follow the existing literature \citep[][among others]{ahoniemi2013overnight, christensen2022amachine} and construct daily realized variance series for all target and source assets from 5-minute intraday returns. The objective of this paper is to forecast the 1-day-ahead realized variance ($RV_{t+1, d}$) of a given target asset. 

The related literature considered a variety of different predictors to forecast daily $RV$, see, e.g., \citet{corsi2009simple}, \citet{patton2015good}, \citet{duong2015empirical}, \citet{mittnik2015stock}, \citet{christiansen2012comprehensive}, or \citet{kambouroudis2021forecasting}. In this paper, we primarily use the two sets of predictors recently evaluated by \citet{christensen2022amachine}. They consider a basic predictor set consisting exclusively of the standard HAR volatility components, and an extended predictor set that additionally includes firm-specific and macroeconomic variables. With respect to the more comprehensive predictor set, \citet{christensen2022amachine} report a superior 1-day-ahead forecasting accuracy across a variety of different forecasting models compared to a standard HAR model. In the following, we briefly review both of these predictor sets.

The first predictor set, hereafter referred to as $Q_{std}$, contains the daily ($RV_{t, d}$), weekly, and monthly volatility components of the standard HAR model \citep{corsi2009simple}, where the weekly volatility component is defined by
\begin{equation}
RV_{t,w} = \frac{1}{5} \sum^{4}_{i=0} RV_{t-i, d}, 
\end{equation}
and the monthly volatility component is defined by
\begin{equation}
RV_{t,m} = \frac{1}{22} \sum^{21}_{i=0} RV_{t-i, d}.
\end{equation}

The second predictor set, referred to as $Q_{ext}$ additionally includes the 1-week price momentum defined by
\begin{equation}
MOM_{t} = p_{close, t} - p_{close, t-5}, 
\end{equation}
where $p_{close, t}$ denotes the log closing price of the respective asset on day $t$. It also includes the first-differenced log-transformed dollar trading volume,
\begin{equation}
DV_{t} = \log(\exp(p_{close, t}) \times v_{t}) - \log(\exp(p_{close, t-1}) \times v_{t-1}),
\end{equation}
where $v_{t}$ is the trading volume on day $t$. To capture potential volatility shifts due to earnings information, $Q_{ext}$ also includes a binary earnings announcement indicator ($EA$) which is set to 1 if an earnings announcement is scheduled for the next day and 0 otherwise. Lastly, $Q_{ext}$ contains several macroeconomic indicators: the first-differenced US 3-month Treasury bill rate ($US3M$); the daily squared log return of the Hang Seng stock index ($HSI$); the Aruoba-Diebold-Scotti business conditions index ($ADS$; \citet{aruoba2008real}); the economic policy uncertainty index ($EPU$) introduced by \citet{baker2016measuring}; and the CBOE Volatility Index ($VIX$) which captures market expectations regarding the volatility of the S\&P 500 index.\footnote{The earnings announcement schedule and $HSI$ data has been retrieved from Yahoo Finance. The $US3M$ data is available from \cite{fed2023}, while the $ADS$ data can be accessed in \cite{phil2023}, and the $EPU$ data is provided by \cite{epu2023}. The $VIX$ data has been retrieved from \cite{cboe2023}.} Notably, our extended data set $Q_{ext}$ coincides with that of \citet{christensen2022amachine}, but excludes the implied volatilities of individual assets derived from options contracts, as options contracts are not available immediately after the initial trading day for most new issues and spin-offs.

\section{Realized Variance Forecasting Models}
\label{sec:rv_forecasting_models}
\noindent
Our proposed transfer learning approach offers flexibility as it is model-agnostic, making it compatible with any forecasting model. The wide range of models developed to predict realized volatility of assets with extensive historical data sets provides a diverse set of candidate models to consider. From this substantial pool, we focus on well-established models, selecting both linear time series models and non-linear, data-driven models that have demonstrated robust and strong forecasting performance in prior studies. Specifically, we consider linear HAR models \citep{andersen2007roughing, corsi2012discrete, patton2015good} as well as non-linear approaches such as FNNs and XGBoost models \citep{hamid2004using, mittnik2015stock, christensen2022amachine, bucci2020realized, teller2022short}. We briefly review these models, with an emphasis on the specific configurations used in this paper. 

\subsection{HAR Model}
\label{sec:har_model}
\noindent
The HAR model, originally introduced by \citet{corsi2009simple}, is primarily characterized by its lagged volatility components, which are measured over different temporal horizons, i.e., a day, a week, and a month. This allows the HAR to approximate the slowly decaying (possibly long-memory) autocorrelation often observed in realized variance series. HAR models can be estimated by ordinary least squares (OLS), making them both straightforward to implement and computationally efficient. In its standard form, the HAR model is defined by
\begin{equation}
\label{eq:standard_har}
RV_{t+1, d} = \beta_{0} + \beta_{d} \: RV_{t,d} + \beta_{w} \: RV_{t,w} + \beta_{m} \: RV_{t,m} + \epsilon_{t+1, d}. 
\end{equation}
Additionally, we extend the HAR model with the variables from $Q_{ext}$, as discussed in the previous section, which results in the following HAR-EXT specification: 
\begin{equation}
\begin{split}
RV_{t+1, d} = & \beta_{0} + \beta_{d} \: RV_{t,d} + \beta_{w} \: RV_{t,w} + \beta_{m} \: RV_{t,m} + \beta_{mom} \: mom_{t} \\
& + \beta_{DV} \: DV_{t} + \beta_{EA} \: EA_{t} + \beta_{US3M} \: US3M_{t} + \beta_{HSI} \: HSI_{t} \\
& + \beta_{ADS} \: ADS_{t} + \beta_{EPU} \: EPU_{t} + \beta_{VIX} \: VIX_{t} + \epsilon_{t+1, d}.
\end{split}
\label{eq:har_ext}
\end{equation}
We consider both the standard HAR based on $Q_{std}$, in the following referred to as HAR-STD, and its more comprehensive counterpart, the HAR-EXT, which allows for a direct comparison between $Q_{std}$ and $Q_{ext}$. Due to their popularity and linear model specification, the HAR specifications can be regarded as reference models against which we compare non-linear machine learning models such as FNNs and XGBoost models.   

\subsection{Feedforward Neural Networks}
\label{sec:fnns}
\noindent
Artificial neural networks (ANNs) are widely applied in various fields, e.g., visual recognition \citep{yan2015hd}, machine translation \citep{devlin2014fast}, and healthcare \citep{han2020keratinocytic}, due to their non-linear modeling capabilities. Although a variety of specialized ANN frameworks have been introduced, such as convolutional neural networks \citep{lecun1995convolutional}, long short-term memory networks \citep{hochreiter1997long}, and transformer networks \citep{vaswani2017attention}, in this article we focus on the most parsimonious class of ANN models, i.e., FNNs. As pointed out by \citet{christensen2022amachine}, FNNs are able to outperform various other data-driven models when it comes to predicting daily realized variance, while offering a favorable computational efficiency in contrast to recurrent neural networks such as long short-term memory networks. A detailed overview of FNNs is provided in \ref{sec:appendix_fnns}.

For forecasting volatility of new issues or spin-offs, we consider a three-hidden-layer FNN. The first hidden layer consists of eight, the second layer of four, and the third layer of two artificial neurons, as suggested by \citet{christensen2022amachine}. Each hidden layer contains ReLU activation functions \citep{agarap2018deep} defined by
\begin{equation}
f(x) = \max(0, x).
\end{equation} 
The resulting FNN architecture is trained based on $Q_{std}$ (FNN-STD) and $Q_{ext}$ (FNN-EXT) via backpropagation combined with ADAM optimization to minimize a mean squared error loss function. 

\subsection{Extreme Gradient Boosting}
\label{sec:xgboost}
\noindent
Gradient boosting, as introduced by \citet{freund1997decision} and \citet{friedman2001greedy}, is characterized by the sequential training of an ensemble of weak learners, where each weak learner, typically a tree learner, is estimated to correct the residuals of the previous one. This idea was later extended by \citet{chen2016xgboost} in the so-called XGBoost framework, which includes additional regularization and more precise weak learner estimation using second-order Taylor expansion. We briefly outline the architecture and estimation of XGBoost models in \ref{sec:appendix_xgboost}. 

In our empirical application, we estimate XGBoost models based on the mean squared error loss function with 40 boosting iterations, i.e., regression trees, and a maximum tree depth of five, to predict future realized variance. The considered XGBoost models are, like the previously introduced forecasting methods, estimated and evaluated based on the standard HAR predictors in $Q_{std}$ (XGB-STD) as well as on the extended predictor set $Q_{ext}$ (XGB-EXT). 

All of the presented models are combined with our transfer learning approach introduced in Section \ref{sec:da_rv}.
 
\section{Setup of the Empirical Study}
\label{sec:experimental_setup}
\noindent
This section presents the partitioning of the data sets into training and evaluation sets and introduces three different forecasting approaches for predicting the volatility of new issues and spin-offs. Furthermore, we detail the estimation of the forecasting models considered, as well as the evaluation metrics and statistical tests used to examine the validity of our results. 

\subsection{Specification of Training and Evaluation Data Sets}
\label{sec:data_preparation}
\noindent
We trim a copy of each new issue/spin-off data set after the first 150, 250, 350, 450, and the first 550 trading days. The last 100 trading days of each of the five trimmed data sets are designated as the out-of-sample evaluation set (rolling test set), in which all realized variance forecasting models are re-estimated every 5 trading days. The preceding observations are used exclusively as training data. To prevent information spillover from future source data observations, we align the end date of each source data set with that of the corresponding new issue/spin-off training data by trimming it accordingly. 

\begin{figure}[t]
    \centering
    \includegraphics[scale=0.6]{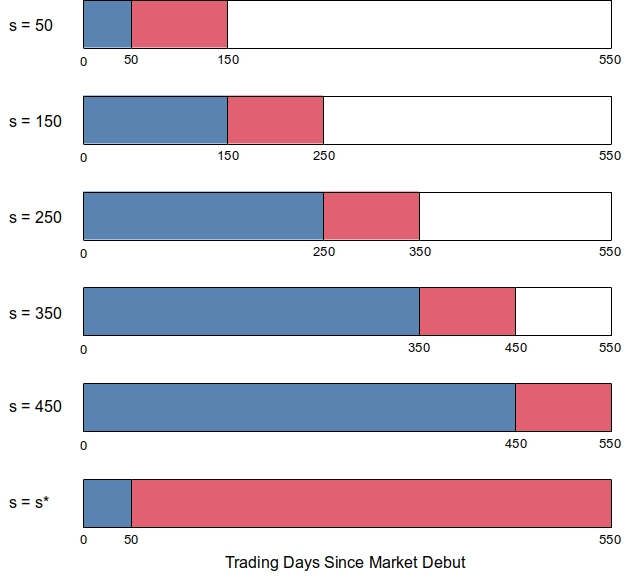} \\
    \caption{Training and evaluation data sets for new issues and spin-offs across different sample periods $s$. Blue intervals represent exclusive training periods, while red intervals indicate evaluation periods during which all realized variance forecasting models are re-estimated every 5 trading days. An overview of the training and evaluation periods used in our forecast evaluation starting immediately after the first trading day is provided in Figure \ref{fig:short_train_eval_schema}.}
    \label{fig:train_eval_schema}
\end{figure}

Based on this setup, we assess the 1-day-ahead realized variance forecasting performance over the entire evaluation set, i.e., over the observations 51 to 550, as well as over the individual 100-day subperiods following the first 50, 150, 250, 350, and 450 trading days of the respective new issue or spin-off. We refer to these sample periods as $s$, where $s=s^{*}$ represents the entire 500-day evaluation period, $s=50$ represents the 100-day evaluation period following the initial 50 trading days, $s=150$ represents the 100-day evaluation period following the initial 150 trading days, and so forth. The division of training and evaluation periods for each sample period is illustrated in Figure~\ref{fig:train_eval_schema}. It is important to note that, in the target training data sets, the actual number of observations is smaller than the number of available past trading days, due to the inclusion of predictors that are constructed using lagged data, such as weekly or monthly realized variance components. Recall that the monthly realized variance component at time period $t$ is based on the realized variance of day $t$ and the previous 21 daily realized variance observations. Consequently, after, e.g., 50 trading days, the target training set for a new issue/spin-off consists of only 29 data points. 

In addition to evaluating the forecasting performance subsequent to the initial 50 trading days, we also consider the extreme case, where forecasts are computed directly after the first trading day of the new issue or spin-off using correspondingly reduced predictor sets. The approach is described in detail in Section~\ref{sec:forecast_immediate_vici}.  

Furthermore, note that the input features for all FNN models are normalized, whereas scaling is omitted for the HAR and XGBoost models, as these models are invariant to scaled input features. 

\subsection{Forecasting Approaches}
\label{sec:forecasting_approaches}
\noindent
We consider three forecasting approaches that differ in terms of the training data set used to estimate the various realized variance forecasting models. In the first approach, all forecasting models are estimated exclusively based on the individual target training data of each new issue/spin-off. We refer to these models as "target only" (TO) models. In the second approach, the forecasting models are trained on the pooled target training data consisting of the training data of the individual new issue/spin-off and of the entire source data set. The training data sets of these models maintain the temporal alignment between the target and source asset data series as discussed in the previous section. We refer to models estimated in this manner as "naive pooling" (NP) models. In the third approach, the realized variance forecasting models are estimated based on the pooled target training data consisting of the training data of the individual new issue/spin-off and of all source subsequences selected by our proposed transfer learning method discussed in Section~\ref{sec:da_rv}. Recall that the number of selected source subsequences is determined by the threshold parameter $\epsilon$. To evaluate its impact, we analyze different inclusion rates (25th, 50th, and 75th percentile) for $\epsilon$. This allows us to assess how the number of selected subsequences influences the precision of realized variance forecasts. The resulting multi-source transfer learning (MTL) models are hereafter referred to as MTL-25, MTL-50, and MTL-75 models. Finally, we also consider a naive forecast (NF), which is defined as the previous day's realized variance of the target asset.

\subsection{Model Estimation}
\label{sec:model_estimation}
\noindent
The estimation of the predictive models under consideration is conducted individually for each target asset, predictor set, and training data set approach based on the specific estimation method pertinent to each model class. Specifically, the HAR models are estimated using ordinary least squares (OLS) regression, the XGBoost models are trained using Newton boosting, and the FNN models are optimized using the ADAM optimization algorithm in conjunction with the backpropagation method. 

\begin{figure}[t]
    \centering
    \includegraphics[scale=0.55]{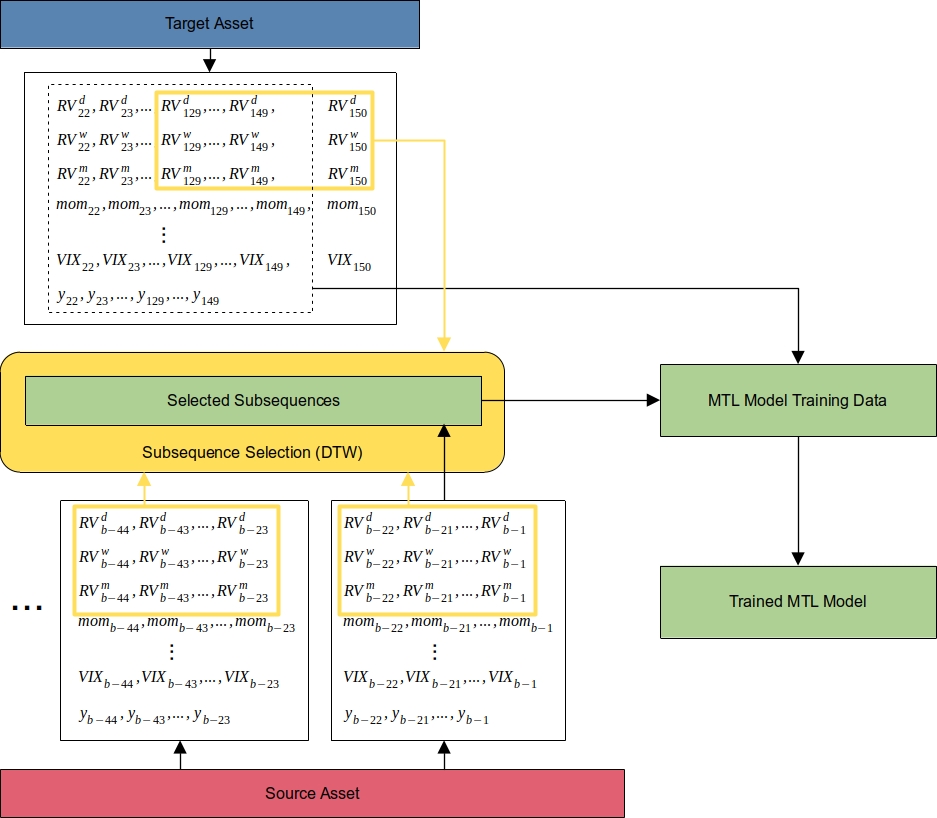} \\
    \caption{Example MTL model subsequence selection process after the 150th trading day of a new issue/spin-off. Depicted is the selection of subsequences from a single source asset for a subsequence size of $m=22$. The date of the source data observation $b$ corresponds to the 150th trading day of the new issue/spin-off. The DTW distance is computed based on the realized variance components. The number of selected subsequences is determined by the threshold parameter $\epsilon$; in the depicted example, only the most recent source subsequence is selected. This process is repeated at each estimation step for the progressively extended target and source series while maintaining temporal alignment.}
    \label{fig:mtl_training_schema}
\end{figure}

At the beginning of each 100-day sample period, the TO models are estimated based on all available target observations, whereas the NP model training data sets additionally include all source data observations up to the first day of the 100-day sample period. The MTL models are estimated based on the target data and selected source data subsequences. An illustration of the MTL subsequence selection process is provided in Figure \ref{fig:mtl_training_schema}. The source subsequences are selected based on their similarity to the input feature values of the target training data set over the most recent month ($m=22$), defined as the last 22 trading days. To determine the similarity between the target observations and the source subsequences, we calculate the DTW distance based on the lagged volatility predictors, i.e., $RV_{d}$, $RV_{w}$, $RV_{m}$. Finally, the selected source subsequences are combined with the entire corresponding target training data set. 

We then compute 1-day-ahead realized variance forecasts, where each model is re-estimated after 5 consecutive trading days. Consequently, the training data set of the TO models increases by five observations at each re-estimation point. Similarly, the combined training data set of the NP models is expanded by five observations from each source asset and by five observations from the target asset. Subsequences for MTL models are selected based on the current last $m=22$ target input feature values from the re-estimation point. This approach inevitably means that the data sets of MTL models are also incrementally expanded and continually updated based on the latest target input feature values at each estimation point.  

For the FNN models, the training is limited to a maximum of 500 epochs at the first estimation point of each sample period, while an early termination mechanism stops the training prematurely if the mean squared error on a 10\% validation set does not improve for 100 epochs. The resulting epoch count is then applied consistently to all subsequent re-estimation points of the same sample period. Unlike NP FNNs, which are trained with a batch size of 1024, and MTL FNNs which use a batch size of 512, TO FNNs require a significantly reduced batch size of four. This adjustment is attributed to the limited volume of target training data available for new issues/spin-offs. Table~\ref{tab:hyperparams} presents the adopted hyperparameter values of the FNN and XGBoost models.

Apart from optimizing the epoch count of the FNNs, we do not perform any other hyperparameter optimizations. The reasons for this are two-fold. First, the target data is already sparse. Thus, additional splitting to generate validation data would result in extremely small validation sets with questionable robustness regarding validation set errors. Second, the proposed rolling window forecasting and re-estimation methodology is computationally expensive. Additional hyperparameter tuning would significantly increase the computational time. However, XGBoost models and FNNs naturally exhibit a large number of potential hyperparameters. Therefore, it can be expected that these model classes would benefit the most from potential hyperparameter optimizations compared to HAR models. 

\begin{table}[t]
\centering
\caption{Hyperparameters of the XGBoost and FNN models.}
\scalebox{0.7}{
\begin{tabular}[t]{ll|ll}
\hline
\hline
\textbf{XGBoost Hyperparameters} & \textbf{Value} & \textbf{FNN Hyperparameters} & \textbf{Value} \\ \hline
Num. of Tree Learners & 40 & Learning Rate & 0.001 \\
Max. Tree Depth & 5 & Num. of Hidden Layers & 3 \\
Shrinkage (Learning Rate) & 0.1  & Units per Hidden Layer & [8,4,2]\\
Num. of Leaves Regularization ($\gamma$) & 0.1 & Batch Size & 4 / 512 / 1024\\
L1 Regularization & 0 &Hidden Layers Act. Functions & ReLU \\
L2 Regularization & 1 & Opt. Algorithm & ADAM \\
Subsample Ratio per Tree & 0.75 &&\\
Min. Num. of Instances per Leaf & 1&&\\
\hline
\hline
\end{tabular}
}
\label{tab:hyperparams}
\end{table}

\subsection{Forecast Evaluation}
\label{sec:evaluation}
\noindent
Before evaluating the generated realized variance forecasts, we enforce a non-negativity constraint on the predicted values. In particular, we follow \citet{christensen2022amachine} and replace negative forecast values with the empirical minimum realized variance of the target training data set. Notably, negative forecast values are extremely rare in our application. The out-of-sample performance of the different models is evaluated based on the same metric as is used in the estimation of the FNN and XGBoost forecasting models, i.e., the mean squared error (MSE). In addition, we report the out-of-sample performance for the less outlier-sensitive mean absolute error (MAE). 

Based on these error metrics, we assess the predictive accuracy of the forecasting models in two ways. First, we analyze the performance of each model over the entire test period, i.e., for $s=s^{*}$, in a pairwise comparison. To this end we compute for each new issue/spin-off the MSE (MAE) of one model ($M_{i}$) relative to the MSE (MAE) of another model ($M_{j}$) and report the cross-sectional average over all new issues/spin-offs of these pairwise relative MSEs (MAEs). For each target asset, we also examine the significance of the differences in the forecasts of individual model pairs ($M_{i} , M_{j}$) by applying a one-sided Diebold-Mariano (DM) test at a 5\% significance level. Specifically, we report whether the null hypothesis of equal forecast performance between forecasting models $M_{i}$ and $M_{j}$, i.e. $H_{0}: \mbox{MSE}_{i} = \mbox{MSE}_{j}$ ($H_{0}: \mbox{MAE}_{i} = \mbox{MAE}_{j}$), is rejected in favor of the alternative hypothesis, $H_{1}: \mbox{MSE}_{i} < \mbox{MSE}_{j}$ ($H_{1}: \mbox{MAE}_{i} < \mbox{MAE}_{j}$), for more than 50\% of the target assets considered. 

In a second forecast evaluation, we investigate the forecast accuracy for the individual 100-day sample periods. As this results in large volumes of forecast evaluation measurements, we only report the cross-sectional average of the MSE (MAE) of each model relative to the MSE (MAE) of the NF. 

Finally, for each sample period and new issue/spin-off, we construct a model confidence set (MCS; \citet{hansen2011model}) using a 95\% confidence level and 5000 bootstrap replications, which contains the most accurate forecasting models for the MSE (MAE) metrics. We enumerate the instances in which a particular forecasting model is included in the MCS for a new issue/spin-off during a specific 100-day sample period, and present the total number of occurrences for individual 100-day sample periods as well as the cumulative occurrences across all 100-day sample periods. In addition, we generate MCSs for the entire test period $s=s^{*}$ and report the total number of occurrences of individual forecasting models across all new issues and spin-offs.

\section{Results}
\label{sec:results}
\noindent
This section presents our empirical results. We first focus on evaluating the forecasting performance of our multi-source transfer learning approach compared to forecasts that are exclusively based on the target data (TO) or based on the entire source and target data (NP). We then discuss in more detail the properties of the source subsequences selected by our transfer learning approach. Finally, we assess the accuracy of our forecasting approaches under the extreme scenario of forecasting immediately one day after the distribution date of each new issue/spin-off. 

\subsection{Realized Variance Forecasting for New Issues and Spin-Offs}
\label{sec:forecasting_performance}
\noindent
The evaluation results of the realized variance forecasts, conducted through a pairwise comparison of all considered models across all new issues and spin-offs for the entire test period ($s=s^{*}$), are presented in Tables~\ref{tab:forecasting_results_mse} and \ref{tab:forecasting_results_mae} in \ref{sec:app_forecasting_results} for MSE and MAE metrics, respectively. This analysis shows significant differences in the performance of the various models. In particular, the MTL-75 XGB-EXT model stands out by achieving cross-sectional MSE and MAE ratios of one or less relative to all other models. As indicated by the rejection of the null hypothesis of the DM test for more than 50\% of all new issues/spin-offs, the MTL-75 XGB-EXT also significantly outperforms the NF and any TO model. Notably, according to the MAE, the MTL-50 XGB-EXT achieves an average performance that is on par with the MTL-75 XGB-EXT.

Focusing on the aggregated MSE (MAE) ratios of each model relative to the TO HAR-STD, i.e., the well-established standard HAR model, it becomes evident that the MTL-75 XGB-EXT leads to the greatest improvement in forecast accuracy compared to the TO HAR-STD. In fact, the MSE improves by 15.4\%. Similarly, the MTL-50 XGB-EXT and MTL-75 XGB-EXT yield an improvement of 14.5\% in the MAE. When comparing each model with the NF, the MTL-75 FNN-EXT reduces the MSE by 33.2\%, while the MTL-50 XGB-EXT and MTL-75 XGB-EXT provide a 17.4\% reduction in the average MAE relative to the NF.

This strong predictive performance of our multi-source transfer learning approach is not limited to the aforementioned models, but generalizes to all MTL-based HAR, FNN, and XGBoost models. These models, whether using the standard HAR predictors in $Q_{std}$ or the extended predictor set in $Q_{ext}$, consistently surpass the NF and all TO models by exhibiting aggregated MSE and MAE ratios below one when evaluated against each other. In addition, all MTL models demonstrate significantly higher forecast accuracy than the NF in terms of MSE and MAE for more than 50\% of all new issues/spin-offs according to the results of the DM tests. The same holds true for the majority of MTL STD and MTL EXT models for MAE, and MTL EXT models for MSE, when compared to individual TO models. 

A comparison of the MTL models with the NP models reveals that each of the MTL EXT models, as well as the majority of the MTL STD models, provides more accurate forecasts on average in terms of MSE and MAE than their respective NP counterparts. However, in most one-to-one comparisons between the MTL and NP models, the null hypothesis of the DM tests cannot be rejected for more than 50\% of all new issues/spin-offs. This indicates that MTL models lead to a significant improvement in realized variance forecast accuracy compared to models trained on the entire target and source data set, but not necessarily for every new issue/spin-off.

The analysis of the MTL STD and MTL EXT models shows that, in most cases, each MTL EXT model exhibits a cross-sectional relative MSE and MAE of less than one when compared to the corresponding MTL STD model. In particular, the MTL XGB-EXT models achieve improvements in the average MSE of up to 12\% over the MTL XGB-STD models. For the other MTL models, the improvements of using the extended set of predictors rather than the standard one are less pronounced, often amounting to only a few percentage points. 

Among the MTL EXT models, the MTL XGB-EXT models generally demonstrate higher prediction accuracy compared to MTL HAR-EXT and MTL FNN-EXT models in terms of MSE and MAE, followed by MTL FNN-EXT models, which outperform the MTL HAR-EXT models in the majority of cases. The reduction in MSE (MAE) of the respective superior model is mostly within a range of up to 5\%.

The results of the pairwise comparison based on the entire test sample period show that our multi-source transfer-learning approach yields superior forecasts in comparison to models trained exclusively on the target data or the naively pooled data, which consists of the entire source and target data. This suggests that instance selection is advantageous for forecasting the realized variance of new issues and spin-offs. In order to gain further insights into the relevance of the length of the available target data history, we now turn to an analysis of the individual test sample periods.

\begin{table}[t!]
\centering
\caption{MSEs and MAEs of each forecasting model relative to the MSE (MAE) of the NF averaged over all new issues and spin-offs considered. The MSE and MAE error metrics are obtained for 100 rolling 1-day-ahead forecasts based on $s=50$, $150$, $250$, $350$, and $450$ trading days after the distribution day of the new issue/spin-off. Blue values indicate, for each sample period ($s$) and error metric, the best-performing model within each models class, i.e., HAR, FNN, and XGB. The MSE (MAE) criteria of the overall best-performing model are highlighted in bold blue.}
\scalebox{0.59}{
\begin{tabular}{p{4.0cm}p{1.75cm}p{1.75cm}p{1.75cm}p{1.75cm}p{1.75cm}p{1.75cm}p{1.75cm}p{1.75cm}p{1.75cm}p{1.75cm}}
\hline
\hline
\multicolumn{1}{c}{Model} & \multicolumn{2}{c}{s = 50} & \multicolumn{2}{c}{s = 150} & \multicolumn{2}{c}{s = 250} & \multicolumn{2}{c}{s = 350} & \multicolumn{2}{c}{s = 450}\\
& MSE & MAE & MSE & MAE &  MSE & MAE &  MSE & MAE &  MSE & MAE\\ 
\hline
TO HAR-STD & 0.747 & 1.002 & 0.810 & 0.991 & 0.792 & 0.968 & 0.689 & 0.901 & 0.759 & 0.993\\
TO HAR-EXT & 1660.889 & 4.162 & 1.220 & 1.066 & 1.443 & 1.297 & 0.868 & 1.020 & 1.081 & 1.142\\
\hdashline
NP HAR-STD & 0.719 & 0.906 & 0.733 & 0.911 & 0.769 & 0.933 & 0.708 & 0.915 & 0.751 & 0.939\\
NP HAR-EXT & 0.664 & 0.915 & 0.780 & 0.957 & 0.908 & 0.979 & 0.741 & 0.964 & 1.037 & 1.026\\
\hdashline
MTL-25 HAR-STD & 0.663 & 0.822 & 0.720 & 0.844 & 0.731 & 0.851 & 0.664 & 0.822 & 0.718 & 0.869\\
MTL-50 HAR-STD & 0.660 & 0.816 & 0.713 & \textcolor{blue}{0.838} & 0.726 & 0.843 & 0.665 & 0.821 & 0.713 & 0.861\\
MTL-75 HAR-STD & 0.660 & 0.804 & 0.860 & 0.854 & 0.919 & 0.915 & 0.624 & 0.806 & 0.776 & 0.855\\
\hdashline
MTL-25 HAR-EXT & 0.581 & 0.784 & 0.709 & 0.853 & 0.689 & 0.831 & 0.616 & 0.797 & 0.697 & 0.840\\
MTL-50 HAR-EXT & 0.577 & \textcolor{blue}{0.780} & 0.705 & 0.848 & \textbf{\textcolor{blue}{0.686}} & \textbf{\textcolor{blue}{0.825}} & 0.615 & \textcolor{blue}{0.795} & \textcolor{blue}{0.696} & \textcolor{blue}{0.836}\\
MTL-75 HAR-EXT & \textcolor{blue}{0.575} & 0.786 & \textcolor{blue}{0.699} & 0.844 & 0.694 & 0.831 & \textcolor{blue}{0.611} & 0.796 & 0.708 & 0.844\\
\hline 
TO FNN-STD & 0.809 & 1.056 & 0.864 & 1.063 & 0.879 & 1.033 & 0.687 & 0.917 & 0.791 & 1.022\\
TO FNN-EXT & 177.933 & 2.025 & 0.723 & 0.982 & 0.968 & 1.086 & 0.717 & 0.965 & 1.192 & 1.171\\
\hdashline
NP FNN-STD & 0.687 & 0.878 & 0.715 & 0.888 & 0.732 & 0.894 & 0.683 & 0.882 & 0.721 & 0.900\\
NP FNN-EXT & 0.751 & 0.965 & 0.890 & 1.043 & 1.158 & 1.098 & 1.027 & 1.132 & 1.205 & 1.128\\
\hdashline
MTL-25 FNN-STD & 0.663 & 0.828 & 0.724 & 0.852 & 0.732 & 0.853 & 0.666 & 0.825 & 0.725 & 0.875\\
MTL-50 FNN-STD & 0.660 & 0.818 & 0.717 & 0.843 & 0.727 & 0.845 & 0.665 & 0.822 & 0.715 & 0.863\\
MTL-75 FNN-STD & 0.658 & 0.819 & 0.712 & 0.841 & 0.724 & 0.846 & 0.663 & 0.824 & 0.710 & 0.865\\
\hdashline
MTL-25 FNN-EXT & 0.593 & 0.796 & 0.709 & 0.858 & 0.692 & 0.843 & 0.618 & 0.799 & 0.711 & 0.853\\
MTL-50 FNN-EXT & 0.583 & \textcolor{blue}{0.788} & 0.691 & 0.841 & \textcolor{blue}{0.688} & \textcolor{blue}{0.831} & 0.613 & \textcolor{blue}{0.796} & \textcolor{blue}{0.700} & \textcolor{blue}{0.840}\\
MTL-75 FNN-EXT & \textcolor{blue}{0.573} & 0.791 & \textbf{\textcolor{blue}{0.662}} & \textbf{\textcolor{blue}{0.828}} & 0.698 & 0.838 & \textcolor{blue}{0.608} & 0.800 & 0.708 & 0.844\\
\hline
TO XGB-STD & 1.155 & 1.163 & 1.149 & 1.104 & 1.581 & 1.086 & 0.894 & 1.024 & 0.980 & 1.052\\
TO XGB-EXT & 0.976 & 1.104 & 1.325 & 1.105 & 1.281 & 1.105 & 0.883 & 1.037 & 0.879 & 0.998\\
\hdashline
NP XGB-STD & 0.675 & 0.837 & 0.740 & 0.870 & 0.791 & 0.921 & 0.672 & 0.851 & 0.732 & 0.903\\
NP XGB-EXT & 0.613 & 0.820 & 0.702 & 0.847 & 0.797 & 0.902 & 0.678 & 0.838 & 0.745 & 0.895\\
\hdashline
MTL-25 XGB-STD & 0.659 & 0.803 & 0.771 & 0.863 & 1.445 & 0.910 & 0.669 & 0.816 & 0.744 & 0.852\\
MTL-50 XGB-STD & 0.658 & 0.799 & 0.754 & 0.854 & 0.791 & 0.870 & 0.664 & 0.817 & 0.724 & 0.850\\
MTL-75 XGB-STD & 0.652 & 0.798 & 0.746 & 0.864 & 0.768 & 0.878 & 0.659 & 0.821 & 0.723 & 0.853\\
\hdashline
MTL-25 XGB-EXT & 0.579 & 0.781 & 0.708 & 0.848 & 0.784 & 0.861 & 0.615 & 0.788 & 0.686 & \textbf{\textcolor{blue}{0.821}}\\
MTL-50 XGB-EXT & 0.573 & 0.773 & 0.689 & \textcolor{blue}{0.830} & 0.748 & \textcolor{blue}{0.851} & \textbf{\textcolor{blue}{0.595}} & \textbf{\textcolor{blue}{0.787}} & 0.681 & 0.823\\
MTL-75 XGB-EXT & \textbf{\textcolor{blue}{0.558}} & \textbf{\textcolor{blue}{0.768}} & \textcolor{blue}{0.676} & 0.831 & \textcolor{blue}{0.730} & 0.861 & 0.597 & 0.791 & \textbf{\textcolor{blue}{0.679}} & 0.827\\
\hline
\hline
\end{tabular}
}
\label{tab:forecasting_results_rel_nf}
\end{table}

Table~\ref{tab:forecasting_results_rel_nf} presents the MSE (MAE) of each model relative to that of the NF approach, averaged across all new issues and spin-offs, for the individual 100-day evaluation periods conditioned on different initial target data lengths (50, 150, 250, 350, and 450 days). We observe that the MTL models using the extended predictor set also dominate in each individual sample period. Every MTL model provides, on average, more accurate realized variance predictions than the NF in each sample period, considering both MSE and MAE. Moreover, it is always an MTL model that provides the greatest improvement in MSE and MAE over the NF in each sample period. In particular, in the first sample period, the MTL models yield improvements in forecast accuracy of up to 44.2\% in the average MSE and up to 23.2\% in the average MAE. The forecast improvements of the TO and NP models are less pronounced, with the NP models achieving greater improvements than their TO counterparts. However, in most cases, these improvements remain smaller than those obtained by the corresponding MTL models, indicating a strong forecasting performance of the MTL approach. Hence, multi-source transfer learning consistently provides superior forecasts, irrespective of the length of the target asset’s available historical data.

The higher average forecast accuracy of the transfer learning models, especially the MTL EXT models, is also reflected in the model confidence sets, as shown in Table~\ref{tab:mcs}. Specifically, for the combined sample periods ($s=s^{*}$), the MTL-75 XGB-EXT is included in three MCSs based on MSEs, while the MTL-50 XGB-EXT is part of three MCSs based on MAEs. Taking into account the aggregated occurrences in MCSs of individual sample periods, the MTL-75 XGB-EXT is contained overall in eight MCSs based on MSEs, surpassing all other models. The MTL-25 XGB-EXT is included in 11 MCSs in terms of MAEs, more often than any other model. In general, considering both MSE and MAE, we can observe that all MTL EXT models (i.e., MTL XGB-EXT, MTL FNN-EXT, and MTL HAR-EXT models) are most often included in the MCSs as compared to the TO and NP models.

\begin{table}[t!]
\centering
\caption{Frequency of inclusion of forecasting models in MCSs based on MSE and MAE of all considered new issues and spin-offs for individual sample periods. The frequency of inclusion of each model over all individual sample periods is aggregated in a separate column (Agg.). The column $s=s^{*}$ reports the inclusion of individual forecasting models in the MCSs of new issues/spin-offs for the entire test period. The model with the highest number of inclusions within each error metric and sample period ($s$), as well as in the aggregate count is highlighted in bold.}
\scalebox{0.635}{
\begin{tabular}{p{4.0cm}p{1.0cm}p{1.0cm}p{1.0cm}p{1.0cm}p{1.0cm}p{1.0cm}p{1.0cm}p{1.0cm}p{1.0cm}p{1.0cm}p{1.0cm}p{1.0cm}p{1.0cm}p{1.0cm}p{1.0cm}p{1.0cm}}
\hline
\hline
\multicolumn{1}{c}{Model} & \multicolumn{2}{c}{s = 50} & \multicolumn{2}{c}{s = 150} & \multicolumn{2}{c}{s = 250} & \multicolumn{2}{c}{s = 350} & \multicolumn{2}{c}{s = 450} & \multicolumn{2}{c}{Agg.} & \multicolumn{2}{c}{$\textnormal{s = s}^{*}$}\\
&  MSE & MAE & MSE & MAE &  MSE & MAE &  MSE & MAE & MSE & MAE & MSE & MAE & MSE & MAE\\
\hline 
NF & 0 & 0 & 0 & 0 & 0 & 0 & 0 & 0 & 0 & 0 & 0 & 0 & 0 & 0\\
\hdashline
TO HAR-STD  & 0 & 0 & 2 & 1 & 1 & 1 & 0 & 0 & 0 & 0 & 3 & 2 & 0 & 0\\
TO HAR-EXT & \textbf{3} & 0 & 0 & 0 & 0 & 0 & \textbf{3} & 0 & 0 & 0 & 6 & 0 & 0 & 0\\
\hdashline
NP HAR-STD & 0 & 0 & 0 & 1 & 0 & 0 & 0 & 0 & 0 & 0 & 0 & 1 & 0 & 0\\
NP HAR-EXT & 0 & 0 & 0 & 0 & 0 & 0 & 0 & 0 & 0 & 0 & 0 & 0 & 0 & 0\\
\hdashline
MTL-25 HAR-STD & 0 & 0 & 0 & \textbf{3} & 0 & 1 & 0 & 0 & 0 & 0 & 0 & 4 & 0 & 2\\
MTL-50 HAR-STD & 0 & 0 & 0 & \textbf{3} & 0 & 1 & 0 & 1 & 0 & 0 & 0 & 5 & 0 & 2 \\
MTL-75 HAR-STD & 0 & 0 & 1 & 1 & 0 & 0 & 0 & 1 & 1 & 1 & 2 & 3 & 0 & 0 \\
\hdashline
MTL-25 HAR-EXT & 2 & \textbf{3} & 0 & 2 & 0 & 0 & 2 & 1 & \textbf{1} & 0 & 5 & 6 & 0 & 2\\
MTL-50 HAR-EXT & 2 & 1 & 0 & 1 & 0 & \textbf{2} & 2 & 0 & \textbf{1} & \textbf{2} & 5 & 6 & 0 & 2\\
MTL-75 HAR-EXT & 1 & 2 & 0 & 2 & 1 & \textbf{2} & \textbf{3} & 1 & 0 & \textbf{2} & 5 & 9 & 1 & 2\\
\hline 
TO FNN-STD & 0 & 0 & 0 & 0 & 0 & 0 & 0 & 0 & \textbf{1} & 0 & 1 & 0 & 0 & 0 \\
TO FNN-EXT & 0 & 0 & \textbf{3} & 0 & 0 & 0 & 2 & 1 & \textbf{1} & 0 & 6 & 1 & 0 & 0\\
\hdashline
NP FNN-STD & 0 & 0 & 1 & 0 & 1 & 0 & 0 & 0 & 0 & 0 & 2 & 0 & 1 & 0 \\
NP FNN-EXT & 2 & 1 & 0 & 1 & 1 & 0 & 0 & 0 & 0 & 0 & 3 & 2 & 0 & 0 \\
\hdashline
MTL-25 FNN-STD & 0 & 0 & 0 & 2 & 0 & 0 & 0 & 0 & 0 & 1 & 0 & 3 & 0 & 0\\
MTL-50 FNN-STD & 0 & 0 & 0 & 2 & 0 & 0 & 0 & 0 & 0 & 1 & 0 & 3 & 0 & 1\\
MTL-75 FNN-STD & 0 & 0 & 0 & 1 & 0 & 0 & 0 & 1 & \textbf{1} & 1 & 1 & 3 & 0 & 0\\
\hdashline
MTL-25 FNN-EXT & 1 & 2 & 1 & \textbf{3} & \textbf{3} & 0 & 1 & 1 & 0 & 1 & 6 & 7 & 0 & 1 \\
MTL-50 FNN-EXT & 1 & 1 & 1 & 1 & 0 & 1 & 2 & 1 & 0 & 1 & 4 & 5 & 1 & 2 \\
MTL-75 FNN-EXT & 2 & 1 & \textbf{3} & \textbf{3} & 0 & 0 & 1 & 1 & 0 & 1 & 6 & 6 & 2 & 1\\
\hline
TO XGB-STD & 0 & 0 & 0 & 0 & 0 & 0 & 0 & 0 & 0 & 0 & 0 & 0 & 0 & 0 \\
TO XGB-EXT & 0 & 0 & 1 & 0 & 0 & 0 & 0 & 0 & \textbf{1} & 1 & 2 & 1 & 1 & 0\\
\hdashline
NP XGB-STD & 0 & 1 & 2 & 1 & 1 & 0 & 1 & 1 & \textbf{1} & 0 & 5 & 3 & 1 & 1 \\
NP XGB-EXT & 0 & 1 & 1 & 2 & 0 & 1 & 1 & 1 & \textbf{1} & 0 & 3 & 5 & 1 & 1\\
\hdashline
MTL-25 XGB-STD & 0 & 1 & 1 & 2 & 0 & 1 & 0 & 1 & \textbf{1} & 1 & 2 & 6 & 0 & 1\\
MTL-50 XGB-STD & 0 & \textbf{3} & 1 & 1 & 1 & 1 & 0 & 1 & \textbf{1} & 1 & 3 & 7 & 0 & 1 \\
MTL-75 XGB-STD & 1 & 1 & 1 & 0 & 0 & 0 & 0 & 0 & \textbf{1} & 1 & 3 & 2 & 0 & 0\\
\hdashline
MTL-25 XGB-EXT & 1 & 2 & 1 & 1 & 0 & \textbf{2} & \textbf{3} & \textbf{4} & \textbf{1} & \textbf{2} & 6 & \textbf{11} & 0 & 1\\
MTL-50 XGB-EXT & 1 & \textbf{3} & 1 & 2 & 0 & \textbf{2} & \textbf{3} & 1 & \textbf{1} & 1 & 6 & 9 & 1 & \textbf{3} \\
MTL-75 XGB-EXT & \textbf{3} & 1 & 1 & 1 & 1 & 0 & \textbf{3} & 1 & 0 & 0 & \textbf{8} & 3 & \textbf{3} & 2\\
\hline
\hline
\end{tabular}
}
\label{tab:mcs}
\end{table}

In the following, we assess how the number of selected subsequences, determined by the choice of $\epsilon$, influences the forecast accuracy of the MTL models by comparing the forecasting performance of MTL-25, MTL-50, and MTL-75 models within each model specification (i.e., HAR, FNN, and XGB) with the same set of predictors (i.e., STD or EXT). The results presented in Tables~\ref{tab:forecasting_results_mse} and \ref{tab:forecasting_results_mae} reveal that there are only minor differences in the forecasting performance, with variations being typically within a narrow range of less than 5\% for both the relative MSE and MAE. However, the relative MSE tends to decrease slightly as more source data is integrated, such as in the MTL-75 models. In a large number of cases, the smallest average MAE values are obtained, when utilizing 50\% of the available source data. These findings suggest that there are two types of source data observations. On the one hand, there is a relatively small fraction of source observations that significantly improve the volatility forecasts compared to models trained exclusively on the target data, and, on the other hand, there is a small set of observations that are detrimental to the prediction. Thus, the main benefit of transfer learning in this context is to identify the former and exclude the latter from the training data sets. As our results suggest, this can be achieved by exploiting the similarity between target and source observations.

After the detailed analysis and discussion of the performance of the MTL models, we now proceed to analyze the performance of the non-MTL models. Comparing the forecasting accuracy of the NP models in relation to the NF, see Tables~\ref{tab:forecasting_results_mse} and \ref{tab:forecasting_results_mae}, we find that all NP models show average MSE (MAE) ratios of less than one. At the same time, the majority of the NP models achieves a significant reduction in MSE and MAE compared to the NF for more than 50\% of the new issues/spin-offs according to the DM tests. Moreover, we observe that specifically the NP HAR and NP XGB models exhibit greater forecast accuracy on average than their TO counterparts when employing the same predictor set. For example, NP XGB-EXT reduces the MSE by 25.1\% in relation to TO XGB-EXT. On the other hand, the NP FNN-EXT performs worse than TO FNN-EXT with respect to MSE (MAE) ratios. Interestingly, the NP HAR-STD and NP FNN-STD models outperform their EXT counterparts when compared to each other, while the NP XGB-EXT exhibits small improvements in MSE and MAE, when compared to the NP XGB-STD models. Contrary to these observations, the NP HAR models are generally less frequently included in the model confidence sets of individual sample periods compared to the TO HAR models, see Table~\ref{tab:mcs}. Meanwhile, the NP XGB models are included more frequently in the model confidence sets of individual sample periods than the TO XGB models. In summary, models trained on both the target and entire source data, i.e., NP models, generally outperform both the NF models and their TO counterparts in terms of forecast accuracy, with the exception of NP FNN-EXT. We also note that less complex NP STD models, such as NP HAR-STD and NP FNN-STD, outperform their more complex EXT counterparts.

Concerning TO models, our results show that the TO HAR-STD model consistently outperforms the NF and all other TO models in terms of average MSE (MAE) ratios. Furthermore, we observe that the TO HAR-STD and the TO FNN-STD demonstrate superior performance compared to the NF and their TO counterparts utilizing the extended predictor sets. When considering the analysis of individual sample periods in Table~\ref{tab:forecasting_results_rel_nf}, this distinction appears to be most pronounced during the initial sample period, where the TO HAR-EXT and TO FNN-EXT models yield substantial MSEs and MAEs relative to the NF. These findings are consistent with the general understanding that parsimonious models tend to excel when trained on limited data as opposed to more complex models. Interestingly, this contrast is less pronounced when considering the TO XGB-EXT, which performs on par with the TO XGB-STD in terms of MSE and MAE. However, both the TO XGB-STD and TO XGB-EXT models simultaneously perform less favorably than the NF. Upon closer examination of the computed model confidence sets presented in Table~\ref{tab:mcs}, an intriguing pattern emerges. Although the TO HAR-STD and TO FNN-STD models show higher forecast accuracy for the MSE and MAE, their counterparts using the extended predictor sets are more frequently included in the model confidence sets of individual sample periods in terms of MSEs, even in the initial sample period. This indicates that the primary reason for the inferior overall performance of TO HAR-EXT and TO FNN-EXT models is largely due to substantial MSEs (MAEs) for one or a limited number of new issues/spin-offs. Consequently, this scenario can lead to instances where the average MSE (MAE) ratios of two forecasting models are greater than one, regardless of which of the two models serves as the reference model. For example, we observe this circumstance for the average MSE ratios of the TO XGB-STD and TO-HAR EXT models or the NP FNN-EXT and TO HAR-EXT models. 

In summary, our results highlight that complementary source data and transfer learning, in the form of NP and MTL models, leads to significant improvements in the forecast accuracy of both linear and non-linear realized variance forecasting models for new issues and spin-offs. Moreover, our findings demonstrate that our proposed multi-source transfer learning approach (MTL) outperforms models trained on the entire source data set via naive pooling (NP), target only models, and the naive forecast. Furthermore, MTL consistently provides superior forecasts, irrespective of the amount of available target data.  

\subsection{Characteristics of source data subsequences selected by MTL models}
\noindent
In order to gain a deeper understanding of the source subsequences selected by our proposed transfer learning approach, we conduct an additional analysis on the training data sets of the MTL-25, MTL-50, and MTL-75 models. This analysis examines the distribution of selected subsequences across the complementary source assets and their temporal distance from the forecast origin.  

\subsubsection{Distribution of selected source subsequences across complementary assets}
\noindent
The analysis of the origins of selected source subsequences, specifically the source assets from which these series were selected, is presented in Tables~\ref{tab:source_orign_25} - \ref{tab:source_orign_75} in \ref{apdx:subequence_selection_rates}. The tables report, for each target asset, the average selection rates of available subsequences from individual source assets with the average taken across all estimation steps and sample periods. Since the total number of selected subsequences varies depending on the choice of $\epsilon$, the results are presented accordingly. Specifically, Table~\ref{tab:source_orign_25} reports the respective results for MTL-25 models, Table~\ref{tab:source_orign_50} for MTL-50, and Table~\ref{tab:source_orign_75} for MTL-75.  

A closer examination of the selection rates reveals that certain sectors and specific source assets exhibit lower proportions of subsequences included in the training data sets of the MTL models, while others show higher selection rates. For instance, we observe that subsequences of the \emph{consumer staples} sector are percentage-wise less frequently selected for the training sets of certain MTL models, particularly the MTL-25, MTL-50, and MTL-75 models for the target assets TWTR and MRNA, and the MTL-50 models for NCLH, PSX, SYF, and CTVA. Consequently, this implies that the lagged volatility component subsequences of \emph{consumer staple} sector assets exhibit a notable dissimilarity from the volatility component series of the aforementioned target assets. In addition, smaller proportions of subsequences from the majority of \emph{utility} sector assets are selected for training data sets of TWTR and all MRNA MTL models.

Conversely, the training data sets for the TWTR MTL-25 and MTL-50 models exhibit an above-average inclusion rate of subsequences from assets in the \emph{communication services}, \emph{consumer discretionary}, and \emph{information technology} sector. This pattern of higher inclusion rates extends to the TWTR MTL-75 training data set, with respect to \emph{energy} sector assets. Similarly, the MRNA MTL-25, MTL-50, and MTL-75 training sets show a high selection rate of subsequences from the \emph{communication service}, \emph{consumer discretionary}, and \emph{information technology} sector assets. Furthermore, the MRNA target data tends to display a higher realized variance structure similarity with specific \emph{energy} sector assets, as indicated by larger proportions of selected subsequences from \emph{energy} sector assets in the MRNA MTL-25, MTL-50, and MTL-75 training data sets. A comparable trend is also evident in the PSX MTL-25 and MTL-75 training sets. 

A review of the inclusion rates within the respective GICS sectors of each target asset reveals generally moderate selection rates. Notable exceptions are the high inclusion rates of the TWTR MTL-25, TWTR MTL-50, PSX MTL-25 and INVH MTL-50 training data sets relative to their respective sectors. 

Aside from the observations discussed, we note that certain source assets demonstrate lower selection rates across a large number of target MTL model training sets. This is evident in cases such as PG for all MTL models; JNJ for MTL-25 and MTL-50 models; TTWO, TSLA, and DXCM for MTL-50 and MTL-75 models; and FTNT and QRVO for MTL-75 models.

Interestingly, our analysis suggests that selecting source subsequences based on the similarity to recent target observations at the forecast origin does not necessarily result in high selection rates for assets belonging to the same sector as the target asset. Although the corresponding sector seems to be relevant, the assets from which subsequences are selected are spread across all sectors considered. However, for specific target assets, certain source assets and sometimes even entire source asset sectors, appear to exhibit dissimilar patterns in their lagged volatility predictors.  

\subsubsection{Temporal distance between source subsequence extraction points and forecast origins}
\noindent
In addition to analyzing the selection rates of individual complementary assets, we examine the temporal discrepancy between the forecast origins and the points in time from which the selected subsequences are extracted. Specifically, for a given target asset and a given forecast origin, we analyze how many selected subsequences have a start date within the first 100 observations prior to the forecast origin, how many start dates fall within the interval 101 to 200 observations prior to the forecast origin, and so forth. We set each of these counts in relation to the number of source subsequences available in the respective 100-day interval, i.e., all the source subsequences with a start date that falls in that interval. We repeat this for each forecast origin and compute the average of the corresponding ratios for each 100-day interval. The results are illustrated in Figures~\ref{fig:time_diff_25} - \ref{fig:time_diff_75} of \ref{appdx:temp_distances}. It is important to note that the results for the group of subsequences with the greatest temporal distance from the forecast origin are primarily driven by the ratio of selected subsequences in the final re-estimation steps. Consequently, the average is computed from fewer observations, making it more sensitive to variations in the relative quantity of selected subsequences. Therefore, the significance of this class, i.e., the final bars of each asset chart in Figures~\ref{fig:time_diff_25} - \ref{fig:time_diff_75}, should be interpreted with caution, and we refrain from including these groups in our subsequent discussion.

When assessing the selection rates across different temporal distances between the forecast origins and selected subsequences, distinct patterns emerge for different target assets. For example, the MTL models for TWTR and MRNA show a tendency to select subsequences from more distant points in the past relative to the forecast origin, although MRNA also demonstrates high inclusion rates for subsequences in temporal proximity to the forecast origin. In contrast, MTL models for LW and INVH display a preference primarily for subsequences closer to the forecast origin. Meanwhile, MTL models for NCLH, PSX, SYF, CARR, DXC, and CTVA exhibit a relatively stable selection rate for subsequences, irrespective of their temporal proximity to the forecast moment. These findings suggest that target observations close to the forecast origin show high similarities not only to source subsequences from the same time period but also to those that are significantly further back in time. Consequently, this indicates that these far-back source subsequences also enhance the realized variance forecasts for new issues and spin-offs.

\subsection{Forecast evaluation in the immediate vicinity of the first trading day}
\label{sec:forecast_immediate_vici}
\begin{figure}[t]
    \centering
    \includegraphics[scale=0.6]{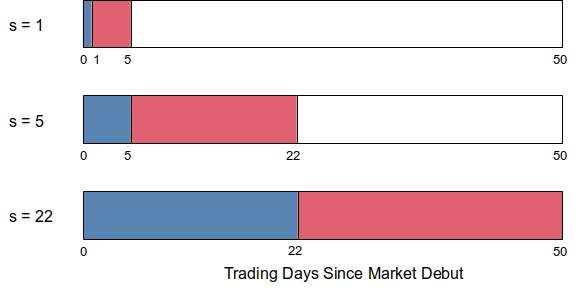} \\
    \caption{Training and evaluation data sets for new issues and spin-offs immediately following their first trading day across different sample periods ($s=1$, $s=5$, and $s=22$). Blue intervals represent exclusive training periods, while red intervals indicate evaluation periods during which all realized variance forecasting models are re-estimated daily in this setting.}
    \label{fig:short_train_eval_schema}
\end{figure}

\noindent
In the forecast assessment presented in the previous subsections, we initiated the forecast analysis 50 days after the first trading day of an asset. However, recognizing that investors may have an interest in volatility forecasts from the very beginning of an asset's life cycle, we conduct a second forecast evaluation to assess the capabilities of of our multi-source transfer learning approach when faced with extreme target data scarcity. In particular, we evaluate the performance of the introduced realized variance forecasting models for new issues and spin-offs, starting immediately after their first trading day. To this end, we report the MSE (MAE) of each model relative to the MSE (MAE) of the NF as a cross-sectional average starting from the first day after the distribution date up to 50 days after the initial trading day. We segment this time period into three sample periods, as illustrated in Figure~ \ref{fig:short_train_eval_schema}. The first sample period ($s=1$) covers the initial 4 trading days following the first trading day. The second sample period ($s=5$) extends from the 6th to the 22nd trading day, while the third sample period ($s=22$) spans from the 23rd to the 50th trading day post-distribution. The evaluation set sizes for the specified periods are 4, 17, and 28 days, respectively. The forecasts are evaluated on a daily basis using a rolling approach, where each forecast is generated for the subsequent trading day.

Consequently, for the forecasts of the first two sample periods, i.e., after 1 and 5 days, we must exclude predictors that represent lagged aggregates or differenced values such as the weekly and monthly volatility components or the $MOM$ and $DV$ predictors. We introduce a corresponding notation: the forecasting models of Section~\ref{sec:forecasting_performance}, that use the entire set of predictors in $Q_{std}$ and $Q_{ext}$, are denoted by an added prefix of "22-", for example, 22-TO HAR-STD. Models without the monthly volatility component and $MOM$ predictor bear a prefix of "5-", while those lacking the $RV_{m}$, $RV_{w}$, $MOM$, $DV$, $US3M$, and $HSI$ predictors are prefixed with "1-". The application of the respective models to the sample periods commences as soon as a sufficient number of target observations is available. Specifically, the 1-models are utilized immediately following the first trading day and are applied to all subsequent sample periods ($s=1$ / $s=5$ / $s=22$). Similarly, the 5-models are initiated after 5 trading days, and then applied consistently throughout the following sample periods ($s=5$ / $s=22$), while the 22-models are applied only to the last sample period, i.e., $s=22$, starting after 22 trading days.

In this forecasting setup, the models are re-estimated daily to address the extremely limited quantity of target observations. The length of source subsequences for MTL models are dynamically matched to the length of the available target data, up to a maximum of 22 days, at each re-estimation point, while the DTW distance between the target and source subsequences is determined based only on the available lagged volatility predictors of the respective predictor sets. In addition, the TO FNNs are trained with a batch size of one and, if only one target observation is available, with a fixed epoch count of 500.

We acknowledge that the inclusion of TO models may be highly controversial due to their limited training data. Nevertheless, we include TO models in our analysis for the purpose of comparison and to maintain consistency.

\subsubsection{Realized Variance Forecasting up to 50 days after the first trading day}
\label{sec:forecast_performance_up_to_50_days}
\noindent
Tables~\ref{tab:after_first_day_1} - \ref{tab:after_first_day_22} in \ref{appdx:vic_of_first_day} report the MSEs (MAEs) of the forecasting models relative to the MSE (MAE) of the NF. Table~\ref{tab:after_first_day_1} reports the results for the 1-models, Table~\ref{tab:after_first_day_5} for the 5-models, and Table~\ref{tab:after_first_day_22} for the 22-models. It becomes evident that the majority of the models that incorporate additional source data in their training data sets, i.e., NP and MTL models, outperform the NF on average in each sample period. This observation holds true for the 1-models, which use only the most reduced predictor set, and those that become available after 5 and 22 days, namely the 5- and 22-models. In addition, our results indicate that MTL models provide an advantage over their respective NP counterparts.

Conversely, the 1-TO, 5-TO, and 22-TO models, with the notable exception of the TO XGB models, in most cases exhibit considerably higher MSEs and MAEs relative to the NF, especially in their initial sample periods. This is not entirely surprising, considering that the 1-TO models are estimated on a maximum of just 49 data points at the last forecast origin, and the 5-TO and 22-TO models are trained on only up to 45 and 28 observations, respectively.

Interestingly, in instances where the forecasting periods of 1-, 5-, and 22-models intersect, there is a noticeable trend for most NP and MTL models to show modest decreases in the relative MSE and MAE, favoring the respective more complex model. This observation suggests a viable approach for NP and MTL models to transition to their 5- and 22-counterparts (5-NP, 5-MTL, 22-NP, and 22-MTL) as soon as a sufficient amount of data becomes available, specifically after 5 and 22 trading days, respectively. To further investigate whether adopting more complex models, which are trained on fewer target observations, is advantageous over simpler models that utilize a larger data set of target observations, we examine the effectiveness of different strategies for transitioning from one predictor set to another in the following subsection.

\subsubsection{Predictor Set Transition Strategies}
\noindent
In total, we assess three different strategies to transition between the predictor sets of all NP and MTL forecasting models after 5 and 22 days from the distribution date of the respective target assets. The first strategy, herein referred to as "Str. 1-1-1", involves uniformly using the predictor sets of the 1-models that lack $RV_{m}$, $RV_{w}$, $MOM$, $DV$, $US3M$, and $HSI$ predictors across all three sample periods. The second approach, denoted as "Str. 1-5-5", entails transitioning from the predictors of 1-models to those of 5-models for the second and third sample periods. The third strategy, "Str. 1-5-22", advocates for a prompt transition to the more complex predictor sets as soon as adequate data becomes available, specifically switching to 5-model predictors after 5 days and to 22-model predictors after 22 days. This probably represents the most intuitive approach to exploit the increasing availability of data over time, allowing more predictors to be integrated progressively. 

To analyze the forecasting performance of these transition strategies, we employ a pairwise comparison approach similar to that in Section~\ref{sec:forecast_performance_up_to_50_days}. Specifically, for each new issue and spin-off, we compute the MSE (MAE) of one strategy relative to another for both the MTL and NP models. Table~\ref{tab:predictor_transitioning} reports the average of these MSE (MAE) ratios, where, for clarity, the averages are computed across the cross-section as well as across all MTL and NP models. The findings reveal a distinct performance hierarchy, with Str. 1-5-22 outperforming Str. 1-5-5, which in turn exceeds the performance of Str. 1-1-1. While the improvements in relative MSE and MAE for Str. 1-5-22 over Str. 1-5-5 are modest, they nonetheless indicate a benefit from promptly adopting more complex predictor sets. Notably, this pattern is also observed individually when separating the results for the MTL and NP models.\footnote{The corresponding results are available from the authors upon request.} 

\begin{table}[t]
\caption{Performance of transition strategies. Reported are the averages of the MSEs (MAEs) of the column strategy relative to the MSEs (MAEs) of the row strategy. MSEs (MAEs) are based on 49 one-day-ahead realized variance forecasts following the first trading day of a new issue/spin-off. The averages are computed across all new issues and spin-offs, as well as across all NP and MTL models.}
\centering
\scalebox{0.8}{
\begin{tabular}[t]{p{2cm}|p{2cm}p{2cm}p{2cm}|p{2cm}p{2cm}p{2cm}}
\hline
\hline
& \multicolumn{3}{c}{relative MSE} |& \multicolumn{3}{c}{relative MAE}\\
\hline
Strategy &  Str. 1-1-1 & Str. 1-5-5 & Str. 1-5-22 & Str. 1-1-1 & Str. 1-5-5 & Str. 1-5-22 \\
\hline 
Str. 1-1-1 & 1.000 & 0.938 & 0.936 & 1.000 & 0.962 & 0.963 \\
Str. 1-5-5  & 1.104 & 1.000 & 0.996 & 1.067 & 1.000 & 1.001 \\
Str. 1-5-22 & 1.133 & 1.023 & 1.000 & 1.071 & 1.002 & 1.000 \\
\hline
\hline
\end{tabular}
}
\label{tab:predictor_transitioning}
\end{table}

Additionally, we apply all transition strategies to TO models and compare their forecasting performance against NP and MTL models using the best-performing NP/MTL transition strategy, i.e., Str. 1-5-22. Table~\ref{tab:strat_forecast_results} reports for each model the cross-sectional average of the MSE (MAE) relative to the NF. Our findings demonstrate that the MTL models consistently deliver the most accurate forecasts within each model class, i.e., HAR, FNN, and XGB. Regarding the MSE, the MTL-75 models demonstrate superior predictive performance. With respect to the MAE, no single MTL model consistently outperforms the others, as MTL-25, MTL-50, and MTL-75 each achieve the highest accuracy within a different model class. Notably, in contrast to the forecast assessment in Sec.~\ref{sec:forecasting_performance}, which begins 50 days after the first trading day of each new issue/spin-off, the STD predictor set MTL models outperform their EXT predictor set counterparts in terms of MSE. Additionally, when considering XGB models, the STD predictor set also provides superior forecast accuracy with respect to the MAE. 

\begin{table}[t]
\caption{Performance of transition strategies for TO models compared to the best-performing MTL and NP model strategy (Str. 1-5-22). Reported are the cross-sectional average MSEs (MAEs) for each forecasting model (HAR, FNN, XGB) under different transition strategies, predictor sets (STD, EXT), and forecasting approaches (TO, NP, MTL) relative to the NF. The MSEs (MAEs) are based on 49 one-day-ahead realized variance forecasts following the first trading day of a new issue/spin-off. The MSE (MAE) criteria of the best-performing model within each model class are marked in bold. The term "$>$ 99" is used in this table to represent values that are exceptionally large, exceeding 99.}
\centering
\scalebox{0.8}{
\begin{tabular}[t]{p{5cm}|p{1.5cm}p{1.5cm}p{1.5cm}|p{1.5cm}p{1.5cm}p{1.5cm}}
\hline
\hline
& \multicolumn{3}{c}{MSE relative to NF} |& \multicolumn{3}{c}{MAE relative to NF}\\
\hline
& HAR & FNN & XGB & HAR & FNN & XGB \\
\hline 
Str. 1-1-1 TO-STD & 1.274 & 1.381 & 1.013 & 1.083 & 1.200 & 1.083 \\
Str. 1-1-1 TO-EXT & 31.496 & 1.815 & 0.885 & 2.261 & 1.344 & 1.008 \\
\hdashline
Str. 1-5-5 TO-STD & 1.729 & 1.549 & 0.972 & 1.128 & 1.184 & 0.985 \\
Str. 1-5-5 TO-EXT & 7.027 & $>$ 99 & 0.911 & 1.931 & $>$ 99 & 0.958 \\
\hdashline
Str. 1-5-22 TO-STD & 13.022 & 1.884 & 0.933 & 1.526 & 1.384 & 0.943 \\
Str. 1-5-22 TO-EXT & 41.758 & $>$ 99 & 0.860 & 2.978 & $>$ 99 & 0.913 \\
\hline
Str. 1-5-22 NP-STD  & 0.754 & 0.737 & 0.765 & 0.896 & 0.883 & 0.875 \\
Str. 1-5-22 NP-EXT  & 0.815 & 0.881 & 0.771 & 0.902 & 0.941 & 0.872 \\
\hline
Str. 1-5-22 MTL-25-STD & 0.707 & 0.704 & 0.749 & 0.827 & 0.828 & 0.848 \\
Str. 1-5-22 MTL-25-EXT & 0.721 & 0.724 & 0.873 & 0.810 & \textbf{0.812} & 0.859 \\
\hdashline
Str. 1-5-22 MTL-50-STD & 0.702 & 0.702 & 0.761 & 0.825 & 0.828 & 0.848 \\
Str. 1-5-22 MTL-50-EXT & 0.716 & 0.720 & 0.845 & \textbf{0.809} & 0.817 & 0.860 \\
\hdashline
Str. 1-5-22 MTL-75-STD & \textbf{0.698} & \textbf{0.698} & \textbf{0.729} & 0.828 & 0.829 & \textbf{0.845} \\
Str. 1-5-22 MTL-75-EXT & 0.713 & 0.712 & 0.774 & 0.813 & 0.817 & 0.849 \\
\hline
\hline
\end{tabular}
}
\label{tab:strat_forecast_results}
\end{table}

To summarize, the examination of realized variance forecasts for new issues and spin-offs commencing on the first day after the distribution date underscores another significant finding: the integration of complementary source data and transfer learning enhances realized variance forecasts even immediately following the distribution date. Furthermore, our results suggest that transitioning to more complex models once a sufficient amount of data is available represents a viable option.

\section{Conclusion}
\label{sec:conclusion}
\noindent
In this paper, we considered the problem of forecasting the volatility of assets with limited data availability, such as new issues/spin-offs. We argued that complementary data from financial source assets with a comprehensive historical record can improve volatility forecasts for new issues and spin-offs. To this end, we proposed a model-agnostic transfer learning approach based on the DTW distance between recent target observations and generated source data subsequences. In order to compare our multi-source instance selection approach with methods that either rely solely on target asset data or target data combined with the entirety of the source data, we conducted a realized variance forecast assessment of 10 new issues and spin-offs. This forecast evaluation incorporated both linear and non-linear models, i.e., HAR models, XGBoost models, and FNNs. Moreover, we assessed all forecasting models using the standard HAR predictor set and an extended predictor set. Furthermore, we analyzed the characteristics of source subsequences selected by our transfer learning approach and presented a second forecast assessment in which we evaluated the forecasting precision of the proposed multi-source transfer learning approach in the immediate time period after the first trading day. In this way, we extend existing research by considering volatility forecasting for new issues and spin-offs, and the application of transfer learning methods to forecasting realized variance. 

Our results highlight that knowledge transfer from complementary data increases the predictive accuracy of all models considered. In particular, our proposed transfer learning method in combination with XGBoost models shows superior forecasting performance compared to any other model class or training data set approach. Moreover, we also observe a higher forecast accuracy for our instance selection approach in combination with HAR and FNN models compared to models trained on the target data alone or trained using a naive pooling approach. Hence, multi-source transfer learning improves the forecast performance irrespective of the adopted model class.

The analysis of the subsequences selected by our transfer learning approach reveals that the selected subsequences are neither limited to the GICS sector of the respective target asset nor to more recent source observations. Furthermore, we also observe a strong variability in the number of selected subsequences for different target assets. This suggests that it is useful to consider a wide range of different source assets with extensive historical data in order to improve the predictive accuracy of new issue/spin-off realized variance forecasting models. 

Furthermore, our empirical analysis shows that multi-source transfer learning improves the forecast accuracy even in the extreme, yet practically important, scenario in which volatility forecasts are required immediately after the first trading day of the new issue/spin-off. Our findings confirm the advantages of integrating source data and transfer learning methods into realized variance forecasting models. The majority of models that leverage source data provide superior forecasts compared to those trained solely on limited target data or a naive forecast. In addition, our results suggest that transitioning to more complex models - specifically, those incorporating weekly and monthly volatility components - is advantageous. The benefits of these models manifest immediately upon their availability.

Further research in the area of incorporating complementary data and transfer learning is warranted to explore the full potential of these methods in enhancing the precision of volatility forecasting models for new issues and spin-offs. This may involve an extensive hyperparameter optimization or the application of different forecasting models such as transformer networks. Moreover, our results motivate an assessment of how complementary source data can be used to improve volatility predictions for target assets that already have an extensive data history, i.e., multiple years of available data.

\clearpage

\appendix
\clearpage

\section{Dynamic Time Warping}
\label{sec:dtw}
\noindent
Consider a time series $T = (\textbf{t}_{1}, \textbf{t}_{2}, \dots, \textbf{t}_{n}, \dots, \textbf{t}_{N})$ with $N \in \mathbb{N}$ and $\textbf{t}_{n} \in \mathbb{R}^{p}$ and a time series $S = (\textbf{s}_{1}, \textbf{s}_{2}, \dots, \textbf{s}_{m}, \dots,  \textbf{s}_{M})$ with $M \in \mathbb{N}$ and $\textbf{s}_{m} \in \mathbb{R}^{p}$. In order to align both series, DTW constructs a local cost matrix $C \in \mathbb{R}^{MxN}$ defined as 
\begin{equation}
C(n,m) := c(\textbf{t}_{n}, \textbf{s}_{m}),
\end{equation}
where each cell represents the cost ($c$), i.e., the distance, typically the Euclidean distance, between two observations of $T$ and $S$. The optimal alignment between the two time series is identified by finding the minimum-cost warping path through the cost matrix. We define a warping path $O$ as a sequence of points $O=(o_{1}, o_{2}, \dots, o_{L})$ with  $o_{l} = (o_{i}, o_{j})$, $i \in  N, j \in M$, and $l \in L$, which adheres to the following constraints: 
\begin{itemize}
    \item Boundary condition: $o_{1} = (1,1)$ and $o_{L} = (N,M)$ 
    \item Monotonicity condition: $n_{1} \leq n_{2} \leq \dots \leq n_{L}$ and $m_{1} \leq m_{2} \leq \dots \leq m_{L}$ 
    \item Continuity condition: $o_{l+1} - o_{l} \in \{(1,1), (1,0), (0,1)\}$ 
\end{itemize}
While the boundary condition ensures that the warping path starts at the beginning of both sequences and ends at the end of both sequences, the monotonicity condition prevents the warping path from moving backward in time for either sequence. Additionally, the continuity condition defines the step size of the warping path to avoid overly large jumps in one sequence relative to the other. The total cost of a warping path with respect to the cost matrix is defined by
\begin{equation}
c_{O}= \sum_{l=1}^{L} c(\textbf{t}_{n_{l}}, \textbf{s}_{m_{l}}),
\label{eq:path_costs}
\end{equation}
where $(\textbf{t}_{n_{l}}, \textbf{s}_{m_{l}})$ is a matrix element which also represents the $l$-th element of the warping path $O$. As shown by \citet{bellman1959adaptive}, an optimal warping path, which minimizes Eq.~\eqref{eq:path_costs}, can be determined recursively by dynamic programming. The DTW distance $DTW(T, S)$ between $T$ and $S$ is then defined as the total cost of the optimal path $O^{*}$:
\begin{equation}
DTW(T,S) = c_{O^{*}} = \sum_{l=1}^{L} c(\textbf{t}_{n_{l}}, \textbf{s}_{m_{l}}). 
\end{equation}
\clearpage

\section{Source Data Set}
\label{sec:appendix_source_data}
\begin{table}[ht!]
\centering
\caption{Overview of source data set assets ordered by GICS sectors.}
\scalebox{0.57}{
\begin{tabular}{llll}
\hline \hline
\textbf{Ticker} & \textbf{Name} & \textbf{GICS Sector}\\ 
\hline
DISH & Dish Network & Communication Services\\
EA & Electronic Arts Inc. & Communication Services\\
GOOGL & Alphabet Inc. & Communication Services\\
LUMN & Lumen Technologies Inc. & Communication Services\\
META & Meta Platforms Inc. & Communication Services\\
TTWO & Take Two Interactive Software Inc. & Communication Services\\
\hline
AMZN & Amazon.com Inc. & Communication Discretionary\\
EBAY & eBay Inc. & Communication Discretionary\\
LEN & Lennar Corp. & Communication Discretionary\\
MHK & Mohawk Industries & Communication Discretionary \\
NWL & Newell Brands & Communication Discretionary \\
TSLA & Tesla Inc. & Communication Discretionary\\
\hline
CPB  & Campbell Soup Company & Consumer Staples\\
HSY & Hershey Company & Consumer Staples \\
KMB & Kimberly-Clark Corp. & Consumer Staples\\
PG & Procter \& Gamble Company & Consumer Staples\\
TAP & Molson Coors Beverage Company & Consumer Staples\\
WMT & Walmart Inc. & Consumer Staples\\
\hline
APA & Apache Corporation & Energy\\
CVX & Chevron Corporation & Energy\\
DVN & Devon Energy Corp. 66 & Energy\\
EQT & EQT Corp. & Energy\\
HES & Hess Corp. & Energy\\
XOM & ExxonMobil & Energy\\
\hline
AIZ & Assurant Inc. & Financials\\
BRK-B & Berkshire Hathaway Inc. & Financials\\
JPM & JPMorgan Chase \& Co. & Financials\\
LNC & Lincoln National Corp. & Financials\\
MTB & M\&t Bank Corp. & Financials\\
NDAQ & Nasdaq Inc. & Financials\\
\hline
BIIB & Biogen & Health Care\\
DVA & DaVita HealthCare Partners Inc. & Health Care\\
DXCM & DexCom Inc. & Health Care\\
JNJ & Johnson \& Johnson & Health Care\\
UNH & UnitedHealth Group Inc. & Health Care\\
XRAY & Dentsply Sirona Inc. & Health Care\\
\hline
ALK & Alaska Air Group & Industrials\\
FAST & Fastenal Company & Industrials\\
GNRC & Generac Holdings Inc. & Industrials\\
GWW & W. W. Grainger & Industrials\\
HON & Honeywell International Inc. & Industrials\\
UPS & United Parcel Service of America Inc. & Industrials\\
\hline
AAPL & Apple Inc. & Information Technology\\
FFIV  & F5 Inc. & Information Technology\\
FTNT & Fortinet & Information Technology\\
MSFT & Microsoft Corp. & Information Technology\\
QRVO & Qorvo & Information Technology\\
TEL & TE Connectivity Ltd. & Information Technology\\
\hline
APD & Air Products \& Chemicals Inc. & Materials\\
IFF & International Flavors \& Fragrances Inc. & Materials\\
LIN & Linde plc & Materials\\
SEE & Sealed Air Corp. & Materials\\
VMC & Vulcan Materials Company & Materials\\
WRK & WestRock Company & Materials\\
\hline
AMT & American Tower Corp. & Real Estate\\
AVB & AvalonBay Communities Inc. & Real Estate\\
FRT & Federal Realty Investment Trust & Real Estate\\
PLD & Prologis Inc. & Real Estate\\
VNO & Vornado Realty Trust & Real Estate\\
WY & Weyerhaeuser Company & Real Estate\\
\hline
AWK & American Water Works Company & Utilities\\
DUK & Duke Energy Corp. & Utilities\\
ES & Eversource Energy & Utilities\\
NEE & NextEra Energy Inc. & Utilities\\
NRG & NRG Energy & Utilities\\
PNW & Pinnacle West Capital & Utilities\\
\hline \hline
\end{tabular}
}
\label{tab:source_assets}
\end{table}

\clearpage

\section{Feedforward Neural Networks}
\label{sec:appendix_fnns}
\noindent
FNNs are based on the interconnection of multiple artificial neurons, i.e, computational units, that perform simple calculations in the form of weighted aggregations and (non-linear) transformations. These artificial neurons can be defined by 
\begin{equation}
a = f(\sum_{j=1}^{J} w_{j} z_{j} + b),
\end{equation}    
where the weights, $w_{j}$, and the bias terms, $b$, are trainable parameters, $z_{j}$ are the inputs of the artificial neuron, $J$ denotes the number of inputs, $f$ is an arbitrary activation function, and $a$ is the activation, i.e., the output of the artificial neuron. 

In FNN frameworks, artificial neurons are arranged in multiple layers. The input layer receives training data observations and passes them on to the next layer. The output layer produces the final prediction of the network. Intermediate layers are commonly referred to as hidden layers. In general, for an $L$-layer FNN, the output of the $l$-th layer is given by  
\begin{equation}
\mathbf{a}^{l} = f^{l}(\mathbf{W}^{l} \mathbf{a}^{l-1} + \mathbf{b}^{l} ), \quad 1 \leq l \leq L,
\end{equation} 
where $\mathbf{a}^{l} \in \mathbf{R}^{U_{l}}$ is the activation vector of layer $l$ consisting of $U_{l}$ artificial neurons, $\mathbf{a}^{l-1} \in \mathbf{R}^{U_{l}}$ is the activation vector of the previous layer, $l-1$, consisting of $U_{l-1}$ units, $\mathbf{W}^{l} \in \mathbf{R}^{U_{l} \times U_{l-1}}$ is the weight matrix, $\mathbf{b}^{l} \in \mathbf{R^{U_{l}}}$ is the bias vector, and $f^{l}$ is the activation function of layer $l$. Due to the interconnection of artificial neurons and their non-linear transformations, FNNs with at least one hidden layer are capable of modeling arbitrarily complex non-linear dependency structures \citep{gybenko1989approximation}. 

The number of artificial neurons in the input layer is specified by the number of input features. The number of hidden layers and units within them represent tunable hyperparameters. Similarly, the class of activation functions applied in individual hidden layers also constitutes a tunable hyperparameter. The number of output neurons and their activation functions, on the other hand, are determined by the specific task for which the model is designed. Regression tasks, such as realized variance forecasting, typically require a single output neuron with an identity activation function.  
The training of FNNs is conducted by adjusting the trainable parameters $\theta$, i.e., the connection weights and bias terms, to minimize a cost function $C$, which can be, for instance, the mean squared error loss function,
\begin{equation}
C(\theta) = \frac{1}{n} \sum_{i=1}^{n} (y_i - f_\theta(\textbf{x}_i))^2,
\end{equation}
where $\textbf{x}_{i}$ denotes the $i$-th training example, $y_{i}$ is the corresponding true label, and $n$ is the number of training examples. To minimize $C$, gradient descent or its variants such as Adagrad \citep{duchi2011adaptive} or ADAM optimization \citep{kingma2014adam} are used in combination with the backpropagation algorithm \citep{rumelhart1986learning}. Backpropagation calculates the gradient of the loss function by propagating the model error, i.e., the difference between the predicted output and the true labels, from the last layer to the first layer, while using the chain rule of calculus to determine the partial derivatives of each parameter with respect to $C$. The parameters are then updated by the respective optimization algorithm based on the gradient.

\clearpage

\section{XGBoost}
\label{sec:appendix_xgboost}
\noindent
For a given set of observations $\{\textbf{x}_{i}, y_{i}\}$ of size $n$ with $\textbf{x}_{i} \in \mathbf{R}^{p}$ and $y_{i} \in \mathbf{R}$, consider a single regression tree that predicts according to:
\begin{equation}
f(\textbf{x}_{i}) = \sum_{k=1}^{K} w_{k} I(\textbf{x}_{i} \in R_{k}),\\
\end{equation}
where $K$ is the number of terminal nodes (leaves), $w_{k}$ is the leaf weight of the $k$-th leaf, and $I$ is an indicator function that is one if $\textbf{x}_{i}$ falls into the region $R_{k}$ and zero otherwise. The XGBoost model represents an ensemble of such trees, and predictions are computed according to:
\begin{equation}
\label{eq:min_boost}
F(\textbf{x}_{i}) = \sum_{m=1}^{M} \epsilon f_{m}(\textbf{x}_{i}), \quad f_{m} \in F,\\
\end{equation}
where $f_{m}$ is the $m$-th regression tree, $M$ is the total number of boosting iterations (trees), $\epsilon$ is a shrinkage parameter, and $F$ is the space of regression trees. 

The estimation of this composite model is conducted in a stage-wise manner. In each boosting iteration $m$, a new regression tree $f_{m}$ is added to the existing ensemble and constructed in such a way that it minimizes the following objective:
\begin{equation}
\label{eq:stage_min}
Obj_{m} = \sum_{i=1}^{n} l ( y_{i}, F_{m-1}(\textbf{x}_{i}) + f_{m}(\textbf{x}_{i})) + \Omega(f_{m}), \\
\end{equation}
where $l$ is a differentiable convex loss function, $F_{m-1}$ is the ensemble model up to the $(m-1)$-th iteration, and $\Omega(f_{m})$ is a regularization term. The regularization component, which penalizes the size of the leaf weights (L2 regularization) and the number of terminal nodes, is given by
\begin{equation}
\Omega(f_{m}) = \gamma K_{m} + \frac{1}{2} \lambda \sum^{K_{m}}_{k=1} w^{2}_{m, k}, \\
\end{equation} 
where $\gamma$ and $\lambda$ are regularization parameters, $K_{m}$ is the number of terminal nodes of the $m$-th tree, and $w_{m,k}$ represents the leaf weights of the tree. In contrast to gradient boosting methods, which estimate weak learners with respect to the negative gradient, XGBoost approximates $Obj_{m}$ using second-order Taylor expansion, i.e., Newton boosting, with
\begin{equation}
\label{eq:taylor}
Obj_{m} \simeq \sum_{i=1}^{n} \left[l(y_{i}, F_{m-1}(\textbf{x}_{i})) + g_{i}f_{m}(\textbf{x}_{i}) + \frac{1}{2} h_{i} f_{m}^{2}(\textbf{x}_{i}) \right] + \Omega(f_{m}),
\end{equation} 
where $g_{i}$ and $h_{i}$ are the first and second-order derivatives of the loss function with respect to the predictions of $F_{m-1}$. Empirical evidence suggests that this higher order of approximation combined with regularization leads to more accurate estimates of Eq.~\eqref{eq:min_boost} compared to vanilla gradient tree boosting \citep{sigrist2021gradient}.

For each tree $f_{m}$, after removing the constant terms and considering the leaf weights, the objective simplifies to
\begin{equation}
\label{eq:simp_obj}
\begin{split}
\bar{Obj}_{m} = & \sum_{k=1}^{K_{m}} \left[ \left( \sum_{i=1}^{n} g_{i} I(\textbf{x}_{i} \in R_{m,k}) \right) w_{m, k} \right. \\ 
& \left. + \frac{1}{2} \left( \sum_{i=1}^{n} h_{i} I(\textbf{x}_{i} \in R_{m,k}) + \lambda \right) w^{2}_{m,k} \right] + \gamma K_{m}.
\end{split}
\end{equation} 

When differentiating Eq.~\eqref{eq:simp_obj} with respect to $w_{m,k}$, the optimal weight $w^{*}_{m, k}$ for a given structure $q(\textbf{x})$ is defined by 
\begin{equation}
\label{eq:opt_weights}
w^{*}_{m, k} = - \frac{\sum_{i=1}^{n} g_{i} I(\textbf{x}_{i} \in R_{m,k})}{\sum_{i=1}^{n} h_{i} I(\textbf{x}_{i} \in R_{m,k}) + \lambda}.
\end{equation}
Moreover, substituting the leaf weights of Eq.~\eqref{eq:simp_obj} by the optimal leaf weights of Eq.~\eqref{eq:opt_weights} provides a criterion for evaluating and comparing different tree structures within the boosting process:
\begin{equation}
\label{eq:eval_crit}
\bar{Obj}_{m}(q) = - \frac{1}{2} \sum_{k=1}^{K_{m}} \frac{\left( \sum^{n}_{i=1} g_{i} I(\textbf{x}_{i} \in R_{m,k}) \right)^{2}}{\sum_{i=1}^{n} h_{i} I(\textbf{x}_{i} \in R_{m,k}) + \lambda} + \gamma K_{m}.
\end{equation}
Similar to the Gini impurity measure in classification trees or the mean squared error loss in regression trees, this criterion can be applied to evaluate potential splits during the tree-growing phase of individual weak learners. To determine splits, XGBoost, in its basic implementation, uses an exact greedy algorithm to evaluate all potential split candidates. This involves calculating the potential gain $G$ of a split, based on Eq.~\eqref{eq:eval_crit}, for each split candidate by
\begin{equation}
\label{eq:boost_gain}
\begin{split}
G = & \frac{1}{2} \left[ \frac{(\sum_{i=1}^{n} g_{i} I(\textbf{x}_{i} \in R_{k_{l}}))^2}{\sum_{i=1}^{n} h_{i} I(\textbf{x}_{i} \in R_{k_{l}}) + \lambda} + \frac{(\sum_{i=1}^{n} g_{i} I(\textbf{x}_{i} \in R_{k_{r}}))^2}{\sum_{i=1}^{n} h_{i} I(\textbf{x}_{i} \in R_{k_{r}}) + \lambda} \right. \\ 
& \left. - \frac{(\sum_{i=1}^{n} g_{i} I(\textbf{x}_{i} \in R_{k}))^2}{\sum_{i=1}^{n} h_{i} I(\textbf{x}_{i} \in R_{k}) + \lambda} \right] - \gamma, 
\end{split}
\end{equation}
where $R_{k,l,}$ and $R_{k,r,}$ are the newly generated regions after adding a split. Consequently, the split that produces the highest gain $G$ is selected in the weak learner estimation process. 

Apart from the discussed L2 regularization and the regularization of the number of terminal nodes, XGBoost incorporates  L1 regularization, bottom-up pruning, and top-down pruning methods, e.g., maximum tree depth restrictions or minimum observations per leaf requirements. For a detailed analysis of computational aspects such as approximate split finding or parallel tree learning, we refer to \citet{chen2016xgboost}.

\clearpage

\begin{landscape}

\section{Forecasting Results}
\label{sec:app_forecasting_results}

\newcolumntype{C}[1]{>{\centering\arraybackslash}p{#1}}

\begin{table}[ht!]
\centering
\caption{1-day-ahead cross-sectional average relative MSEs. Each value in this table represents the cross-sectional average of the pairwise realized variance forecast MSE for the model in the selected column relative to the benchmark in the selected row. Each model's MSE is determined for 1-day-ahead realized variance forecasts over 500 days starting 50 days after the first trading day of each target asset. The symbol ($*$) denotes whether the Diebold–Mariano test of equal predictive accuracy is rejected for more than 50\% of the target assets using a 5\% significance level. A rejection for the majority of target assets indicates that the model in the column exhibits a significantly lower MSE than the corresponding benchmark model in the respective row. The corresponding MSE values are additionally marked in blue.}
\scalebox{0.515}{
\begin{tabular}{lC{2cm}C{2cm}C{2cm}C{2cm}C{2cm}C{2cm}C{2cm}C{2cm}C{2cm}C{2cm}C{2cm}C{2cm}C{2cm}C{2cm}C{2cm}C{2cm}}
\toprule
 & NF & \begin{tabular}[c]{@{}c@{}}TO \\ HAR-STD\end{tabular} & \begin{tabular}[c]{@{}c@{}}NP \\ HAR-STD\end{tabular} & \begin{tabular}[c]{@{}c@{}}MTL-25 \\ HAR-STD\end{tabular} &\begin{tabular}[c]{@{}c@{}}MTL-50 \\ HAR-STD\end{tabular} & \begin{tabular}[c]{@{}c@{}}MTL-75 \\ HAR-STD\end{tabular} & \begin{tabular}[c]{@{}c@{}}TO \\ FNN-STD\end{tabular} & \begin{tabular}[c]{@{}c@{}}NP \\ FNN-STD\end{tabular} & \begin{tabular}[c]{@{}c@{}}MTL-25 \\ FNN-STD\end{tabular} & \begin{tabular}[c]{@{}c@{}}MTL-50 \\ FNN-STD\end{tabular} & \begin{tabular}[c]{@{}c@{}}MTL-75 \\ FNN-STD\end{tabular} & \begin{tabular}[c]{@{}c@{}}TO \\ XGB-STD\end{tabular} & \begin{tabular}[c]{@{}c@{}}NP \\ XGB-STD\end{tabular} & \begin{tabular}[c]{@{}c@{}}MTL-25 \\ XGB-STD\end{tabular} & \begin{tabular}[c]{@{}c@{}}MTL-50 \\ XGB-STD\end{tabular} & \begin{tabular}[c]{@{}c@{}}MTL-75 \\ XGB-STD\end{tabular} \\
\midrule
NF & - & 0.807 & \textcolor{blue}{0.745*} & \textcolor{blue}{0.723*} & \textcolor{blue}{0.719*} & \textcolor{blue}{0.717*} & 0.845 & \textcolor{blue}{0.721*} & \textcolor{blue}{0.729*} & \textcolor{blue}{0.724*} & \textcolor{blue}{0.720*} & 1.127 & \textcolor{blue}{0.770*} & \textcolor{blue}{0.789*} & \textcolor{blue}{0.764*} & \textcolor{blue}{0.769*} \\
TO HAR-STD & 1.336 & - & 0.977 & 0.934 & 0.930 & 0.927 & 1.042 & 0.943 & 0.939 & 0.934 & 0.929 & 1.402 & 0.968 & 0.984 & 0.963 & 0.958 \\
TO HAR-EXT & 1.249 & 0.873 & 0.898 & 0.849 & 0.845 & 0.841 & 0.908 & 0.865 & 0.853 & 0.847 & 0.842 & 1.259 & 0.856 & 0.867 & 0.856 & 0.840 \\
NP HAR-STD & 1.353 & 1.063 & - & 0.964 & 0.960 & 0.956 & 1.112 & 0.967 & 0.971 & 0.965 & 0.960 & 1.480 & 1.015 & 1.039 & 1.009 & 1.012 \\
NP HAR-EXT & 1.378 & 1.049 & 1.007 & 0.967 & 0.962 & 0.959 & 1.099 & 0.974 & 0.974 & 0.967 & 0.962 & 1.468 & 1.003 & 1.028 & 1.001 & 0.998 \\
MTL-25 HAR-STD & 1.415 & 1.094 & 1.042 & - & 0.996 & 0.993 & 1.143 & 1.006 & 1.007 & 1.001 & 0.996 & 1.521 & 1.046 & 1.070 & 1.041 & 1.042 \\
MTL-50 HAR-STD & 1.421 & 1.099 & 1.046 & 1.004 & - & 0.997 & 1.148 & 1.010 & 1.012 & 1.005 & 1.000 & 1.528 & 1.050 & 1.074 & 1.046 & 1.046 \\
MTL-75 HAR-STD & 1.425 & 1.102 & 1.049 & 1.007 & 1.003 & - & 1.151 & 1.013 & 1.015 & 1.008 & 1.003 & 1.533 & 1.054 & 1.078 & 1.049 & 1.050 \\
MTL-25 HAR-EXT & 1.521 & 1.151 & 1.112 & 1.063 & 1.058 & 1.055 & 1.203 & 1.073 & 1.070 & 1.063 & 1.058 & 1.594 & 1.101 & 1.124 & 1.098 & 1.094 \\
MTL-50 HAR-EXT & 1.526 & 1.153 & 1.115 & 1.066 & 1.062 & 1.058 & 1.205 & 1.077 & 1.073 & 1.066 & 1.061 & 1.598 & 1.103 & 1.127 & 1.100 & 1.096 \\
MTL-75 HAR-EXT & 1.537 & 1.160 & 1.123 & 1.073 & 1.069 & 1.065 & 1.213 & 1.084 & 1.080 & 1.073 & 1.068 & 1.612 & 1.110 & 1.134 & 1.108 & 1.103 \\
TO FNN-STD & 1.288 & 0.961 & 0.942 & 0.899 & 0.896 & \textcolor{blue}{0.893*} & - & 0.908 & 0.905 & 0.899 & \textcolor{blue}{0.895*} & 1.352 & 0.932 & 0.948 & 0.927 & \textcolor{blue}{0.923*} \\
TO FNN-EXT & 1.307 & 0.921 & 0.942 & 0.891 & 0.887 & 0.884 & 0.957 & 0.907 & 0.895 & 0.890 & 0.885 & 1.310 & 0.902 & 0.913 & 0.901 & 0.887 \\
NP FNN-STD & 1.401 & 1.097 & 1.035 & 0.997 & 0.992 & 0.989 & 1.147 & - & 1.004 & 0.998 & 0.993 & 1.527 & 1.048 & 1.073 & 1.042 & 1.045 \\
NP FNN-EXT & 1.258 & 0.974 & 0.922 & 0.890 & 0.885 & 0.882 & 1.024 & 0.894 & 0.897 & 0.891 & 0.885 & 1.375 & 0.928 & 0.952 & 0.925 & 0.924 \\
MTL-25 FNN-STD & 1.407 & 1.084 & 1.035 & 0.993 & 0.989 & 0.986 & 1.133 & 0.999 & - & 0.994 & 0.989 & 1.508 & 1.037 & 1.060 & 1.033 & 1.033 \\
MTL-50 FNN-STD & 1.415 & 1.091 & 1.041 & 0.999 & 0.995 & 0.992 & 1.140 & 1.005 & 1.006 & - & 0.995 & 1.518 & 1.044 & 1.068 & 1.040 & 1.040 \\
MTL-75 FNN-STD & 1.422 & 1.097 & 1.046 & 1.004 & 1.000 & 0.997 & 1.146 & 1.010 & 1.011 & 1.005 & - & 1.527 & 1.050 & 1.073 & 1.045 & 1.045 \\
MTL-25 FNN-EXT & 1.509 & 1.143 & 1.103 & 1.055 & 1.051 & 1.047 & 1.195 & 1.065 & 1.062 & 1.055 & 1.050 & 1.586 & 1.093 & 1.117 & 1.090 & 1.087 \\
MTL-50 FNN-EXT & 1.531 & 1.162 & 1.120 & 1.071 & 1.067 & 1.063 & 1.215 & 1.081 & 1.079 & 1.072 & 1.066 & 1.613 & 1.111 & 1.135 & 1.108 & 1.105 \\
MTL-75 FNN-ECT & 1.570 & 1.199 & 1.150 & 1.101 & 1.096 & 1.093 & 1.255 & 1.110 & 1.109 & 1.102 & 1.096 & 1.665 & 1.144 & 1.171 & 1.141 & 1.138 \\
TO XGB-STD & 0.991 & \textcolor{blue}{0.742*} & \textcolor{blue}{0.723*} & \textcolor{blue}{0.689*} & \textcolor{blue}{0.686*} & \textcolor{blue}{0.684*} & \textcolor{blue}{0.776*} & \textcolor{blue}{0.697*} & \textcolor{blue}{0.693*} & \textcolor{blue}{0.689*} & \textcolor{blue}{0.686*} & - & \textcolor{blue}{0.715*} & \textcolor{blue}{0.725*} & \textcolor{blue}{0.710*} & \textcolor{blue}{0.707*} \\
TO XGB-EXT & 1.062 & 0.778 & 0.773 & \textcolor{blue}{0.737*} & \textcolor{blue}{0.733*} & \textcolor{blue}{0.731*} & 0.813 & 0.745 & \textcolor{blue}{0.741*} & \textcolor{blue}{0.736*} & \textcolor{blue}{0.732*} & 1.065 & \textcolor{blue}{0.764*} & \textcolor{blue}{0.768*} & \textcolor{blue}{0.755*} & \textcolor{blue}{0.749*} \\
NP XGB-STD & 1.391 & 1.049 & 1.016 & 0.970 & 0.966 & 0.963 & 1.095 & 0.980 & 0.976 & 0.970 & 0.966 & 1.464 & - & 1.026 & 1.001 & 0.997 \\
NP XGB-EXT & 1.465 & 1.109 & 1.068 & 1.020 & 1.016 & 1.012 & 1.161 & 1.031 & 1.027 & 1.020 & 1.015 & 1.545 & 1.049 & 1.080 & 1.053 & 1.048 \\
MTL-25 XGB-STD & 1.359 & 1.021 & 0.993 & 0.948 & 0.944 & 0.941 & 1.066 & 0.958 & 0.954 & 0.948 & 0.944 & 1.419 & 0.981 & - & 0.978 & 0.974 \\
MTL-50 XGB-STD & 1.384 & 1.047 & 1.013 & 0.968 & 0.964 & 0.961 & 1.093 & 0.977 & 0.974 & 0.968 & 0.964 & 1.455 & 1.003 & 1.024 & - & 0.997 \\
MTL-75 XGB-STD & 1.400 & 1.050 & 1.022 & 0.975 & 0.971 & 0.968 & 1.095 & 0.986 & 0.981 & 0.975 & 0.971 & 1.462 & 1.007 & 1.028 & 1.005 & - \\
MTL-25 XGB-EXT & 1.552 & 1.166 & 1.131 & 1.080 & 1.075 & 1.071 & 1.218 & 1.091 & 1.087 & 1.080 & 1.074 & 1.619 & 1.115 & 1.140 & 1.112 & 1.107 \\
MTL-50 XGB-EXT & 1.555 & 1.174 & 1.134 & 1.084 & 1.079 & 1.075 & 1.228 & 1.094 & 1.092 & 1.084 & 1.079 & 1.628 & 1.121 & 1.146 & 1.118 & 1.114 \\
MTL-75 XGB-EXT & 1.618 & 1.216 & 1.178 & 1.126 & 1.121 & 1.117 & 1.271 & 1.137 & 1.134 & 1.126 & 1.120 & 1.695 & 1.162 & 1.189 & 1.160 & 1.154 \\
\bottomrule
\end{tabular}
}
\end{table}

\newpage
\begin{table}[ht!]
\centering
\ContinuedFloat  
\caption{1-day-ahead cross-sectional average relative MSEs (continued). Each value in this table represents the cross-sectional average of the pairwise realized variance forecast MSE for the model in the selected column relative to the benchmark in the selected row. Each model's MSE is determined for 1-day-ahead realized variance forecasts over 500 days starting 50 days after the first trading day of each target asset. The symbol ($*$) denotes whether the Diebold–Mariano test of equal predictive accuracy is rejected for more than 50\% of the target assets using a 5\% significance level. A rejection for the majority of target assets indicates that the model in the column exhibits a significantly lower MSE than the corresponding benchmark model in the respective row. The corresponding MSE values are additionally marked in blue.}
\scalebox{0.515}{
\begin{tabular}{lC{2cm}C{2cm}C{2cm}C{2cm}C{2cm}C{2cm}C{2cm}C{2cm}C{2cm}C{2cm}C{2cm}C{2cm}C{2cm}C{2cm}C{2cm}}
\toprule
 & \begin{tabular}[c]{@{}c@{}}TO \\ HAR-EXT\end{tabular} & \begin{tabular}[c]{@{}c@{}}NP \\ HAR-EXT\end{tabular} & \begin{tabular}[c]{@{}c@{}}MTL-25 \\ HAR-EXT\end{tabular} &\begin{tabular}[c]{@{}c@{}}MTL-50 \\ HAR-EXT\end{tabular} & \begin{tabular}[c]{@{}c@{}}MTL-75 \\ HAR-EXT\end{tabular} & \begin{tabular}[c]{@{}c@{}}TO \\ FNN-EXT\end{tabular} & \begin{tabular}[c]{@{}c@{}}NP \\ FNN-EXT\end{tabular} & \begin{tabular}[c]{@{}c@{}}MTL-25 \\ FNN-EXT\end{tabular} & \begin{tabular}[c]{@{}c@{}}MTL-50 \\ FNN-EXT\end{tabular} & \begin{tabular}[c]{@{}c@{}}MTL-75 \\ FNN-EXT\end{tabular} & \begin{tabular}[c]{@{}c@{}}TO \\ XGB-EXT\end{tabular} & \begin{tabular}[c]{@{}c@{}}NP \\ XGB-EXT\end{tabular} & \begin{tabular}[c]{@{}c@{}}MTL-25 \\ XGB-EXT\end{tabular} & \begin{tabular}[c]{@{}c@{}}MTL-50 \\ XGB-EXT\end{tabular} & \begin{tabular}[c]{@{}c@{}}MTL-75 \\ XGB-EXT\end{tabular} \\
\midrule
NF & 21.310 & \textcolor{blue}{0.782*} & \textcolor{blue}{0.698*} & \textcolor{blue}{0.698*} & \textcolor{blue}{0.695*} & 2.939 & 0.916 & \textcolor{blue}{0.702*} & \textcolor{blue}{0.690*} & \textcolor{blue}{0.668*} & 1.198 & \textcolor{blue}{0.751*} & \textcolor{blue}{0.699*} & \textcolor{blue}{0.691*} & \textcolor{blue}{0.673*} \\
TO HAR-STD & 13.402 & 1.002 & \textcolor{blue}{0.881*} & \textcolor{blue}{0.879*} & \textcolor{blue}{0.874*} & 2.282 & 1.204 & \textcolor{blue}{0.888*} & \textcolor{blue}{0.874*} & \textcolor{blue}{0.854*} & 1.388 & \textcolor{blue}{0.951*} & \textcolor{blue}{0.877*} & \textcolor{blue}{0.872*} & \textcolor{blue}{0.846*} \\
TO HAR-EXT & - & 0.873 & \textcolor{blue}{0.778*} & \textcolor{blue}{0.773*} & \textcolor{blue}{0.765*} & 0.861 & 1.059 & \textcolor{blue}{0.784*} & \textcolor{blue}{0.773*} & \textcolor{blue}{0.757*} & 1.120 & \textcolor{blue}{0.835*} & \textcolor{blue}{0.762*} & 0.764 & \textcolor{blue}{0.733*} \\
NP HAR-STD & 23.633 & 1.038 & \textcolor{blue}{0.922*} & \textcolor{blue}{0.922*} & \textcolor{blue}{0.917*} & 3.405 & 1.226 & \textcolor{blue}{0.928*} & \textcolor{blue}{0.912*} & \textcolor{blue}{0.886*} & 1.543 & 0.990 & 0.921 & \textcolor{blue}{0.911*} & \textcolor{blue}{0.887*} \\
NP HAR-EXT & 18.259 & - & 0.910 & 0.907 & 0.900 & 2.817 & 1.170 & 0.916 & 0.901 & 0.874 & 1.462 & 0.969 & 0.897 & 0.893 & 0.862 \\
MTL-25 HAR-STD & 21.317 & 1.077 & 0.951 & 0.950 & 0.946 & 3.191 & 1.285 & 0.958 & \textcolor{blue}{0.942*} & 0.917 & 1.566 & 1.021 & 0.949 & 0.940 & \textcolor{blue}{0.915*} \\
MTL-50 HAR-STD & 21.499 & 1.081 & 0.955 & 0.954 & 0.950 & 3.214 & 1.289 & 0.962 & \textcolor{blue}{0.946*} & 0.921 & 1.572 & 1.025 & 0.953 & 0.944 & \textcolor{blue}{0.918*} \\
MTL-75 HAR-STD & 21.574 & 1.085 & 0.959 & 0.958 & 0.953 & 3.224 & 1.294 & 0.965 & \textcolor{blue}{0.949*} & 0.923 & 1.577 & 1.028 & 0.956 & 0.947 & 0.921 \\
MTL-25 HAR-EXT & 18.121 & 1.135 & - & 0.998 & 0.993 & 2.904 & 1.363 & 1.007 & 0.992 & 0.966 & 1.602 & 1.069 & 0.995 & 0.987 & 0.959 \\
MTL-50 HAR-EXT & 17.763 & 1.137 & 1.002 & - & 0.995 & 2.867 & 1.366 & 1.009 & 0.994 & 0.968 & 1.600 & 1.071 & 0.996 & 0.989 & 0.960 \\
MTL-75 HAR-EXT & 17.859 & 1.140 & 1.008 & 1.006 & - & 2.882 & 1.370 & 1.015 & 1.000 & 0.973 & 1.608 & 1.076 & 1.001 & 0.994 & 0.964 \\
TO FNN-STD & 12.542 & 0.968 & \textcolor{blue}{0.849*} & \textcolor{blue}{0.847*} & \textcolor{blue}{0.843*} & 2.159 & 1.169 & \textcolor{blue}{0.856*} & \textcolor{blue}{0.843*} & \textcolor{blue}{0.823*} & 1.337 & \textcolor{blue}{0.917*} & \textcolor{blue}{0.846*} & \textcolor{blue}{0.841*} & \textcolor{blue}{0.816*} \\
TO FNN-EXT & 1.869 & 0.930 & \textcolor{blue}{0.821*} & \textcolor{blue}{0.817*} & 0.811 & - & 1.129 & \textcolor{blue}{0.829*} & 0.818 & 0.802 & 1.192 & 0.885 & 0.810 & 0.811 & \textcolor{blue}{0.781*} \\
NP FNN-STD & 23.788 & 1.075 & \textcolor{blue}{0.953*} & \textcolor{blue}{0.952*} & \textcolor{blue}{0.948*} & 3.452 & 1.276 & 0.959 & \textcolor{blue}{0.943*}& \textcolor{blue}{0.916*} & 1.587 & 1.023 & 0.951 & 0.941 & \textcolor{blue}{0.916*} \\
NP FNN-EXT & 19.305 & 0.904 & 0.840 & 0.837 & 0.830 & 2.855 & - & 0.845 & 0.830 & 0.803 & 1.374 & 0.896 & 0.826 & 0.820 & 0.792 \\
MTL-25 FNN-STD & 20.482 & 1.070 & 0.944 & 0.943 & 0.938 & 3.095 & 1.277 & 0.950 & \textcolor{blue}{0.935*} & 0.910 & 1.547 & 1.013 & 0.942 & 0.933 & \textcolor{blue}{0.908*} \\
MTL-50 FNN-STD & 20.839 & 1.076 & 0.950 & 0.949 & 0.944 & 3.139 & 1.284 & 0.956 & \textcolor{blue}{0.941*} & 0.916 & 1.558 & 1.019 & 0.948 & 0.939 & \textcolor{blue}{0.913*} \\
MTL-75 FNN-STD & 21.096 & 1.081 & 0.955 & 0.954 & 0.949 & 3.170 & 1.290 & 0.961 & \textcolor{blue}{0.945*} & 0.920 & 1.567 & 1.024 & 0.952 & 0.943 & 0.917 \\
MTL-25 FNN-EXT & 18.363 & 1.127 & 0.993 & 0.991 & 0.986 & 2.922 & 1.352 & - & 0.984 & 0.959 & 1.595 & 1.061 & 0.987 & 0.980 & 0.951 \\
MTL-50 FNN-EXT & 19.246 & 1.144 & 1.009 & 1.007 & 1.002 & 3.030 & 1.372 & 1.016 & - & 0.974 & 1.625 & 1.077 & 1.003 & 0.995 & 0.966 \\
MTL-75 FNN-ECT & 21.313 & 1.172 & 1.038 & 1.036 & 1.030 & 3.276 & 1.401 & 1.045 & 1.028 & - & 1.684 & 1.105 & 1.031 & 1.023 & 0.992 \\
TO XGB-STD & 10.240 & 0.742 & \textcolor{blue}{0.648*} & \textcolor{blue}{0.647*} & \textcolor{blue}{0.644*} & 1.727 & 0.903 & \textcolor{blue}{0.653*} & \textcolor{blue}{0.644*} & \textcolor{blue}{0.629*} & 1.008 & \textcolor{blue}{0.702*} & \textcolor{blue}{0.645*} & \textcolor{blue}{0.641*} & \textcolor{blue}{0.624*} \\
TO XGB-EXT & 6.394 & 0.774 & \textcolor{blue}{0.685*} & \textcolor{blue}{0.683*} & \textcolor{blue}{0.678*} & 1.352 & 0.943 & \textcolor{blue}{0.692*} & \textcolor{blue}{0.681*} & \textcolor{blue}{0.666*} & - & \textcolor{blue}{0.749*} & \textcolor{blue}{0.678*} & \textcolor{blue}{0.678*} & \textcolor{blue}{0.655*} \\
NP XGB-STD & 16.410 & 1.038 & 0.914 & 0.913 & 0.908 & 2.636 & 1.245 & 0.921 & 0.907 & 0.884 & 1.479 & 0.973 & 0.910 & 0.903 & 0.877 \\
NP XGB-EXT & 18.731 & 1.079 & 0.956 & 0.954 & 0.948 & 2.923 & 1.298 & 0.963 & 0.947 & 0.920 & 1.554 & - & 0.948 & 0.940 & 0.910 \\
MTL-25 XGB-STD & 14.710 & 1.019 & 0.893 & 0.891 & 0.887 & 2.437 & 1.225 & 0.900 & 0.886 & 0.865 & 1.421 & 0.959 & \textcolor{blue}{0.889*} & 0.883 & \textcolor{blue}{0.858*} \\
MTL-50 XGB-STD & 16.577 & 1.040 & 0.914 & 0.912 & 0.908 & 2.653 & 1.247 & 0.920 & 0.906 & 0.883 & 1.468 & 0.980 & 0.910 & 0.902 & \textcolor{blue}{0.877*} \\
MTL-75 XGB-STD & 15.133 & 1.045 & 0.918 & 0.916 & 0.911 & 2.504 & 1.256 & 0.924 & 0.910 & 0.888 & 1.462 & 0.982 & 0.912 & 0.906 & 0.880 \\
MTL-25 XGB-EXT & 17.920 & 1.139 & 1.012 & 1.009 & 1.002 & 2.889 & 1.364 & 1.019 & 1.003 & 0.976 & 1.609 & 1.077 & - & 0.994 & 0.962 \\
MTL-50 XGB-EXT & 19.200 & 1.149 & 1.018 & 1.016 & 1.009 & 3.032 & 1.372 & 1.025 & 1.008 & 0.981 & 1.633 & 1.083 & 1.008 & - & 0.970 \\
MTL-75 XGB-EXT & 19.063 & 1.183 & 1.055 & 1.051 & 1.044 & 3.049 & 1.413 & 1.061 & 1.044 & 1.015 & 1.680 & 1.118 & 1.040 & 1.034 & - \\
\bottomrule
\end{tabular}
}
\label{tab:forecasting_results_mse}
\end{table}

\clearpage

\begin{table}[ht!]
\centering
\caption{1-day-ahead cross-sectional average relative MAEs. Each value in this table represents the cross-sectional average of the pairwise realized variance forecast MAE for the model in the selected column relative to the benchmark in the selected row. Each model's MAE is determined for 1-day-ahead realized variance forecasts over 500 days starting 50 days after the first trading day of each target asset. The symbol ($*$) denotes whether the Diebold–Mariano test of equal predictive accuracy is rejected for more than 50\% of the target assets using a 5\% significance level. A rejection for the majority of target assets indicates that the model in the column exhibits a significantly lower MAE than the corresponding benchmark model in the respective row. The corresponding MAE values are additionally marked in blue.}
\scalebox{0.515}{
\begin{tabular}{lC{2cm}C{2cm}C{2cm}C{2cm}C{2cm}C{2cm}C{2cm}C{2cm}C{2cm}C{2cm}C{2cm}C{2cm}C{2cm}C{2cm}C{2cm}C{2cm}}
\toprule
 & NF & \begin{tabular}[c]{@{}c@{}}TO \\ HAR-STD\end{tabular} & \begin{tabular}[c]{@{}c@{}}NP \\ HAR-STD\end{tabular} & \begin{tabular}[c]{@{}c@{}}MTL-25 \\ HAR-STD\end{tabular} &\begin{tabular}[c]{@{}c@{}}MTL-50 \\ HAR-STD\end{tabular} & \begin{tabular}[c]{@{}c@{}}MTL-75 \\ HAR-STD\end{tabular} & \begin{tabular}[c]{@{}c@{}}TO \\ FNN-STD\end{tabular} & \begin{tabular}[c]{@{}c@{}}NP \\ FNN-STD\end{tabular} & \begin{tabular}[c]{@{}c@{}}MTL-25 \\ FNN-STD\end{tabular} & \begin{tabular}[c]{@{}c@{}}MTL-50 \\ FNN-STD\end{tabular} & \begin{tabular}[c]{@{}c@{}}MTL-75 \\ FNN-STD\end{tabular} & \begin{tabular}[c]{@{}c@{}}TO \\ XGB-STD\end{tabular} & \begin{tabular}[c]{@{}c@{}}NP \\ XGB-STD\end{tabular} & \begin{tabular}[c]{@{}c@{}}MTL-25 \\ XGB-STD\end{tabular} & \begin{tabular}[c]{@{}c@{}}MTL-50 \\ XGB-STD\end{tabular} & \begin{tabular}[c]{@{}c@{}}MTL-75 \\ XGB-STD\end{tabular} \\
\midrule
NF & - & 0.970 & \textcolor{blue}{0.923*} & \textcolor{blue}{0.850*} & \textcolor{blue}{0.845*} & \textcolor{blue}{0.847*} & 1.018 & \textcolor{blue}{0.895*} & \textcolor{blue}{0.857*} & \textcolor{blue}{0.849*} & \textcolor{blue}{0.850*} & 1.109 & \textcolor{blue}{0.877*} & \textcolor{blue}{0.860*} & \textcolor{blue}{0.850*} & \textcolor{blue}{0.858*} \\
TO HAR-STD & 1.043 & - & 0.963 & \textcolor{blue}{0.884*} & \textcolor{blue}{0.878*} & \textcolor{blue}{0.881*} & 1.046 & \textcolor{blue}{0.932*} & \textcolor{blue}{0.890*} & \textcolor{blue}{0.882*} & \textcolor{blue}{0.883*} & 1.142 & \textcolor{blue}{0.912*} & \textcolor{blue}{0.889*} & \textcolor{blue}{0.881*} & \textcolor{blue}{0.887*} \\
TO HAR-EXT & 0.913 & 0.857 & \textcolor{blue}{0.840*} & \textcolor{blue}{0.769*} & \textcolor{blue}{0.764*} & \textcolor{blue}{0.765*} & 0.891 & \textcolor{blue}{0.812*} & \textcolor{blue}{0.773*} & \textcolor{blue}{0.766*} & \textcolor{blue}{0.766*} & 0.975 & \textcolor{blue}{0.790*} & \textcolor{blue}{0.765*} & \textcolor{blue}{0.761*} & \textcolor{blue}{0.762*} \\
NP HAR-STD & 1.085 & 1.052 & - & \textcolor{blue}{0.921*} & \textcolor{blue}{0.915*} & \textcolor{blue}{0.918*} & 1.105 & 0.971 & \textcolor{blue}{0.928*} & \textcolor{blue}{0.920*} & \textcolor{blue}{0.920*} & 1.200 & \textcolor{blue}{0.949*} & \textcolor{blue}{0.931*} & \textcolor{blue}{0.921*} & \textcolor{blue}{0.929*} \\
NP HAR-EXT & 1.040 & 1.005 & \textcolor{blue}{0.957*} & \textcolor{blue}{0.882*} & \textcolor{blue}{0.877*} & \textcolor{blue}{0.879*} & 1.054 & \textcolor{blue}{0.929*} & \textcolor{blue}{0.889*} & \textcolor{blue}{0.880*} & \textcolor{blue}{0.881*} & 1.146 & \textcolor{blue}{0.907*} & \textcolor{blue}{0.890*} & \textcolor{blue}{0.880*} & \textcolor{blue}{0.887*} \\
MTL-25 HAR-STD & 1.180 & 1.141 & 1.088 & - & 0.994 & 0.997 & 1.196 & 1.055 & 1.008 & 0.998 & 0.999 & 1.300 & 1.031 & 1.010 & 1.000 & 1.008 \\
MTL-50 HAR-STD & 1.187 & 1.147 & 1.094 & 1.006 & - & 1.003 & 1.203 & 1.061 & 1.014 & 1.004 & 1.005 & 1.307 & 1.038 & 1.016 & 1.005 & 1.014 \\
MTL-75 HAR-STD & 1.184 & 1.144 & 1.091 & 1.003 & 0.997 & - & 1.200 & 1.058 & 1.011 & 1.002 & 1.003 & 1.304 & 1.035 & 1.013 & 1.002 & 1.011 \\
MTL-25 HAR-EXT & 1.200 & 1.154 & 1.105 & 1.015 & 1.009 & 1.012 & 1.208 & 1.071 & 1.023 & 1.013 & 1.014 & 1.313 & 1.046 & 1.022 & 1.013 & 1.020 \\
MTL-50 HAR-EXT & 1.206 & 1.158 & 1.110 & 1.020 & 1.014 & 1.016 & 1.213 & 1.076 & 1.027 & 1.018 & 1.018 & 1.318 & 1.051 & 1.027 & 1.017 & 1.024 \\
MTL-75 HAR-EXT & 1.203 & 1.156 & 1.108 & 1.018 & 1.011 & 1.014 & 1.210 & 1.073 & 1.025 & 1.015 & 1.016 & 1.315 & 1.048 & 1.024 & 1.015 & 1.021 \\
TO FNN-STD & 1.001 & \textcolor{blue}{0.957*} & 0.924 & \textcolor{blue}{0.848*} & \textcolor{blue}{0.842*} & \textcolor{blue}{0.845*} & - & \textcolor{blue}{0.894*} & \textcolor{blue}{0.854*} & \textcolor{blue}{0.846*} & \textcolor{blue}{0.846*} & 1.093 & \textcolor{blue}{0.875*} & \textcolor{blue}{0.852*} & \textcolor{blue}{0.844*} & \textcolor{blue}{0.850*} \\
TO FNN-EXT & 0.958 & 0.905 & \textcolor{blue}{0.881*} & \textcolor{blue}{0.808*} & \textcolor{blue}{0.802*} & \textcolor{blue}{0.804*} & 0.943 & \textcolor{blue}{0.852*} & \textcolor{blue}{0.813*} & \textcolor{blue}{0.805*} & \textcolor{blue}{0.805*} & 1.029 & \textcolor{blue}{0.831*} & \textcolor{blue}{0.807*} & \textcolor{blue}{0.802*} & \textcolor{blue}{0.804*} \\
NP FNN-STD & 1.119 & 1.082 & 1.031 & \textcolor{blue}{0.949*} & \textcolor{blue}{0.943*} & \textcolor{blue}{0.946*} & 1.135 & - & \textcolor{blue}{0.956*} & \textcolor{blue}{0.947*} & \textcolor{blue}{0.948*} & 1.234 & 0.978 & \textcolor{blue}{0.958*} & \textcolor{blue}{0.948*} & \textcolor{blue}{0.956*} \\
NP FNN-EXT & 0.954 & 0.924 & \textcolor{blue}{0.877*} & \textcolor{blue}{0.810*} & \textcolor{blue}{0.805*} & \textcolor{blue}{0.807*} & 0.971 & \textcolor{blue}{0.853*} & \textcolor{blue}{0.816*} & \textcolor{blue}{0.808*} & \textcolor{blue}{0.809*} & 1.057 & \textcolor{blue}{0.833*} & \textcolor{blue}{0.818*} & \textcolor{blue}{0.809*} & \textcolor{blue}{0.815*} \\
MTL-25 FNN-STD & 1.172 & 1.132 & 1.080 & \textcolor{blue}{0.992*} & \textcolor{blue}{0.987*} & 0.989 & 1.186 & 1.047 & - & \textcolor{blue}{0.991*} & 0.992 & 1.289 & 1.024 & 1.002 & 0.992 & 1.000 \\
MTL-50 FNN-STD & 1.183 & 1.142 & 1.090 & 1.002 & 0.996 & 0.999 & 1.197 & 1.056 & 1.009 & - & 1.001 & 1.301 & 1.033 & 1.011 & 1.001 & 1.009 \\
MTL-75 FNN-STD & 1.181 & 1.141 & 1.089 & 1.001 & 0.995 & 0.997 & 1.196 & 1.055 & 1.009 & 0.999 & - & 1.300 & 1.032 & 1.010 & 1.000 & 1.008 \\
MTL-25 FNN-EXT & 1.187 & 1.141 & 1.093 & 1.004 & 0.998 & 1.000 & 1.195 & 1.059 & 1.011 & 1.002 & 1.003 & 1.300 & 1.035 & 1.011 & 1.002 & 1.009 \\
MTL-50 FNN-EXT & 1.204 & 1.158 & 1.109 & 1.019 & 1.012 & 1.015 & 1.213 & 1.074 & 1.026 & 1.016 & 1.017 & 1.319 & 1.050 & 1.026 & 1.016 & 1.023 \\
MTL-75 FNN-ECT & 1.207 & 1.162 & 1.112 & 1.022 & 1.015 & 1.018 & 1.218 & 1.077 & 1.029 & 1.020 & 1.020 & 1.324 & 1.053 & 1.030 & 1.019 & 1.027 \\
TO XGB-STD & 0.921 & \textcolor{blue}{0.883*} & \textcolor{blue}{0.848*} & \textcolor{blue}{0.778*} & \textcolor{blue}{0.773*} & \textcolor{blue}{0.775*} & \textcolor{blue}{0.923*} & \textcolor{blue}{0.821*} & \textcolor{blue}{0.784*} & \textcolor{blue}{0.776*} & \textcolor{blue}{0.777*} & - & \textcolor{blue}{0.802*} & \textcolor{blue}{0.782*} & \textcolor{blue}{0.776*} & \textcolor{blue}{0.781*} \\
TO XGB-EXT & 0.935 & 0.891 & \textcolor{blue}{0.861*} & \textcolor{blue}{0.790*} & \textcolor{blue}{0.785*} & \textcolor{blue}{0.787*} & 0.931 & \textcolor{blue}{0.833*} & \textcolor{blue}{0.795*} & \textcolor{blue}{0.788*} & \textcolor{blue}{0.788*} & 1.012 & \textcolor{blue}{0.815*} & \textcolor{blue}{0.793*} & \textcolor{blue}{0.786*} & \textcolor{blue}{0.790*} \\
NP XGB-STD & 1.150 & 1.112 & 1.058 & 0.974 & 0.968 & 0.971 & 1.165 & 1.027 & 0.981 & 0.972 & 0.973 & 1.265 & - & 0.982 & 0.972 & 0.979 \\
NP XGB-EXT & 1.173 & 1.132 & 1.079 & 0.993 & 0.987 & 0.989 & 1.187 & 1.047 & 1.000 & 0.991 & 0.991 & 1.288 & 1.020 & 1.001 & 0.991 & 0.998 \\
MTL-25 XGB-STD & 1.174 & 1.129 & 1.082 & 0.993 & 0.987 & 0.990 & 1.182 & 1.048 & 1.001 & 0.991 & 0.992 & 1.285 & 1.024 & - & 0.990 & 0.998 \\
MTL-50 XGB-STD & 1.184 & 1.141 & 1.091 & 1.003 & 0.996 & 0.999 & 1.195 & 1.057 & 1.010 & 1.001 & 1.002 & 1.299 & 1.033 & 1.010 & - & 1.007 \\
MTL-75 XGB-STD & 1.177 & 1.132 & 1.085 & 0.997 & 0.990 & 0.993 & 1.185 & 1.051 & 1.004 & 0.994 & 0.995 & 1.289 & 1.026 & 1.003 & 0.993 & - \\
MTL-25 XGB-EXT & 1.213 & 1.166 & 1.117 & 1.027 & 1.020 & 1.023 & 1.221 & 1.082 & 1.034 & 1.024 & 1.025 & 1.328 & 1.057 & 1.033 & 1.023 & 1.030 \\
MTL-50 XGB-EXT & 1.223 & 1.176 & 1.126 & 1.035 & 1.028 & 1.031 & 1.232 & 1.091 & 1.042 & 1.032 & 1.033 & 1.339 & 1.066 & 1.042 & 1.031 & 1.039 \\
MTL-75 XGB-EXT & 1.224 & 1.177 & 1.127 & 1.036 & 1.029 & 1.032 & 1.233 & 1.092 & 1.044 & 1.033 & 1.034 & 1.341 & 1.066 & 1.043 & 1.032 & 1.039 \\
\bottomrule
\end{tabular}
}
\end{table}

\newpage

\begin{table}[ht!]
\centering
\ContinuedFloat  
\caption{1-day-ahead cross-sectional average relative MAEs (continued). Each value in this table represents the cross-sectional average of the pairwise realized variance forecast MAE for the model in the selected column relative to the benchmark in the selected row. Each model's MAE is determined for 1-day-ahead realized variance forecasts over 500 days starting 50 days after the first trading day of each target asset. The symbol ($*$) denotes whether the Diebold–Mariano test of equal predictive accuracy is rejected for more than 50\% of the target assets using a 5\% significance level. A rejection for the majority of target assets indicates that the model in the column exhibits a significantly lower MAE than the corresponding benchmark model in the respective row. The corresponding MAE values are additionally marked in blue.}
\scalebox{0.515}{
\begin{tabular}{lC{2cm}C{2cm}C{2cm}C{2cm}C{2cm}C{2cm}C{2cm}C{2cm}C{2cm}C{2cm}C{2cm}C{2cm}C{2cm}C{2cm}C{2cm}}
\toprule
 & \begin{tabular}[c]{@{}c@{}}TO \\ HAR-EXT\end{tabular} & \begin{tabular}[c]{@{}c@{}}NP \\ HAR-EXT\end{tabular} & \begin{tabular}[c]{@{}c@{}}MTL-25 \\ HAR-EXT\end{tabular} &\begin{tabular}[c]{@{}c@{}}MTL-50 \\ HAR-EXT\end{tabular} & \begin{tabular}[c]{@{}c@{}}MTL-75 \\ HAR-EXT\end{tabular} & \begin{tabular}[c]{@{}c@{}}TO \\ FNN-EXT\end{tabular} & \begin{tabular}[c]{@{}c@{}}NP \\ FNN-EXT\end{tabular} & \begin{tabular}[c]{@{}c@{}}MTL-25 \\ FNN-EXT\end{tabular} & \begin{tabular}[c]{@{}c@{}}MTL-50 \\ FNN-EXT\end{tabular} & \begin{tabular}[c]{@{}c@{}}MTL-75 \\ FNN-EXT\end{tabular} & \begin{tabular}[c]{@{}c@{}}TO \\ XGB-EXT\end{tabular} & \begin{tabular}[c]{@{}c@{}}NP \\ XGB-EXT\end{tabular} & \begin{tabular}[c]{@{}c@{}}MTL-25 \\ XGB-EXT\end{tabular} & \begin{tabular}[c]{@{}c@{}}MTL-50 \\ XGB-EXT\end{tabular} & \begin{tabular}[c]{@{}c@{}}MTL-75 \\ XGB-EXT\end{tabular} \\
\midrule
NF & 1.542 & 0.970 & \textcolor{blue}{0.842*} & \textcolor{blue}{0.839*} & \textcolor{blue}{0.841*} & 1.169 & 1.081 & \textcolor{blue}{0.851*} & \textcolor{blue}{0.838*} & \textcolor{blue}{0.835*} & 1.117 & \textcolor{blue}{0.862*} & \textcolor{blue}{0.834*} & \textcolor{blue}{0.826*} & \textcolor{blue}{0.826*} \\
TO HAR-STD & 1.460 & 1.009 & \textcolor{blue}{0.870*} & \textcolor{blue}{0.867*} & \textcolor{blue}{0.869*} & 1.165 & 1.127 & \textcolor{blue}{0.880*} & \textcolor{blue}{0.867*} & \textcolor{blue}{0.865*} & 1.140 & \textcolor{blue}{0.896*} & \textcolor{blue}{0.862*} & \textcolor{blue}{0.855*} & \textcolor{blue}{0.855*} \\
TO HAR-EXT & - & 0.871 & \textcolor{blue}{0.747*} & \textcolor{blue}{0.742*} & \textcolor{blue}{0.744*} & 0.921 & 0.975 & \textcolor{blue}{0.756*} & \textcolor{blue}{0.746*} & \textcolor{blue}{0.745*} & 0.949 & \textcolor{blue}{0.773*} & \textcolor{blue}{0.739*} & \textcolor{blue}{0.735*} & \textcolor{blue}{0.734*} \\
NP HAR-STD & 1.658 & 1.050 & \textcolor{blue}{0.911*} & \textcolor{blue}{0.908*} & \textcolor{blue}{0.910*} & 1.261 & 1.168 & \textcolor{blue}{0.921*} & \textcolor{blue}{0.907*} & \textcolor{blue}{0.904*} & 1.209 & \textcolor{blue}{0.932*} & \textcolor{blue}{0.903*} & \textcolor{blue}{0.895*} & \textcolor{blue}{0.894*} \\
NP HAR-EXT & 1.530 & - & \textcolor{blue}{0.870*} & \textcolor{blue}{0.867*} & \textcolor{blue}{0.868*} & 1.188 & 1.110 & \textcolor{blue}{0.880*} & \textcolor{blue}{0.867*} & \textcolor{blue}{0.863*} & 1.147 & \textcolor{blue}{0.889*} & \textcolor{blue}{0.861*} & \textcolor{blue}{0.853*} & \textcolor{blue}{0.853*} \\
MTL-25 HAR-STD & 1.763 & 1.143 & \textcolor{blue}{0.988*} & 0.985 & 0.988 & 1.358 & 1.275 & 0.999 & 0.985 & 0.982 & 1.308 & 1.013 & 0.979 & 0.971 & 0.971 \\
MTL-50 HAR-STD & 1.771 & 1.149 & 0.994 & \textcolor{blue}{0.991*} & 0.993 & 1.365 & 1.283 & 1.005 & 0.990 & 0.987 & 1.314 & 1.019 & 0.985 & 0.976 & 0.976 \\
MTL-75 HAR-STD  & 1.764 & 1.146 & 0.991 & \textcolor{blue}{0.988*} & \textcolor{blue}{0.990*} & 1.360 & 1.279 & 1.002 & 0.987 & 0.984 & 1.310 & 1.016 & 0.982 & 0.973 & 0.973 \\
MTL-25 HAR-EXT & 1.705 & 1.158 & - & 0.996 & 0.999 & 1.349 & 1.293 & 1.011 & 0.997 & 0.994 & 1.314 & 1.027 & 0.990 & 0.982 & 0.982 \\
MTL-50 HAR-EXT & 1.703 & 1.163 & 1.004 & - & 1.003 & 1.352 & 1.299 & 1.016 & 1.001 & 0.998 & 1.318 & 1.031 & 0.994 & 0.986 & 0.986 \\
MTL-75 HAR-EXT & 1.698 & 1.159 & 1.002 & 0.998 & - & 1.348 & 1.296 & 1.013 & 0.998 & 0.995 & 1.314 & 1.028 & 0.992 & 0.984 & 0.983 \\
TO FNN-STD & 1.370 & \textcolor{blue}{0.968*} & \textcolor{blue}{0.834*} & \textcolor{blue}{0.830*} & \textcolor{blue}{0.833*} & 1.107 & 1.084 & \textcolor{blue}{0.844*} & \textcolor{blue}{0.831*} & \textcolor{blue}{0.829*} & 1.090 & \textcolor{blue}{0.859*} & \textcolor{blue}{0.826*} & \textcolor{blue}{0.820*} & \textcolor{blue}{0.819*} \\
TO FNN-EXT & 1.151 & 0.918 & \textcolor{blue}{0.789*} & \textcolor{blue}{0.784*} & \textcolor{blue}{0.786*} & - & 1.027 & \textcolor{blue}{0.798*} & \textcolor{blue}{0.787*} & \textcolor{blue}{0.785*} & 1.012 & \textcolor{blue}{0.814*} & \textcolor{blue}{0.781*} & \textcolor{blue}{0.776*} & \textcolor{blue}{0.775*} \\
NP FNN-STD & 1.686 & 1.083 & \textcolor{blue}{0.938*} & \textcolor{blue}{0.935*} & \textcolor{blue}{0.937*} & 1.292 & 1.208 & \textcolor{blue}{0.948*} & \textcolor{blue}{0.934*} & \textcolor{blue}{0.931*} & 1.241 & \textcolor{blue}{0.961*} & \textcolor{blue}{0.929*} & \textcolor{blue}{0.921*} & \textcolor{blue}{0.921*} \\
NP FNN-EXT & 1.418 & \textcolor{blue}{0.913*} & \textcolor{blue}{0.799*} & \textcolor{blue}{0.796*} & \textcolor{blue}{0.797*} & 1.095 & - & \textcolor{blue}{0.807*} & \textcolor{blue}{0.795*} & \textcolor{blue}{0.792*} & 1.053 & \textcolor{blue}{0.817*} & \textcolor{blue}{0.790*} & \textcolor{blue}{0.783*} & \textcolor{blue}{0.783*} \\
MTL-25 FNN-STD & 1.736 & 1.134 & \textcolor{blue}{0.980*} & \textcolor{blue}{0.977*} & \textcolor{blue}{0.980*} & 1.343 & 1.266 & 0.991 & \textcolor{blue}{0.977*} & 0.974 & 1.296 & 1.006 & 0.971 & 0.963 & 0.963 \\
MTL-50 FNN-STD & 1.753 & 1.144 & 0.990 & \textcolor{blue}{0.986*} & \textcolor{blue}{0.989*} & 1.356 & 1.278 & 1.001 & 0.986 & 0.983 & 1.308 & 1.015 & 0.980 & 0.972 & 0.972 \\
MTL-75 FNN-STD & 1.752 & 1.143 & 0.989 & \textcolor{blue}{0.985*} & \textcolor{blue}{0.988*} & 1.354 & 1.276 & 1.000 & 0.985 & 0.981 & 1.306 & 1.013 & 0.979 & 0.971 & 0.971 \\
MTL-25 FNN-EXT & 1.691 & 1.145 & 0.989 & 0.985 & 0.988 & 1.336 & 1.277 & - & \textcolor{blue}{0.986*} & 0.983 & 1.300 & 1.016 & 0.980 & 0.972 & 0.972 \\
MTL-50 FNN-EXT & 1.723 & 1.161 & 1.004 & 1.000 & 1.002 & 1.357 & 1.297 & 1.015 & - & 0.997 & 1.319 & 1.030 & 0.994 & 0.986 & 0.985 \\
MTL-75 FNN-ECT & 1.741 & 1.164 & 1.007 & 1.003 & 1.005 & 1.366 & 1.300 & 1.018 & 1.003 & - & 1.325 & 1.033 & 0.997 & 0.989 & 0.988 \\
TO XGB-STD & 1.270 & \textcolor{blue}{0.889*} & \textcolor{blue}{0.766*} & \textcolor{blue}{0.762*} & \textcolor{blue}{0.765*} & 1.022 & 0.998 & \textcolor{blue}{0.775*} & \textcolor{blue}{0.764*} & \textcolor{blue}{0.761*} & 1.000 & \textcolor{blue}{0.787*} & \textcolor{blue}{0.758*} & \textcolor{blue}{0.753*} & \textcolor{blue}{0.753*} \\
TO XGB-EXT & 1.219 & 0.898 & \textcolor{blue}{0.774*} & \textcolor{blue}{0.771*} & \textcolor{blue}{0.772*} & 1.011 & 1.004 & \textcolor{blue}{0.783*} & \textcolor{blue}{0.772*} & \textcolor{blue}{0.770*} & - & \textcolor{blue}{0.798*} & \textcolor{blue}{0.766*} & \textcolor{blue}{0.761*} & \textcolor{blue}{0.761*} \\
NP XGB-STD & 1.699 & 1.109 & 0.961 & 0.958 & 0.960 & 1.316 & 1.239 & 0.973 & 0.958 & 0.955 & 1.272 & 0.982 & 0.952 & 0.944 & 0.944 \\
NP XGB-EXT & 1.722 & 1.129 & 0.979 & 0.975 & 0.977 & 1.337 & 1.261 & 0.991 & 0.976 & 0.972 & 1.292 & - & 0.969 & 0.961 & 0.960 \\
MTL-25 XGB-STD & 1.675 & 1.134 & 0.979 & 0.975 & 0.978 & 1.323 & 1.267 & 0.990 & 0.975 & 0.973 & 1.287 & 1.005 & 0.969 & 0.961 & 0.961 \\
MTL-50 XGB-STD & 1.717 & 1.144 & 0.989 & 0.985 & 0.988 & 1.344 & 1.278 & 1.000 & 0.985 & 0.982 & 1.303 & 1.014 & 0.979 & 0.971 & 0.971 \\
MTL-75 XGB-STD & 1.676 & 1.136 & 0.982 & 0.977 & 0.980 & 1.325 & 1.269 & 0.993 & 0.978 & 0.975 & 1.291 & 1.007 & 0.972 & 0.964 & 0.963 \\
MTL-25 XGB-EXT & 1.722 & 1.169 & 1.011 & 1.006 & 1.009 & 1.363 & 1.305 & 1.022 & 1.007 & 1.004 & 1.327 & 1.037 & - & 0.992 & 0.992 \\
MTL-50 XGB-EXT & 1.753 & 1.178 & 1.019 & 1.015 & 1.017 & 1.380 & 1.315 & 1.031 & 1.015 & 1.012 & 1.340 & 1.045 & 1.008 & - & 1.000 \\
MTL-75 XGB-EXT & 1.751 & 1.178 & 1.020 & 1.016 & 1.018 & 1.379 & 1.315 & 1.031 & 1.016 & 1.013 & 1.341 & 1.045 & 1.009 & 1.001 & - \\
\bottomrule
\end{tabular}
}
\label{tab:forecasting_results_mae}
\end{table}

\end{landscape}

\clearpage

\section{Selection rates of source assets subsequences for MTL models}
\label{apdx:subequence_selection_rates}
\vspace*{\fill}

\begin{table}[ht!]
\centering
\caption{Average selection rates of source asset subsequences for MTL-25 models of individual target assets. Source assets that belong to the same GICS sector as the respective target asset are highlighted in gray. Selection rates greater than 45\% are highlighted in blue, while selection rates of less than 5\% are highlighted in red. Hyphens (-) indicate the absence of available subsequences for selection due to the later distribution date of the source asset compared to the target asset.} 
\scalebox{0.5}{
\begin{tabular}{llx{1.75cm}x{1.75cm}x{1.75cm}x{1.75cm}x{1.75cm}x{1.75cm}x{1.75cm}x{1.75cm}x{1.75cm}x{1.75cm}}
\hline \hline
\textbf{GICS Sector} & \textbf{Source Asset}  & \multicolumn{10}{c}{\textbf{Target Asset}}  \\ 
&  & \textbf{TWTR} &  \textbf{NCLH} & \textbf{LW} & \textbf{PSX} &  \textbf{SYF} & \textbf{MRNA} & \textbf{CARR} &  \textbf{DXC} & \textbf{CTVA} &  \textbf{INVH} \\ 
\hline
\multirow{6}{*}{Communication Services} & DISH  & \cellcolor{lightgray} \color{blue}  0.524 & 0.415 &  0.248 &  0.392 &  0.374 & \color{blue} 0.521 &  0.384 &  0.227 & 0.439 &  0.100 \\
& EA    & \cellcolor{lightgray} \color{blue} 0.620 &  0.328 &  0.177 &  0.223 &  0.310 &  \color{blue} 0.572 &  0.358 &  0.182 &  0.382 & 0.094 \\
& GOOGL & \cellcolor{lightgray} 0.140 &  0.189 &  0.229 &  0.172 &  0.217 &  0.198 &  0.170 &  0.229 &  0.154 &  0.250 \\
& LUMN  & \cellcolor{lightgray} \color{red} 0.042 &  0.130 &  0.290 &  0.137 &  0.168 &  0.241 &  0.242 &  0.292 &  0.223 &  0.323 \\
& META  & \cellcolor{lightgray} \color{blue} 0.658 &  0.379 &  0.193 &  0.284 &  0.337 & 0.409 &  0.303 &  0.190 &  0.301 &  0.134 \\
& TTWO  & \cellcolor{lightgray} \color{blue} 0.697 &  0.294 &  0.146 &  0.139 &  0.296 & \color{blue} 0.687 &  0.395 &  0.158 & 0.418 & 0.054 \\
\hline
\multirow{6}{*}{Communication Discretionary} & AMZN  & 0.429 & \cellcolor{lightgray} 0.412 &  0.211 & 0.412 &  0.365 &  0.350 &  0.260 &  0.182 &  0.291 &  0.105 \\
& EBAY  &  0.400 & \cellcolor{lightgray} \color{blue} 0.479 &  0.365 & \color{blue} 0.515 & 0.426 &  0.228 &  0.363 &  0.309 &  0.377 &  0.242 \\
& LEN   & \color{blue} 0.707 & \cellcolor{lightgray} 0.309 &  0.208 &  0.145 &  0.297 & \color{blue} 0.595 &  0.366 &  0.206 &  0.390 &  0.113 \\
& MHK   & \color{blue} 0.584 & \cellcolor{lightgray} \color{blue} 0.455 &  0.284 &  0.381 &  0.399 & 0.401 &  0.360 &  0.257 &  0.391 &  0.187 \\
& NWL   &  0.233 & \cellcolor{lightgray} 0.346 &  0.338 & 0.431 &  0.328 &  0.323 &  0.302 &  0.304 &  0.300 &  0.291 \\
& TSLA  & \color{blue} 0.487 & \cellcolor{lightgray} 0.148 &  0.096 &  0.061 &  0.190 & \color{blue} 0.868 &  0.335 &  0.215 &  0.328 &  0.088 \\
\hline
\multirow{6}{*}{Consumer Staples} & CPB   & \color{red}  0.026 & 0.097 & \cellcolor{lightgray} 0.222 &  0.106 &  0.102 & 0.052 &  0.159 &  0.233 &  0.142 &  0.291 \\
& HSY   & \color{red}  0.032 & 0.086 & \cellcolor{lightgray} 0.185 &  0.107 & 0.090 & \color{red}  0.024 &  0.105 &  0.221 & 0.084 &  0.288 \\
& KMB   & \color{red}  0.022 & 0.065 & \cellcolor{lightgray} 0.133 & 0.077 &  0.064 & \color{red}  0.017 &  0.091 &  0.162 & 0.072 &  0.217 \\
& PG    & \color{red}  0.021 & \color{red}  0.047 & \cellcolor{lightgray} 0.076 &  0.059 & \color{red}  0.046 & \color{red}  0.017 & 0.052 &  0.116 & \color{red}  0.037 &  0.166 \\
& TAP   & 0.064 &  0.156 & \cellcolor{lightgray} 0.280 &  0.142 &  0.171 & 0.091 &  0.197 &  0.273 &  0.167 &  0.319 \\
& WMT   & \color{red}  0.024 & 0.055 & \cellcolor{lightgray} 0.120 & 0.054 & 0.055 & \color{red}  0.025 & 0.066 &  0.156 & 0.055 &  0.225 \\
\hline
\multirow{6}{*}{Energy} & APA   &  0.306 &  0.366 &  0.268 & \cellcolor{lightgray} \color{blue} 0.462 &  0.357 & \color{blue} 0.527 &  0.339 &  0.257 &  0.387 &  0.189 \\
& CVX   &  0.102 &  0.159 &  0.223 & \cellcolor{lightgray} 0.190 &  0.191 &  0.104 &  0.194 &  0.250 &  0.170 &  0.316 \\
& DVN   &  0.313 &  0.390 &  0.270 & \cellcolor{lightgray} \color{blue} 0.456 &  0.365 &  \color{blue} 0.508 &  0.338 &  0.243 &  0.389 &  0.161 \\
& EQT   &  0.321 &  0.437 &  0.313 & \cellcolor{lightgray} \color{blue} 0.452 &  0.423 &  0.438 &  0.348 &  0.255 &  0.418 &  0.146 \\
& HES   &  0.381 &  0.395 &  0.246 & \cellcolor{lightgray} \color{blue} 0.481 &  0.368 &  \color{blue} 0.548 &  0.388 &  0.233 &  0.438 &  0.139 \\
& XOM   & 0.060 &  0.112 &  0.176 & \cellcolor{lightgray} 0.125 &  0.135 & 0.071 &  0.141 &  0.201 &  0.114 &  0.264 \\
\hline
\multirow{6}{*}{Financials} &  AIZ   &  0.136 &  0.253 &  0.354 &  0.306 & \cellcolor{lightgray} 0.276 &  0.114 &  0.244 &  0.347 &  0.224 &  0.380 \\
& BRK-B & 0.078 & 0.090 & 0.091 &  0.111 & \cellcolor{lightgray} 0.084 & 0.051 & 0.078 &  0.110 & 0.076 &  0.141 \\
& JPM   &  0.225 &  0.337 &  0.310 &  0.440 & \cellcolor{lightgray} 0.327 &  0.135 &  0.248 &  0.308 &  0.240 &  0.337 \\
& LNC   & \color{blue} 0.455 &  0.415 &  0.321 &  0.367 & \cellcolor{lightgray} 0.381 &  0.337 &  0.375 &  0.280 &  0.413 &  0.194 \\
& MTB   &  0.194 &  0.212 &  0.310 &  0.230 & \cellcolor{lightgray} 0.218 &  0.155 &  0.249 &  0.327 &  0.219 &  0.370 \\
& NDAQ  &  0.236 &  0.374 &  0.350 &  0.423 & \cellcolor{lightgray} 0.377 &  0.140 &  0.261 &  0.314 &  0.255 &  0.316 \\
\hline
\multirow{6}{*}{Health Care} & BIIB  &  0.405 &  0.357 &  0.289 &  0.342 &  0.371 & \cellcolor{lightgray} 0.424 &  0.384 &  0.257 &  0.416 &  0.167 \\
& DVA   & 0.069 &  0.163 &  0.309 &  0.167 &  0.181 & \cellcolor{lightgray} 0.136 &  0.224 &  0.315 &  0.194 &  0.375 \\
& DXCM  & \color{blue} 0.630 &  0.353 & 0.056 &  0.147 &  0.263 & \cellcolor{lightgray} \color{blue} 0.929 &  0.297 &  0.203 &  0.282 & \color{red}  0.025 \\
& JNJ   & \color{red}  0.025 & \color{red}  0.045 & 0.079 & 0.055 & 0.050 & \cellcolor{lightgray} \color{red}  0.033 & 0.058 &  0.102 & \color{red}  0.048 &  0.136 \\
& UNH   &  0.218 &  0.288 &  0.294 &  0.325 &  0.286 & \cellcolor{lightgray} 0.157 &  0.248 &  0.289 &  0.240 &  0.330 \\
& XRAY  & 0.077 &  0.185 &  0.350 &  0.207 &  0.214 & \cellcolor{lightgray} 0.105 &  0.220 &  0.340 &  0.186 &  0.409 \\
\hline
\multirow{6}{*}{Industrials} & ALK   & \color{blue} 0.677 & \color{blue} 0.480 &  0.252 &  0.398 &  0.418 & \color{blue} 0.477 & \cellcolor{lightgray} \color{blue} 0.456 &  0.244 &\color{blue}  0.542 & 0.084 \\
& FAST  &  0.240 &  0.402 &  0.387 &  0.431 &  0.394 &  0.172 & \cellcolor{lightgray} 0.326 &  0.323 &  0.354 &  0.284 \\
& GNRC  & \color{blue} 0.509 &  0.371 &  0.201 &  0.241 &  0.326 & \color{blue} 0.616 & \cellcolor{lightgray} 0.347 &  0.233 &  0.401 & 0.092 \\
& GWW   &  0.166 &  0.281 &  0.363 &  0.291 &  0.313 &  0.239 & \cellcolor{lightgray} 0.328 &  0.318 &  0.321 &  0.298 \\
& HON   &  0.121 &  0.195 &  0.221 &  0.256 &  0.195 & 0.077 & \cellcolor{lightgray} 0.151 &  0.239 &  0.138 &  0.287 \\
& UPS   & 0.082 &  0.117 &  0.145 &  0.152 &  0.114 & 0.072 & \cellcolor{lightgray} 0.142 &  0.181 &  0.124 &  0.234 \\
\hline
\multirow{6}{*}{Information Technology} & AAPL  &  0.178 &  0.278 &  0.311 &  0.270 &  0.282 &  0.123 &  0.232 & \cellcolor{lightgray} 0.288 &  0.221 &  0.302 \\
& FFIV  & \color{blue} 0.670 &  0.361 &  0.224 &  0.223 &  0.333 & \color{blue} 0.465 &  0.369 & \cellcolor{lightgray} 0.218 &  0.388 &  0.138 \\
& FTNT  & \color{blue} 0.655 &  0.291 &  0.169 &  0.107 &  0.254 & \color{blue} 0.618 &  0.345 & \cellcolor{lightgray} 0.174 &  0.384 &  0.074 \\
& MSFT  &  0.110 &  0.258 &  0.342 &  0.262 &  0.297 &  0.112 &  0.214 & \cellcolor{lightgray} 0.328 &  0.189 &  0.379 \\
& QRVO  & \color{blue} 0.696 &  0.416 &  0.145 &    - &  0.279 & \color{blue} 0.671 &  0.307 & \cellcolor{lightgray} 0.214 &  0.346 &  0.080 \\
& TEL   &  0.195 &  0.301 &  0.304 &  0.385 &  0.305 &  0.126 &  0.234 & \cellcolor{lightgray} 0.299 &  0.215 &  0.346 \\
\hline
\multirow{6}{*}{Materials} & APD   &  0.125 &  0.167 &  0.245 &  0.167 &  0.176 &  0.083 &  0.168 &  0.261 & \cellcolor{lightgray} 0.149 &  0.323 \\
& IFF   &  0.101 &  0.161 &  0.300 &  0.183 &  0.186 &  0.103 &  0.200 &  0.302 & \cellcolor{lightgray} 0.177 &  0.358 \\
& LIN   &    - &    - & \color{blue} 0.733 &    - &    - &  0.136 &  0.192 & \color{blue} 0.558 & \cellcolor{lightgray} 0.191 &  0.423 \\
& SEE   &  0.314 &  0.384 &  0.350 &  0.388 &  0.394 &  0.216 &  0.327 &  0.307 & \cellcolor{lightgray} 0.320 &  0.277 \\
& VMC   & \color{blue} 0.544 &  0.413 &  0.252 &  0.354 &  0.371 &  0.432 &  0.403 &  0.231 & \cellcolor{lightgray} 0.446 &  0.116 \\
& WRK   & \color{blue} 0.809 & \color{blue} 0.938 &  0.284 &    - & \color{blue} 0.574 &  0.390 &  0.383 &  0.270 & \cellcolor{lightgray} 0.407 &  0.197 \\
\hline
\multirow{6}{*}{Real Estate} & AMT   & \color{red} 0.044 &  0.126 &  0.261 &  0.199 &  0.129 & \color{red}  0.030 &  0.167 &  0.294 &  0.133 & \cellcolor{lightgray} 0.399 \\
& AVB   &  0.140 &  0.223 &  0.354 &  0.244 &  0.254 &  0.089 &  0.218 &  0.367 &  0.175 & \cellcolor{lightgray} \color{blue} 0.465 \\
& FRT   &  0.143 &  0.192 &  0.344 &  0.199 &  0.226 &  0.110 &  0.226 &  0.369 &  0.186 & \cellcolor{lightgray} 0.450 \\
& PLD   &  0.261 &  0.329 &  0.329 &  0.397 &  0.317 &  0.135 &  0.254 &  0.330 &  0.234 & \cellcolor{lightgray} 0.384 \\
& VNO   & \color{red}  0.034 & 0.089 &  0.255 &  0.105 &  0.108 & 0.052 &  0.144 &  0.299 & 0.098 & \cellcolor{lightgray} 0.400 \\
& WY    &  0.314 &  0.394 &  0.353 & \color{blue} 0.476 &  0.357 &  0.191 &  0.313 &  0.327 &  0.304 & \cellcolor{lightgray} 0.333 \\
\hline
\multirow{6}{*}{Utilities} & AWK  & 0.056 &  0.184 &  0.355 &  0.216 &  0.218 & \color{red}  0.027 &  0.188 &  0.399 &  0.137 & \color{blue} 0.534 \\
& DUK   & \color{red}  0.031 &  0.124 &  0.186 &  0.115 &  0.135 & \color{red}  0.023 &  0.105 &  0.260 & 0.077 &  0.342 \\
& ES    & \color{red}  0.036 &  0.125 &  0.246 &  0.140 &  0.148 & \color{red}  0.024 &  0.146 &  0.298 &  0.107 &  0.398 \\
& NEE   & \color{red}  0.036 &  0.121 &  0.227 &  0.138 &  0.138 & \color{red}  0.023 &  0.128 &  0.279 & 0.093 &  0.388 \\
& NRG   &  0.385 &  0.443 &  0.279 & \color{blue} 0.482 &  0.414 &  0.426 &  0.401 &  0.259 &  0.428 &  0.174 \\
& PNW   & \color{red}  0.030 &  0.094 &  0.233 &  0.101 &  0.129 & \color{red}  0.029 &  0.150 &  0.260 &  0.102 &  0.356 \\
\hline \hline
\end{tabular}
}
\label{tab:source_orign_25}
\end{table}
\vspace*{\fill}
\clearpage

\vspace*{\fill}
\begin{table}[ht!]
\centering
\caption{Average selection rates of source asset subsequences for MTL-50 models of individual target assets. Source assets that belong to the same GICS sector as the respective target asset are highlighted in gray. Selection rates greater than 70\% are highlighted in blue, while selection rates of less than 30\% are highlighted in red. Hyphens (-) indicate the absence of available subsequences for selection due to the later distribution date of the source asset compared to the target asset.} 
\scalebox{0.5}{
\begin{tabular}{llx{1.75cm}x{1.75cm}x{1.75cm}x{1.75cm}x{1.75cm}x{1.75cm}x{1.75cm}x{1.75cm}x{1.75cm}x{1.75cm}}
\hline \hline
\textbf{GICS Sector} & \textbf{Source Asset}  & \multicolumn{10}{c}{\textbf{Target Asset}}  \\ 
&  & \textbf{TWTR} &  \textbf{NCLH} & \textbf{LW} & \textbf{PSX} &  \textbf{SYF} & \textbf{MRNA} & \textbf{CARR} &  \textbf{DXC} & \textbf{CTVA} &  \textbf{INVH} \\ 
\hline
\multirow{6}{*}{Communication Services} & DISH  & \cellcolor{lightgray} \color{blue} 0.863 &  0.610 & 0.388 & 0.536 &  0.556 & \color{blue} 0.926 &  0.591 & 0.345 &  0.655 & \color{red} 0.179 \\
& EA    & \cellcolor{lightgray} \color{blue} 0.834 & 0.475 & 0.311 & 0.353 & 0.452 & \color{blue} 0.881 &  0.572 & 0.310 &  0.603 & \color{red} 0.182 \\
& GOOGL & \cellcolor{lightgray} 0.416 & 0.493 & 0.497 & 0.518 & 0.501 & 0.431 & 0.442 &  0.479 & 0.439 & 0.503 \\
& LUMN  & \cellcolor{lightgray} \color{red} 0.208 & 0.420 & 0.593 &  0.418 &  0.490 &  0.454 &  0.499 &  0.590 &  0.495 &  0.617 \\
& META  & \cellcolor{lightgray} \color{blue} 0.861 &  0.521 &  0.383 &  0.330 &  0.510 &  0.690 &  0.561 &  0.385 &  0.571 &  0.317 \\
& TTWO  & \cellcolor{lightgray} \color{blue} 0.831 &  0.380 & \color{red} 0.234 & \color{red} 0.229 &  0.394 & \color{blue}  0.956 &  0.566 & \color{red} 0.248 &  0.593 & \color{red} 0.104 \\
\hline
\multirow{6}{*}{Communication Discretionary} & AMZN  & \color{blue} 0.853 & \cellcolor{lightgray} 0.641 &  0.373 &  0.570 &  0.579 & \color{blue} 0.730 &  0.501 &  0.347 &  0.531 & \color{red} 0.243 \\
& EBAY  & \color{blue} 0.848 & \cellcolor{lightgray} \color{blue} 0.747 &  0.600 &  0.697 &  0.693 &  0.656 &  0.670 &  0.550 & \color{blue} 0.710 &  0.451 \\
& LEN   & \color{blue} 0.839 & \cellcolor{lightgray} 0.418 &  0.327 & \color{red} 0.217 &  0.424 & \color{blue} 0.882 &  0.588 &  0.335 & 0.627 & \color{red} 0.202 \\
& MHK   & \color{blue} 0.883 & \cellcolor{lightgray} 0.657 &  0.474 &  0.517 &  0.607 & \color{blue} 0.766 &  0.607 &  0.442 &  0.659 &  0.325 \\
& NWL   &  0.647 & \cellcolor{lightgray} 0.697 &  0.603 & \color{blue} 0.705 &  0.665 &  0.625 &  0.592 &  0.566 &  0.601 &  0.520 \\
& TSLA  &  0.559 & \cellcolor{lightgray} \color{red} 0.188 & \color{red} 0.141 & \color{red} 0.070 & \color{red} 0.219 & \color{blue} 0.968 &  0.349 & \color{red} 0.250 &  0.397 & \color{red} 0.067 \\
\hline
\multirow{6}{*}{Consumer Staples} & CPB   & \color{red} 0.093 & \color{red} 0.295 & \cellcolor{lightgray} 0.571 & \color{red} 0.282 &  0.337 & \color{red} 0.217 &  0.421 &  0.601 &  0.382 & \color{blue} 0.732 \\
& HSY   & \color{red} 0.135 & \color{red} 0.274 & \cellcolor{lightgray} 0.531 &  0.336 &  0.307 & \color{red} 0.141 &  0.367 &  0.574 &  0.314 & \color{blue} 0.725 \\
& KMB   &  \color{red} 0.075 & \color{red} 0.206 & \cellcolor{lightgray} 0.429 & \color{red} 0.232 & \color{red} 0.230 & \color{red} 0.105 &  0.303 &  0.501 & \color{red} 0.255 &  0.661 \\
& PG    &  0.059 & \color{red} 0.159 & \cellcolor{lightgray} 0.309 & \color{red} 0.186 &  \color{red} 0.165 & \color{red} 0.051 & \color{red} 0.185 &  0.398 & \color{red} 0.144 &  0.572 \\
& TAP   & \color{red} 0.268 &  0.456 & \cellcolor{lightgray} 0.617 &  0.466 &  0.502 &  0.358 &  0.508 &  0.607 &  0.478 &  0.661 \\
& WMT   & \color{red} 0.061 & \color{red} 0.184 & \cellcolor{lightgray} 0.419 & \color{red} 0.217 & \color{red} 0.218 & \color{red} 0.098 & \color{red} 0.254 &  0.494 &  \color{red} 0.203 &  0.665 \\
\hline
\multirow{6}{*}{Energy} & APA   &  0.670 &  0.685 &  0.456 & \cellcolor{lightgray} \color{blue} 0.725 &  0.634 & \color{blue} 0.800 &  0.527 &  0.435 &  0.602 &  0.327 \\
& CVX   & \color{red} 0.259 &  0.390 &  0.532 & \cellcolor{lightgray} 0.495 &  0.426 & \color{red} 0.272 &  0.414 &  0.555 &  0.376 &  0.657 \\
& DVN   & \color{blue} 0.755 & \color{blue} 0.717 &  0.434 & \cellcolor{lightgray} \color{blue} 0.730 &  0.661 & \color{blue} 0.839 &  0.553 &  0.405 &  0.620 & \color{red} 0.280 \\
& EQT   & \color{blue} 0.844 & \color{blue} 0.734 &  0.490 & \cellcolor{lightgray} \color{blue} 0.710 &  0.685 & \color{blue} 0.902 &  0.574 &  0.433 &  0.658 & \color{red} 0.275 \\
& HES   & \color{blue} 0.770 &  0.692 &  0.409 & \cellcolor{lightgray} 0.671 &  0.620 & \color{blue} 0.860 &  0.588 &  0.389 &  0.648 & \color{red} 0.257 \\
& XOM   & \color{red} 0.196 &  0.321 &  0.473 & \cellcolor{lightgray} 0.394 &  0.367 & \color{red} 0.203 &  0.348 &  0.519 & \color{red} 0.290 &  0.652 \\
\hline
\multirow{6}{*}{Financials} & AIZ   &  0.434 &  0.620 &  0.678 & \color{blue} 0.736 & \cellcolor{lightgray} 0.628 &  0.408 &  0.590 &  0.650 &  0.592 &  0.655 \\
& BRK-B & \color{red} 0.163 & \color{red} 0.235 &  0.304 &  0.309 & \cellcolor{lightgray} \color{red} 0.232 & \color{red} 0.140 & \color{red} 0.229 &  0.390 & \color{red} 0.202 &  0.526 \\
& JPM   &  0.606 &  0.665 &  0.615 & \color{blue} 0.732 & \cellcolor{lightgray} 0.642 &  0.408 &  0.544 &  0.602 &  0.521 &  0.631 \\
& LNC   & \color{blue} 0.829 &  0.645 &  0.521 &  0.531 & \cellcolor{lightgray} 0.616 & \color{blue} 0.779 &  0.668 &  0.483 & \color{blue} 0.721 &  0.354 \\
& MTB   &  0.389 &  0.473 &  0.616 &  0.563 & \cellcolor{lightgray} 0.488 &  0.389 &  0.538 &  0.617 &  0.524 &  0.656 \\
& NDAQ  & \color{blue} 0.738 & \color{blue} 0.736 &  0.651 & \color{blue} 0.724 & \cellcolor{lightgray} \color{blue} 0.713 &  0.473 &  0.607 &  0.608 &  0.598 &  0.608 \\
\hline
\multirow{6}{*}{Health Care} & BIIB  &\color{blue}  0.786 &  0.643 &  0.470 &  0.677 &  0.621 & \cellcolor{lightgray} \color{blue} 0.844 &  0.658 &  0.440 & \color{blue} 0.704 &  0.314 \\
& DVA   & \color{red} 0.277 &  0.515 &  0.662 &  0.572 &  0.537 & \cellcolor{lightgray} 0.363 &  0.508 &  0.648 &  0.495 & \color{blue} 0.706 \\
& DXCM  & \color{blue} 0.708 &  0.404 & \color{red} 0.087 & \color{red} 0.216 &  0.308 & \cellcolor{lightgray} \color{blue} 0.978 &  0.337 & \color{red} 0.200 &  0.356 & \color{red} 0.029 \\
& JNJ   & \color{red} 0.076 & \color{red} 0.144 & \color{red} 0.282 & \color{red} 0.169 & \color{red} 0.163 & \cellcolor{lightgray} \color{red} 0.103 & \color{red} 0.207 &  0.376 & \color{red} 0.164 &  0.523 \\
& UNH   &  0.540 &  0.629 &  0.597 &  0.683 &  0.623 & \cellcolor{lightgray} 0.423 &  0.556 &  0.574 &  0.544 &  0.598 \\
& XRAY  & \color{red} 0.297 &  0.540 & \color{blue} 0.707 &  0.635 &  0.584 & \cellcolor{lightgray} 0.336 &  0.537 &  0.682 &  0.525 & \color{blue} 0.736 \\
\hline
\multirow{6}{*}{Industrials} & ALK   & \color{blue} 0.933 &  0.630 &  0.379 &  0.516 &  0.569 & \color{blue} 0.949 & \cellcolor{lightgray} 0.662 &  0.345 & \color{blue} 0.745 & \color{red}  0.145 \\
& FAST  & \color{blue} 0.751 & \color{blue} 0.746 &  0.648 & \color{blue} 0.742 & \color{blue} 0.710 &  0.654 & \cellcolor{lightgray} \color{blue} 0.705 &  0.588 & \color{blue} 0.734 &  0.495 \\
& GNRC  & \color{blue} 0.854 &  0.547 & \color{red} 0.278 &  0.325 &  0.525 & \color{blue} 0.977 & \cellcolor{lightgray} 0.528 & \color{red} 0.278 &  0.595 & \color{red} 0.116 \\
& GWW   &  0.524 &  0.655 &  0.640 &  0.666 &  0.662 &  0.621 & \cellcolor{lightgray} 0.662 &  0.596 &  0.690 &  0.534 \\
& HON   &  0.322 &  0.483 &  0.519 &  0.623 &  0.488 & \color{red} 0.231 & \cellcolor{lightgray} 0.387 &  0.547 &  0.347 &  0.652 \\
& UPS   & \color{red} 0.191 &  0.308 &  0.440 &  0.384 &  0.325 & \color{red} 0.204 & \cellcolor{lightgray} 0.363 &  0.503 &  0.307 &  0.638 \\
\hline
\multirow{6}{*}{Information Technology} & AAPL  &  0.510 &  0.648 &  0.627 &  0.664 &  0.650 &  0.389 &  0.539 & \cellcolor{lightgray} 0.600 &  0.538 &  0.607 \\
& FFIV  & \color{blue} 0.871 &  0.500 &  0.383 &  0.321 &  0.493 & \color{blue} 0.802 &  0.635 & \cellcolor{lightgray} 0.378 &  0.672 & \color{red} 0.248 \\
& FTNT  & \color{blue} 0.801 &  0.371 & \color{red} 0.268 & \color{red} 0.160 &  0.366 & \color{blue} 0.938 &  0.555 & \cellcolor{lightgray} \color{red} 0.277 &  0.598 & \color{red} 0.133 \\
& MSFT  &  0.432 &  0.642 &  0.670 & \color{blue} 0.711 &  0.676 &  0.306 &  0.523 & \cellcolor{lightgray} 0.646 &  0.522 &  0.687 \\
& QRVO  & \color{blue} 0.734 &  0.459 & \color{red} 0.185 &    - &  0.356 & \color{blue} 0.945 &  0.452 & \cellcolor{lightgray} \color{red} 0.244 &  0.494 & \color{red} 0.081 \\
& TEL   &  0.541 &  0.631 &  0.616 & \color{blue} 0.714 &  0.617 &  0.375 &  0.525 & \cellcolor{lightgray} 0.601 &  0.496 &  0.640 \\
\hline
\multirow{6}{*}{Materials} & APD   & \color{red} 0.296 &  0.458 &  0.573 &  0.499 &  0.489 & \color{red} 0.244 &  0.443 &  0.577 & \cellcolor{lightgray} 0.405 &  0.674 \\
& IFF   &  0.302 &  0.467 &  0.623 &  0.513 &  0.513 &  0.357 &  0.510 &  0.614 & \cellcolor{lightgray} 0.485 &  0.653 \\
& LIN   &    - &    - &  0.689 &    - &    - & \color{red} 0.245 &  0.484 & \color{blue} 0.748 & \cellcolor{lightgray} 0.381 & \color{blue} 0.727 \\
& SEE   & \color{blue} 0.757 &  0.696 &  0.606 &  0.668 &  0.671 &  0.590 &  0.649 &  0.557 & \cellcolor{lightgray} 0.669 &  0.487 \\
& VMC   & \color{blue} 0.884 &  0.618 &  0.415 &  0.492 &  0.575 & \color{blue} 0.895 &  0.684 &  0.384 & \cellcolor{lightgray} \color{blue} 0.729 & \color{red} 0.211 \\
& WRK   & \color{blue} 0.882 & \color{blue} 0.938 &  0.480 &    - &  0.591 & \color{blue} 0.753 &  0.626 &  0.448 & \cellcolor{lightgray} 0.680 &  0.320 \\
\hline
\multirow{6}{*}{Real Estate} & AMT   & \color{red} 0.124 &  0.409 &  0.637 &  0.483 &  0.446 & \color{red} 0.184 &  0.458 &  0.652 &  0.390 & \cellcolor{lightgray} \color{blue} 0.783 \\
& AVB   &  0.371 &  0.526 &  0.690 &  0.613 &  0.551 & \color{red} 0.264 &  0.508 &  0.678 &  0.476 & \cellcolor{lightgray} \color{blue} 0.762 \\
& FRT   &  0.318 &  0.475 &  0.673 &  0.495 &  0.529 &  0.309 &  0.512 &  0.677 &  0.490 & \cellcolor{lightgray} \color{blue} 0.733 \\
& PLD   &  0.611 &  0.624 &  0.623 &  0.648 &  0.615 &  0.398 &  0.542 &  0.601 &  0.518 & \cellcolor{lightgray} 0.635 \\
& VNO   & \color{red} 0.103 & \color{red} 0.232 &  0.546 & \color{red} 0.243 &  0.318 & \color{red} 0.162 &  0.370 &  0.542 &  0.320 & \cellcolor{lightgray} 0.619 \\
& WY    & \color{blue} 0.705 &  0.697 &  0.633 & \color{blue} 0.716 &  0.649 &  0.509 &  0.614 &  0.601 &  0.620 & \cellcolor{lightgray} 0.586 \\
\hline
\multirow{6}{*}{Utilities} & AWK   & \color{red} 0.262 &  0.519 & \color{blue} 0.737 &  0.583 &  0.576 & \color{red} 0.170 &  0.482 & \color{blue} 0.738 &  0.424 & \color{blue} 0.862 \\
& DUK   & \color{red} 0.065 & \color{red} 0.250 &  0.510 &  0.315 &  0.312 & \color{red} 0.069 &  0.308 &  0.569 & \color{red} 0.248 & \color{blue} 0.760 \\
& ES    & \color{red} 0.178 &  0.360 &  0.603 &  0.426 &  0.423 & \color{red} 0.136 &  0.389 &  0.638 &  0.323 & \color{blue} 0.796 \\
& NEE   & \color{red} 0.170 &  0.343 &  0.578 &  0.371 &  0.397 & \color{red} 0.125 &  0.362 &  0.609 & \color{red} 0.297 & \color{blue} 0.767 \\
& NRG   & \color{blue} 0.774 & \color{blue} 0.749 &  0.448 & \color{blue} 0.745 &  0.665 & \color{blue} 0.797 &  0.663 &  0.427 & \color{blue} 0.708 & \color{red} 0.294 \\
& PNW   & \color{red} 0.157 &  0.311 &  0.562 &  0.318 &  0.375 & \color{red} 0.130 &  0.390 &  0.600 &  0.327 & \color{blue} 0.752 \\
\hline \hline
\end{tabular}
}
\label{tab:source_orign_50}
\end{table}
\vspace*{\fill}
\clearpage

\vspace*{\fill}
\begin{table}[ht!]
\centering
\caption{Average selection rates of source asset subsequences for MTL-75 models of individual target assets. Source assets that belong to the same GICS sector as the respective target asset are highlighted in gray. Selection rates greater than 95\% are highlighted in blue, while selection rates of less than 55\% are highlighted in red. Hyphens (-) indicate the absence of available subsequences for selection due to the later distribution date of the source asset compared to the target asset.} 
\scalebox{0.5}{
\begin{tabular}{llx{1.75cm}x{1.75cm}x{1.75cm}x{1.75cm}x{1.75cm}x{1.75cm}x{1.75cm}x{1.75cm}x{1.75cm}x{1.75cm}}
\hline \hline
\textbf{GICS Sector} & \textbf{Source Asset}  & \multicolumn{10}{c}{\textbf{Target Asset}}  \\ 
&  & \textbf{TWTR} &  \textbf{NCLH} & \textbf{LW} & \textbf{PSX} &  \textbf{SYF} & \textbf{MRNA} & \textbf{CARR} &  \textbf{DXC} & \textbf{CTVA} &  \textbf{INVH} \\ 
\hline
\multirow{6}{*}{Communication Services} & DISH  & \cellcolor{lightgray} 0.909 &  0.727 &  0.617 &  0.663 &  0.712 & \color{blue} 0.995 &  0.722 &  0.613 &  0.751 & \color{red} 0.544 \\
& EA    & \cellcolor{lightgray} 0.886 &  0.569 & \color{red} 0.462 & \color{red} 0.477 &  0.559 & \color{blue} 0.975 &  0.703 & \color{red} 0.488 &  0.717 & \color{red} 0.403 \\
& GOOGL & \cellcolor{lightgray} 0.820 &  0.777 &  0.725 &  0.757 &  0.755 &  0.802 &  0.743 &  0.724 &  0.732 &  0.724 \\
& LUMN  & \cellcolor{lightgray} 0.651 &  0.813 &  0.833 &  0.865 &  0.833 &  0.780 &  0.759 &  0.821 &  0.774 &  0.825 \\
& META  & \cellcolor{lightgray} 0.921 &  0.617 &  0.584 & \color{red} 0.387 &  0.621 &  0.892 &  0.755 &  0.602 &  0.760 &  0.553 \\
& TTWO  & \cellcolor{lightgray} 0.864 & \color{red} 0.455 & \color{red} 0.364 & \color{red} 0.320 & \color{red} 0.472 & \color{blue} 0.989 &  0.663 & \color{red} 0.399 &  0.676 & \color{red} 0.304 \\
\hline
\multirow{6}{*}{Communication Discretionary} & AMZN  &  0.908 & \cellcolor{lightgray} 0.746 &  0.611 &  0.687 &  0.723 &  0.905 &  0.688 &  0.621 &  0.695 &  0.588 \\
& EBAY  & \color{blue} 0.956 & \cellcolor{lightgray} 0.870 &  0.817 &  0.813 &  0.858 &  0.946 &  0.878 &  0.821 &  0.892 &  0.789 \\
& LEN   &  0.875 & \cellcolor{lightgray} \color{red} 0.498 & \color{red} 0.456 & \color{red} 0.291 & \color{red} 0.509 & \color{blue} 0.988 &  0.695 & \color{red} 0.490 &  0.718 & \color{red} 0.403 \\
& MHK   &  0.945 & \cellcolor{lightgray} 0.766 &  0.677 &  0.673 &  0.749 & \color{blue} 0.961 &  0.772 &  0.684 &  0.806 &  0.621 \\
& NWL   &  0.933 & \cellcolor{lightgray} 0.864 &  0.787 &  0.797 &  0.846 &  0.922 &  0.787 &  0.790 &  0.796 &  0.761 \\
& TSLA  &  0.597 & \cellcolor{lightgray} \color{red} 0.188 & \color{red} 0.173 & \color{red} 0.080 & \color{red} 0.204 & \color{blue} 0.971 & \color{red} 0.385 & \color{red} 0.212 & \color{red} 0.433 & \color{red} 0.130 \\
\hline
\multirow{6}{*}{Consumer Staples} & CPB   & \color{red} 0.444 &  0.745 & \cellcolor{lightgray} 0.938 &  0.851 &  0.788 & \color{red} 0.517 &  0.784 &  0.916 &  0.759 & \color{blue} 0.979 \\
& HSY   & \color{red} 0.399 &  0.711 & \cellcolor{lightgray} 0.911 &  0.846 &  0.747 & \color{red} 0.429 &  0.740 &  0.887 &  0.704 & \color{blue} 0.962 \\
& KMB   & \color{red} 0.291 &  0.581 & \cellcolor{lightgray} 0.854 &  0.651 &  0.655 & \color{red} 0.343 &  0.660 &  0.854 &  0.599 & \color{blue} 0.976 \\
& PG    & \color{red} 0.212 & \color{red} 0.521 & \cellcolor{lightgray} 0.771 &  0.637 &  0.594 & \color{red} 0.200 & \color{red} 0.534 &  0.790 & \color{red} 0.456 &  0.949 \\
& TAP   &  0.724 &  0.879 & \cellcolor{lightgray} 0.916 & \color{blue} 0.952 &  0.873 &  0.729 &  0.853 &  0.890 &  0.858 &  0.907 \\
& WMT   & \color{red} 0.259 &  0.595 & \cellcolor{lightgray} 0.839 &  0.769 &  0.653 & \color{red} 0.294 &  0.623 &  0.836 &  0.565 &  0.946 \\
\hline
\multirow{6}{*}{Energy} & APA   &  0.946 &  0.855 &  0.616 & \cellcolor{lightgray} 0.828 &  0.793 & \color{blue} 0.970 &  0.665 &  0.627 &  0.721 &  0.552 \\
& CVX   &  0.558 &  0.762 &  0.866 & \cellcolor{lightgray} 0.884 &  0.769 & \color{red} 0.528 &  0.745 &  0.848 &  0.719 &  0.896 \\
& DVN   & \color{blue} 0.976 &  0.865 &  0.603 & \cellcolor{lightgray} 0.849 &  0.801 & \color{blue} 0.985 &  0.677 &  0.615 &  0.727 & \color{red} 0.547 \\
& EQT   & \color{blue} 0.956 &  0.850 &  0.696 & \cellcolor{lightgray} 0.821 &  0.803 & \color{blue} 0.984 &  0.662 &  0.698 &  0.733 &  0.646 \\
& HES   & 0.945 & 0.834 & 0.588 & \cellcolor{lightgray} 0.775 &  0.768 & \color{blue} 0.972 &  0.721 & 0.597 &  0.755 & \color{red} 0.519 \\
& XOM   & \color{red} 0.457 & 0.709 &  0.852 & \cellcolor{lightgray} 0.836 &  0.744 & \color{red} 0.398 &  0.671 &  0.837 &  0.634 &  0.924 \\
\hline
\multirow{6}{*}{Financials} & AIZ   &  0.899 &  0.902 &  0.884 &  0.903 & \cellcolor{lightgray} 0.880 &  0.859 &  0.888 &  0.881 &  0.897 &  0.876 \\
& BRK-B & \color{red} 0.331 &  0.590 &  0.743 &  0.756 & \cellcolor{lightgray} 0.621 & \color{red} 0.269 & \color{red} 0.546 &  0.766 & \color{red} 0.472 &  0.908 \\
& JPM   &  0.880 &  0.865 &  0.857 &  0.830 & \cellcolor{lightgray} 0.845 &  0.719 &  0.824 &  0.847 &  0.819 &  0.853 \\
& LNC   &  0.931 &  0.759 &  0.713 &  0.667 & \cellcolor{lightgray} 0.757 & \color{blue} 0.989 &  0.812 &  0.726 &  0.845 &  0.676 \\
& MTB   &  0.729 &  0.798 &  0.852 &  0.802 & \cellcolor{lightgray} 0.797 &  0.756 &  0.815 &  0.843 &  0.827 &  0.849 \\
& NDAQ  &  0.944 &  0.865 &  0.852 &  0.818 & \cellcolor{lightgray} 0.851 &  0.777 &  0.853 &  0.840 &  0.851 &  0.839 \\
\hline
\multirow{6}{*}{Health Care} &  BIIB  & \color{blue} 0.951 &  0.800 &  0.638 &  0.830 &  0.749 & \cellcolor{lightgray} \color{blue} 0.987 &  0.798 &  0.649 &  0.823 &  0.585 \\
& DVA   &  0.796 &  0.883 &  0.901 &  0.908 &  0.873 & \cellcolor{lightgray} 0.758 &  0.848 &  0.886 &  0.851 &  0.892 \\
& DXCM  &  0.742 & \color{red} 0.361 & \color{red} 0.103 & \color{red} 0.247 & \color{red} 0.304 & \cellcolor{lightgray} \color{blue} 0.981 & \color{red} 0.387 & \color{red} 0.154 & \color{red} 0.404 & \color{red} 0.048 \\
& JNJ   & \color{red} 0.205 & \color{red} 0.485 &  0.744 &  0.589 &  0.570 & \cellcolor{lightgray} \color{red} 0.245 & \color{red} 0.532 &  0.767 & \color{red} 0.452 &  0.923 \\
& UNH   &  0.904 &  0.881 &  0.844 &  0.849 &  0.870 & \cellcolor{lightgray} 0.771 &  0.842 &  0.837 &  0.833 &  0.842 \\
& XRAY  &  0.808 &  0.888 &  0.912 &  0.902 &  0.890 & \cellcolor{lightgray} 0.806 &  0.863 &  0.900 &  0.870 &  0.904 \\
\hline
\multirow{6}{*}{Industrials} & ALK   &  0.947 &  0.730 &  0.623 &  0.646 &  0.710 & \color{blue} 0.993 & \cellcolor{lightgray} 0.778 &  0.636 &  0.824 &  0.555 \\
& FAST  & \color{blue} 0.953 &  0.897 &  0.861 &  0.846 &  0.888 &  0.942 & \cellcolor{lightgray} 0.906 &  0.856 &  0.911 &  0.837 \\
& GNRC  &  0.896 &  0.653 & \color{red} 0.459 & \color{red} 0.447 &  0.639 & \color{blue} 0.996 & \cellcolor{lightgray} 0.632 & \color{red} 0.473 &  0.673 & \color{red} 0.391 \\
& GWW   &  0.935 &  0.907 &  0.831 &  0.908 &  0.900 &  0.942 & \cellcolor{lightgray} 0.879 &  0.827 &  0.894 &  0.795 \\
& HON   &  0.700 &  0.824 &  0.847 &  0.891 &  0.814 & \color{red} 0.489 & \cellcolor{lightgray} 0.711 &  0.839 &  0.670 &  0.909 \\
& UPS   & \color{red} 0.442 &  0.691 &  0.841 &  0.865 &  0.721 & \color{red} 0.424 & \cellcolor{lightgray} 0.694 &  0.831 &  0.642 &  0.922 \\
\hline
\multirow{6}{*}{Information Technology} & AAPL  &  0.905 &  0.904 &  0.880 &  0.905 &  0.895 &  0.774 &  0.843 & \cellcolor{lightgray} 0.870 &  0.843 &  0.880 \\
& FFIV  &  0.914 &  0.598 &  0.559 & \color{red} 0.453 &  0.609 & \color{blue} 0.982 &  0.800 & \cellcolor{lightgray} 0.583 &  0.816 & \color{red} 0.508 \\
& FTNT  &  0.830 & \color{red} 0.441 & \color{red} 0.399 & \color{red} 0.225 & \color{red} 0.447 & \color{blue} 0.992 &  0.660 & \cellcolor{lightgray} \color{red} 0.430 &  0.679 & \color{red} 0.342 \\
& MSFT  &  0.908 &  0.929 &  0.895 &  0.938 &  0.917 &  0.754 &  0.837 & \cellcolor{lightgray} 0.883 &  0.829 &  0.896 \\
& QRVO  &  0.762 & \color{red} 0.377 & \color{red} 0.244 &    - & \color{red} 0.303 & \color{blue} 0.954 & \color{red} 0.532 & \cellcolor{lightgray} \color{red} 0.296 &  0.557 & \color{red} 0.213 \\
& TEL   &  0.875 &  0.862 &  0.859 &  0.864 &  0.852 &  0.728 &  0.837 & \cellcolor{lightgray} 0.849 &  0.833 &  0.859 \\
\hline
\multirow{6}{*}{Materials} & APD   &  0.716 &  0.854 &  0.896 &  0.902 &  0.854 &  0.581 &  0.801 &  0.871 & \cellcolor{lightgray} 0.793 &  0.903 \\
& IFF   &  0.774 &  0.859 &  0.888 &  0.883 &  0.855 &  0.783 &  0.854 &  0.873 & \cellcolor{lightgray} 0.857 &  0.882 \\
& LIN   &    - &    - & \color{blue} 0.956 &    - &    - &  0.583 &  0.785 & \color{blue} 0.986 & \cellcolor{lightgray} 0.773 & \color{blue} 0.995 \\
& SEE   &  0.933 &  0.830 &  0.798 &  0.789 &  0.825 & \color{blue} 0.950 &  0.864 &  0.804 & \cellcolor{lightgray} 0.871 &  0.773 \\
& VMC   &  0.939 &  0.731 &  0.632 &  0.640 &  0.716 & \color{blue} 0.997 &  0.807 &  0.638 & \cellcolor{lightgray} 0.826 &  0.560 \\
& WRK   &  0.882 &  0.938 &  0.667 &    - & \color{red} 0.541 & \color{blue} 0.992 &  0.795 &  0.680 & \cellcolor{lightgray} 0.831 &  0.615 \\
\hline
\multirow{6}{*}{Real Estate} & AMT   &  0.622 &  0.858 & \color{blue} 0.971 &  0.942 &  0.872 &  0.570 &  0.842 &  0.946 &  0.827 & \cellcolor{lightgray} \color{blue} 0.998 \\
& AVB   &  0.766 &  0.844 &  0.906 &  0.863 &  0.856 &  0.701 &  0.846 &  0.894 &  0.845 & \cellcolor{lightgray} 0.914 \\
& FRT   &  0.761 &  0.836 &  0.897 &  0.873 &  0.852 &  0.780 &  0.839 &  0.887 &  0.855 & \cellcolor{lightgray} 0.896 \\
& PLD   &  0.878 &  0.832 &  0.865 &  0.802 &  0.829 &  0.745 &  0.860 &  0.856 &  0.864 & \cellcolor{lightgray} 0.860 \\
& VNO   & \color{red} 0.383 & \color{red} 0.442 &  0.748 & \color{red} 0.282 &  0.555 &  0.582 &  0.710 &  0.746 &  0.714 & \cellcolor{lightgray} 0.765 \\
& WY    &  0.910 &  0.867 &  0.852 &  0.825 &  0.856 &  0.811 &  0.865 &  0.845 &  0.875 & \cellcolor{lightgray} 0.834 \\
\hline
\multirow{6}{*}{Utilities} & AWK   &  0.811 &  0.924 & \color{blue} 0.970 & \color{blue} 0.967 &  0.923 &  0.641 &  0.884 &  0.950 &  0.887 & \color{blue} 0.970 \\
& DUK   & \color{red} 0.297 &  0.639 &  0.896 &  0.659 &  0.755 & \color{red} 0.336 &  0.675 &  0.881 &  0.630 & \color{blue} 0.986 \\
& ES    & \color{red} 0.517 &  0.794 &  0.934 &  0.893 &  0.818 & \color{red} 0.449 &  0.749 &  0.905 &  0.720 & \color{blue} 0.976 \\
& NEE   & \color{red} 0.494 &  0.766 &  0.927 &  0.857 &  0.810 & \color{red} 0.396 &  0.743 &  0.904 &  0.714 & \color{blue} 0.975 \\
& NRG   & \color{blue} 0.986 &  0.905 &  0.626 &  0.881 &  0.846 & \color{blue} 0.988 &  0.819 &  0.633 &  0.829 &  0.571 \\
& PNW   & \color{red} 0.449 &  0.697 &  0.904 &  0.772 &  0.760 & \color{red} 0.450 &  0.737 &  0.889 &  0.708 & \color{blue} 0.965 \\
\hline \hline
\end{tabular}
}
\label{tab:source_orign_75}
\end{table}
\vspace*{\fill}
\clearpage

\section{Temporal distances between subsequence start dates and forecast origins}
\label{appdx:temp_distances}
\vspace*{\fill}
\begin{figure}[ht!]
    \centering
    \includegraphics[scale=0.23]{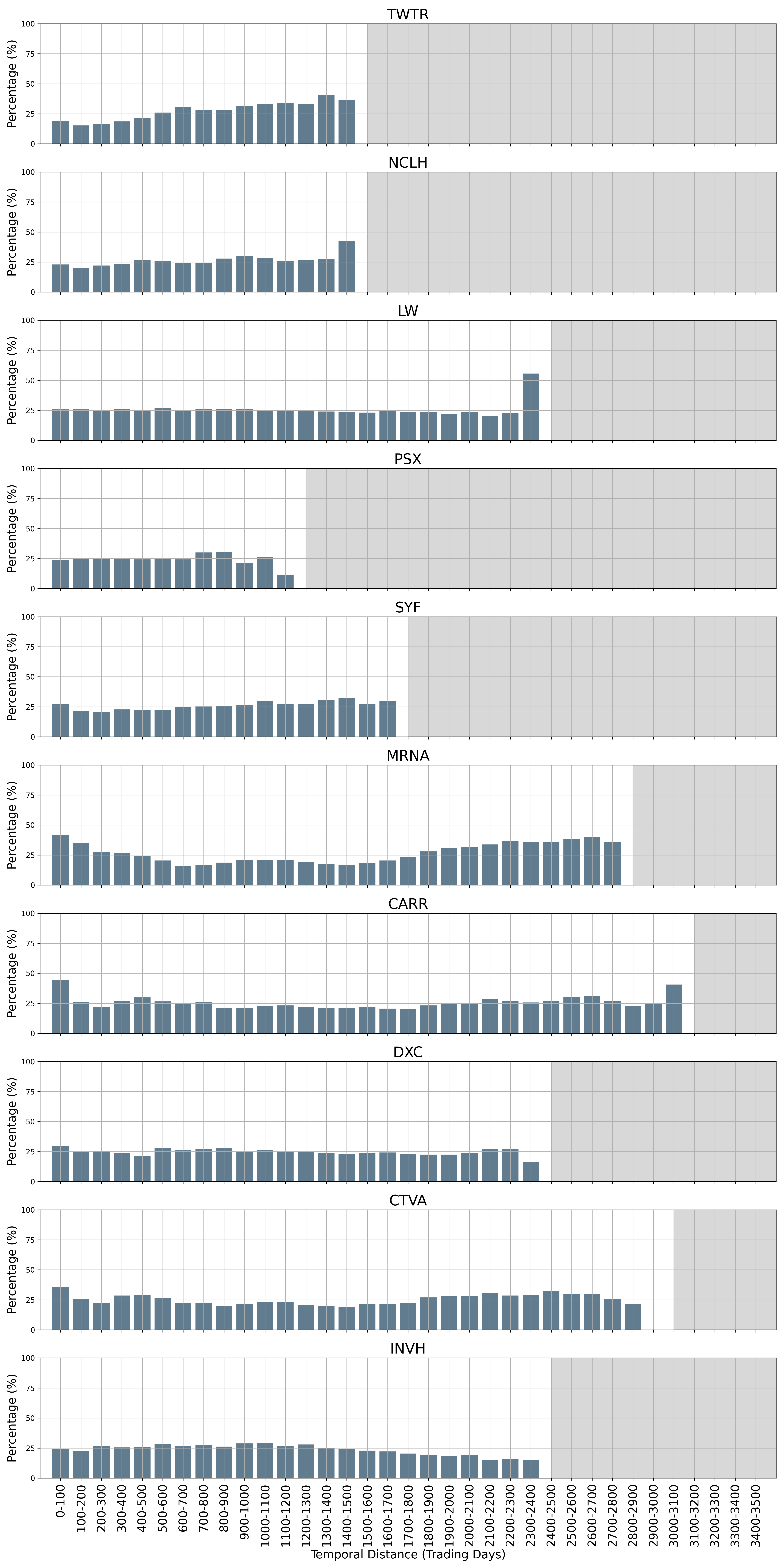} \\
    \caption{Average selection proportions of subsequences by MTL-25 models for each target asset, aggregated across all (re-)estimation steps in $s=s^{*}$. The selection rates are categorized based on the temporal distance between forecast origins and subsequence start dates, grouped into 100-trading-day intervals.}
    \label{fig:time_diff_25}
\end{figure}
\vspace*{\fill}
\clearpage

\vspace*{\fill}
\begin{figure}[ht!]
    \centering
    \includegraphics[scale=0.23]{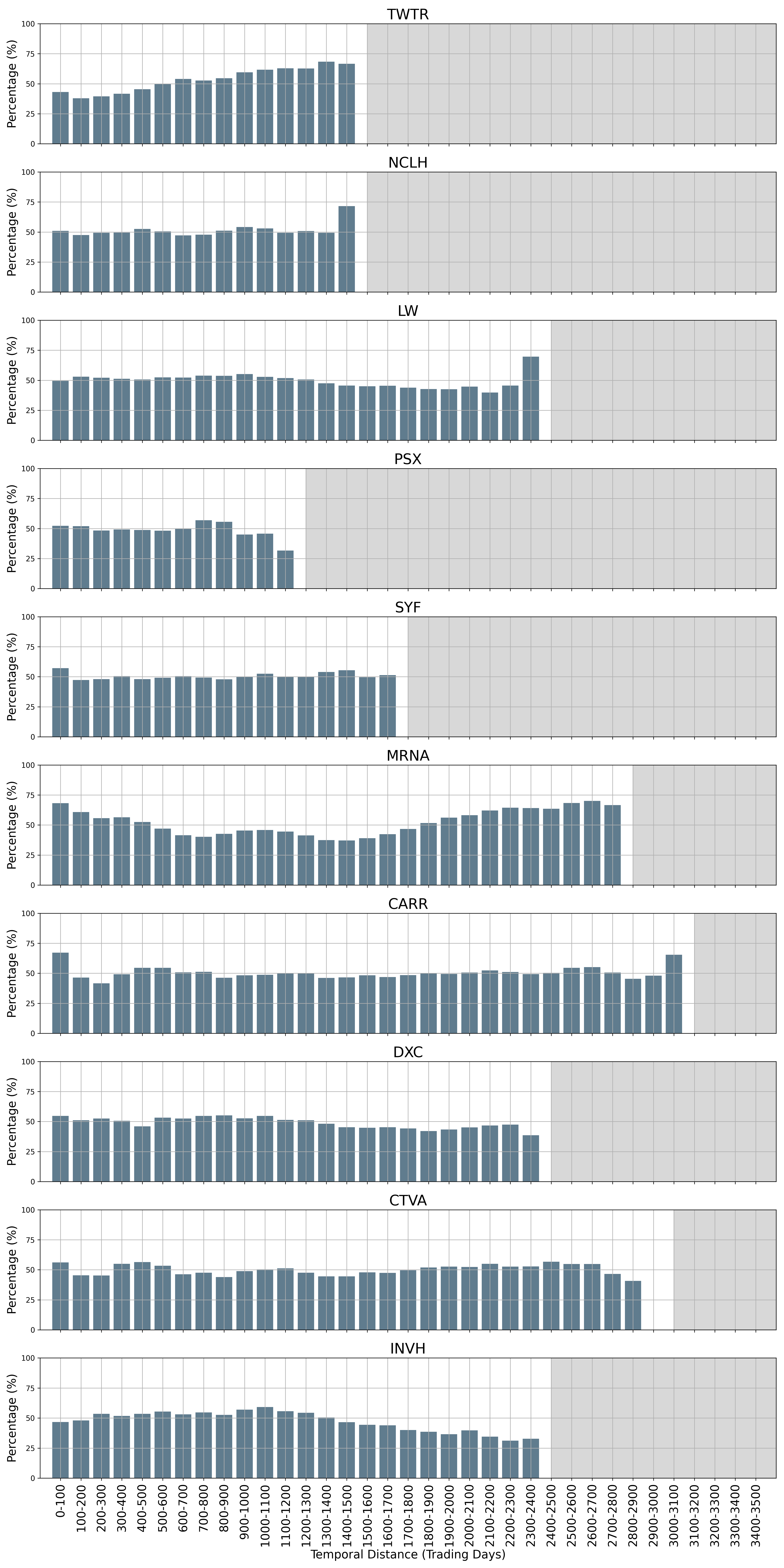} \\
    \caption{Average selection proportions of subsequences by MTL-50 models for each target asset, aggregated across all (re-)estimation steps in $s=s^{*}$. The selection rates are categorized based on the temporal distance between forecast origins and subsequence start dates, grouped into 100-trading-day intervals.}
    \label{fig:time_diff_50}
\end{figure}
\vspace*{\fill}
\clearpage

\vspace*{\fill}
\begin{figure}[ht!]
    \centering
    \includegraphics[scale=0.23]{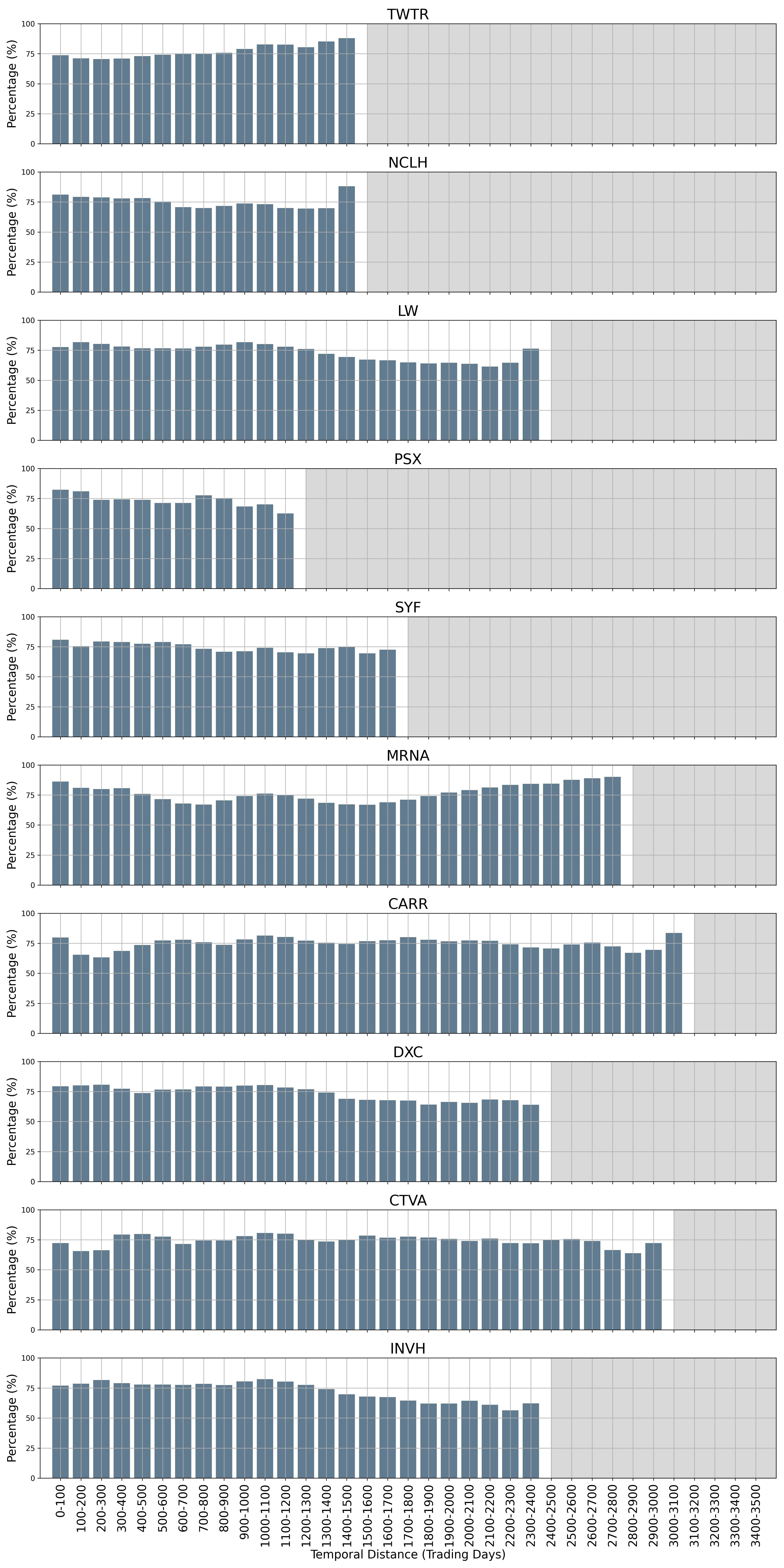} \\
    \caption{Average selection proportions of subsequences by MTL-75 models for each target asset, aggregated across all (re-)estimation steps in $s=s^{*}$. The selection rates are categorized based on the temporal distance between forecast origins and subsequence start dates, grouped into 100-trading-day intervals.}
    \label{fig:time_diff_75}
\end{figure}
\vspace*{\fill}
\clearpage

\section{Realized variance forecast results in the immediate vicinity of the first trading day}
\label{appdx:vic_of_first_day}

\vspace*{\fill}
\begin{table}[ht!]
\centering
\caption{MSEs (MAEs) of forecasting models lacking $RV_{m}$, $RV_{w}$, $MOM$, $DV$, $US3M$, and $HSI$ predictors relative to the MSE (MAE) of the NF, averaged across all new issues and spin-offs considered. The MSE and MAE error metrics are obtained for 4, 17, 28 rolling 1-day-ahead forecasts after 1, 5, and 22 public trading days. The best-performing model for each sample period (s) and error metric has been marked in bold. The term "$>$ 99" is used in this table to represent values that are exceptionally large, exceeding 99. This notation is adopted as a practical measure to avoid the inclusion of excessively large values, which could potentially compromise the clarity and readability of the table.}
\scalebox{0.8}{
\begin{tabular}{p{4.5cm}p{1.25cm}p{1.25cm}p{1.25cm}p{1.25cm}p{1.25cm}p{1.25cm}}
\hline
\hline
\multicolumn{1}{c}{Model} & \multicolumn{2}{c}{s = 1} & \multicolumn{2}{c}{s = 5} & \multicolumn{2}{c}{s = 22} \\
&  MSE & MAE & MSE & MAE & MSE & MAE\\
\hline 
1-TO HAR-STD & 4.066 & 1.715 & 1.000 & 1.016 & \textbf{0.729} & 0.894\\
1-TO HAR-EXT & 6.071 & 1.934 & $>$ 99 & 3.835 & 1.814 & 1.216\\
\hdashline
1-NP HAR-STD & 0.895 & 0.901 & 0.925 & 1.011 & 0.909 & 0.996\\
1-NP HAR-EXT & 0.823 & 0.845 & 0.970 & 1.001 & 0.973 & 0.986\\
\hdashline
1-MTL-25 HAR-STD & 0.706 & 0.862 & 0.895 & 0.927 & 0.787 & 0.868\\
1-MTL-50 HAR-STD & 0.662 & 0.845 & 0.917 & 0.930 & 0.819 & 0.882\\
1-MTL-75 HAR-STD & \textbf{0.646} & 0.845 & 0.927 & 0.939 & 0.836 & 0.893\\
\hdashline
1-MTL-25 HAR-EXT & 0.779 & 0.845 & 0.877 & 0.896 & 0.789 & 0.855\\
1-MTL-50 HAR-EXT & 0.741 & 0.844 & 0.887 & 0.896 & 0.834 & 0.886\\
1-MTL-75 HAR-EXT & 0.729 & 0.848 & 0.899 & 0.905 & 0.856 & 0.905\\
\hline 
1-TO FNN-STD  & 3.845 & 1.998 & 1.603 & 1.232 & 0.749 & 0.953\\
1-TO FNN-EXT  & 4.277 & 2.008 & 1.629 & 1.263 & 1.708 & 1.179\\
\hdashline
1-NP FNN-STD & 0.929 & 0.908 & 0.924 & 1.012 & 0.911 & 0.998\\
1-NP FNN-EXT & 0.829 & 0.829 & 0.994 & 1.008 & 1.016 & 1.009\\
\hdashline
1-MTL-25 FNN-STD & 0.706 & 0.859 & 0.895 & 0.927 & 0.787 & 0.868\\
1-MTL-50 FNN-STD & 0.663 & 0.846 & 0.917 & 0.930 & 0.819 & 0.882\\
1-MTL-75 FNN-STD & \textbf{0.646} & 0.844 & 0.927 & 0.939 & 0.835 & 0.893\\
\hdashline
1-MTL-25 FNN-EXT & 0.788 & 0.848 & 0.910 & 0.910 & 0.771 & \textbf{0.850}\\
1-MTL-50 FNN-EXT & 0.748 & 0.863 & 0.920 & 0.913 & 0.811 & 0.870\\
1-MTL-75 FNN-EXT & 0.706 & \textbf{0.843} & 0.942 & 0.926 & 0.811 & 0.890\\
\hline 
1-TO XGB-STD & 0.885 & 0.930 & 0.975 & 0.999 & 1.119 & 1.071 \\
1-TO XGB-EXT & 0.833 & 0.903 & \textbf{0.765} & \textbf{0.848} & 1.012 & 1.027\\
\hdashline
1-NP XGB-STD & 0.895 & 0.957 & 0.992 & 0.994 & 0.818 & 0.911\\
1-NP XGB-EXT & 0.853 & 0.918 & 0.950 & 0.967 & 0.791 & 0.893\\
\hdashline
1-MTL-25 XGB-STD & 0.865 & 0.934 & 0.937 & 0.939 & 0.754 & \textbf{0.850} \\
1-MTL-50 XGB-STD & 0.884 & 0.944 & 0.954 & 0.934 & 0.767 & 0.854\\
1-MTL-75 XGB-STD & 0.780 & 0.896 & 0.965 & 0.945 & 0.762 & 0.858\\
\hdashline
1-MTL-25 XGB-EXT & 1.147 & 1.013 & 0.872 & 0.880 & 0.804 & 0.872\\
1-MTL-50 XGB-EXT & 1.129 & 1.043 & 0.895 & 0.889 & 0.804 & 0.873\\
1-MTL-75 XGB-EXT & 0.894 & 0.937 & 0.898 & 0.891 & 0.830 & 0.893\\
\hline
\hline
\end{tabular}
} 
\label{tab:after_first_day_1}
\end{table}
\vspace*{\fill}
\clearpage

\vspace*{\fill}
\begin{table}[ht!]
\centering
\caption{MSEs (MAEs) of forecasting models lacking $RV_{m}$ and $MOM$ predictors relative to the MSE (MAE) of the NF, averaged across all new issues and spin-offs considered. The MSE and MAE error metrics are obtained for 17 and 28 rolling 1-day-ahead forecasts after 5 and 22 public trading days. The best-performing model for each sample period (s) and error metric has been marked in bold. The term "$>$ 99" is used in this table to represent values that are exceptionally large, exceeding 99. This notation is adopted as a practical measure to avoid the inclusion of excessively large values, which could potentially compromise the clarity and readability of the table.}
\scalebox{0.8}{
\begin{tabular}{p{4.5cm}p{1.25cm}p{1.25cm}p{1.25cm}p{1.25cm}p{1.25cm}p{1.25cm}}
\hline
\hline
\multicolumn{1}{c}{Model} & \multicolumn{2}{c}{s = 1} & \multicolumn{2}{c}{s = 5} & \multicolumn{2}{c}{s = 22} \\
&  MSE & MAE & MSE & MAE & MSE & MAE\\
\hline 
5-TO HAR-STD & - & - & 2.123 & 1.263 & 0.758 & 0.895\\
5-TO HAR-EXT & - & - & 14.433 & 2.851 & 2.248 & 1.305\\
\hdashline
5-NP HAR-STD & - & - & 0.847 & 0.936 & 0.742 & 0.891\\
5-NP HAR-EXT & - & - & 0.899 & 0.942 & 0.820 & 0.896\\
\hdashline
5-MTL-25 HAR-STD & - & - & 0.811 & 0.868 & 0.727 & 0.817\\
5-MTL-50 HAR-STD & - & - & 0.810 & 0.865 & 0.733 & 0.815\\
5-MTL-75 HAR-STD & - & - & \textbf{0.806} & 0.870 & 0.733 & 0.818\\
\hdashline
5-MTL-25 HAR-EXT & - & - & 0.847 & 0.861 & 0.714 & \textbf{0.800}\\
5-MTL-50 HAR-EXT & - & - & 0.847 & 0.865 & 0.724 & 0.802\\
5-MTL-75 HAR-EXT & - & - & 0.844 & 0.870 & 0.726 & 0.807\\
\hline 
5-TO FNN-STD & - & -  & 2.102 & 1.395 & 0.733 & 0.882\\
5-TO FNN-EXT & - & - & $>$ 99 & $>$ 99 & 1.449 & 1.107\\
\hdashline
5-NP FNN-STD & - & - & 0.832 & 0.937 & 0.748 & 0.898\\
5-NP FNN-EXT  & - & - & 0.918 & 0.954 & 0.876 & 0.929\\
\hdashline
5-MTL-25 FNN-STD  & - & - & 0.810 & 0.868 & 0.723 & 0.814\\
5-MTL-50 FNN-STD  & - & - & 0.812 & 0.871 & 0.733 & 0.816\\
5-MTL-75 FNN-STD  & - & - & 0.808 & 0.871 & 0.733 & 0.819\\
\hdashline
5-MTL-25 FNN-EXT  & - & - & 0.844 & \textbf{0.858} & 0.717 & 0.801\\
5-MTL-50 FNN-EXT  & - & - & 0.837 & 0.859 & 0.723 & 0.802\\
5-MTL-75 FNN-EXT  & - & - & 0.831 & 0.862 & 0.721 & 0.804\\
\hline 
5-TO XGB-STD & - & - & 1.028 & 0.964 & 0.989 & 1.013\\
5-TO XGB-EXT & - & - & 0.900 & 0.899 & 0.981 & 1.019\\
\hdashline
5-NP XGB-STD & - & - & 0.853 & 0.908 & 0.760 & 0.863\\
5-NP XGB-EXT & - & - & 0.902 & 0.919 & 0.744 & 0.856\\
\hdashline
5-MTL-25 XGB-STD & - & - & 0.842 & 0.879 & 0.715 & 0.820\\
5-MTL-50 XGB-STD & - & - & 0.865 & 0.887 & 0.712 & 0.822\\
5-MTL-75 XGB-STD & - & - & 0.839 & 0.883 & 0.704 & 0.819\\
\hdashline
5-MTL-25 XGB-EXT & - & - & 0.841 & 0.860 & 0.728 & 0.808\\
5-MTL-50 XGB-EXT & - & - & 0.879 & 0.873 & \textbf{0.691} & \textbf{0.800}\\
5-MTL-75 XGB-EXT & - & - & 0.863 & 0.871 & 0.712 & 0.822\\
\hline
\hline
\end{tabular}
} 
\label{tab:after_first_day_5}
\end{table}
\vspace*{\fill}
\clearpage

\vspace*{\fill}
\begin{table}[ht!]
\centering
\caption{MSEs (MAEs) of forecasting models relative to the MSE (MAE) of the NF, averaged across all new issues and spin-offs considered. The MSE and MAE error metrics are obtained for 28 rolling 1-day-ahead forecasts after 22 public trading days. The best-performing model for each sample period (s) and error metric has been marked in bold.}
\scalebox{0.8}{
\begin{tabular}{p{4.5cm}p{1.25cm}p{1.25cm}p{1.25cm}p{1.25cm}p{1.25cm}p{1.25cm}}
\hline
\hline
\multicolumn{1}{c}{Model} & \multicolumn{2}{c}{s = 1} & \multicolumn{2}{c}{s = 5} & \multicolumn{2}{c}{s = 22} \\
&  MSE & MAE & MSE & MAE & MSE & MAE\\
\hline 
22-TO HAR-STD & - & - & - & - & 19.434 & 1.593\\
22-TO HAR-EXT & - & - & - & - & 72.575 & 3.190\\
\hdashline
22-NP HAR-STD & - & - & - & - & 0.727 & 0.880\\
22-NP HAR-EXT & - & - & - & - & 0.819 & 0.894\\
\hdashline
22-MTL-25 HAR-STD & - & - & - & - & 0.700 & 0.810\\
22-MTL-50 HAR-STD & - & - & - & - & 0.693 & 0.808\\
22-MTL-75 HAR-STD & - & - & - & - & 0.688 & 0.809\\
\hdashline
22-MTL-25 HAR-EXT & - & - & - & - & 0.690 & 0.786\\
22-MTL-50 HAR-EXT & - & - & - & - & 0.678 & \textbf{0.778}\\
22-MTL-75 HAR-EXT & - & - & - & - & \textbf{0.672} & 0.781\\
\hline 
22-TO FNN-STD & - & - & - & - & 1.475 & 1.271\\
22-TO FNN-EXT & - & - & - & - & 1.597 & 1.286\\
\hdashline
22-NP FNN-STD & - & - & - & - & 0.694 & 0.854\\
22-NP FNN-EXT & - & - & - & - & 0.975 & 0.972\\
\hdashline
22-MTL-25 FNN-STD & - & - & - & - & 0.698 & 0.812\\
22-MTL-50 FNN-STD & - & - & - & - & 0.693 & 0.809\\
22-MTL-75 FNN-STD & - & - & - & - & 0.688 & 0.810\\
\hdashline
22-MTL-25 FNN-EXT & - & - & - & - & 0.703 & 0.791\\
22-MTL-50 FNN-EXT & - & - & - & - & 0.697 & 0.793\\
22-MTL-75 FNN-EXT & - & - & - & - & 0.690 & 0.794\\
\hline 
22-TO XGB-STD & - & - & - & - & 0.884 & 0.929\\
22-TO XGB-EXT & - & - & - & - & 0.854 & 0.928\\
\hdashline
22-NP XGB-STD & - & - & - & - & 0.748 & 0.859\\
22-NP XGB-EXT & - & - & - & - & 0.745 & 0.851\\
\hdashline
22-MTL-25 XGB-STD & - & - & - & - & 0.689 & 0.822\\
22-MTL-50 XGB-STD & - & - & - & - & 0.685 & 0.814\\
22-MTL-75 XGB-STD & - & - & - & - & 0.688 & 0.826\\
\hdashline
22-MTL-25 XGB-EXT & - & - & - & - & 0.718 & 0.819\\
22-MTL-50 XGB-EXT & - & - & - & - & 0.728 & 0.815\\
22-MTL-75 XGB-EXT & - & - & - & - & 0.747 & 0.831\\
\hline
\hline
\end{tabular}
} 
\label{tab:after_first_day_22}
\end{table}

\vspace*{\fill}

\clearpage

\bibliographystyle{apacite} 
\interlinepenalty=10000
\bibliography{ref}

\end{document}